\documentclass{article}

\usepackage{times}
\usepackage{epsfig}
\usepackage{graphicx}
\usepackage{amsmath}
\usepackage{amssymb}

\usepackage{booktabs}       
\usepackage{amsfonts}       
\usepackage{nicefrac}       
\usepackage{microtype}      
\usepackage{scrextend}

\usepackage{color}

\newcommand{\comment}[1]{{}}

\usepackage[utf8]{inputenc} 
\usepackage[T1]{fontenc}    
\usepackage{url}            

\usepackage{graphicx}
\graphicspath{{graphs/}}

\usepackage{afterpage}
\usepackage{placeins}

\usepackage{caption}
\usepackage{subcaption}
\usepackage{float}
\newcommand{\app}{{Suppl}}

\usepackage{amsmath,amsfonts,bm}

\def\eqref#1{equation~\ref{#1}}

\def\1{\bm{1}}

\def\vx{{\bm{x}}}
\def\vy{{\bm{y}}}

\DeclareMathAlphabet{\mathsfit}{\encodingdefault}{\sfdefault}{m}{sl}
\SetMathAlphabet{\mathsfit}{bold}{\encodingdefault}{\sfdefault}{bx}{n}

\def\sN{{\mathbb{N}}}

\def\sR{{\mathbb{R}}}

\def\sX{{\mathbb{X}}}

\usepackage{hyperref}

\usepackage[accepted]{icml2020}

\icmltitlerunning{Let's Agree to Agree: Neural Networks Share Classification Order on Real Datasets}

\begin{document}

\twocolumn[
\icmltitle{Let's Agree to Agree: Neural Networks Share Classification Order on Real Datasets}



\icmlsetsymbol{equal}{*}

\begin{icmlauthorlist}
\icmlauthor{Guy Hacohen}{huji,elsc}
\icmlauthor{Leshem Choshen}{huji}
\icmlauthor{Daphna Weinshall}{huji}
\end{icmlauthorlist}

\icmlaffiliation{huji}{School of Computer Science and Engineering, The Hebrew University of Jerusalem, Jerusalem, Israel}
\icmlaffiliation{elsc}{Edmond \& Lily Safra Center for Brain Sciences, The Hebrew University of Jerusalem, Jerusalem, Israel}

\icmlcorrespondingauthor{Guy Hacohen}{guy.hacohen@mail.huji.ac.il}
\icmlcorrespondingauthor{Leshem Choshen}{leshem.choshen@mail.huji.ac.il}
\icmlcorrespondingauthor{Daphna Weinshall}{daphna@cs.huji.ac.il}

\icmlkeywords{Machine Learning, ICML, machine learning, deep learning, computer vision, neural networks, agreement, tp-agreement, classification order, learning order, network comparison, natural datasets, learning dynamics, Natural language processing, CV, nlp, computational linguistics, cl, learning theory, ml, ai}

\vskip 0.3in
]



\printAffiliationsAndNotice{}  

\begin{abstract}
We report a series of robust empirical observations, demonstrating that deep Neural Networks learn the examples in both the training and test sets in a similar order. This phenomenon is observed in all the commonly used benchmarks we evaluated, including many image classification benchmarks, and one text classification benchmark. While this phenomenon is strongest for models of the same architecture, it also crosses architectural boundaries -- models of different architectures start by learning the same examples, after which the more powerful model may continue to learn additional examples. We further show that this pattern of results reflects the interplay between the way neural networks learn benchmark datasets. Thus, when fixing the architecture, we show synthetic datasets where this pattern ceases to exist. When fixing the dataset, we show that other learning paradigms may learn the data in a different order. We hypothesize that our results reflect how neural networks discover structure in natural datasets.

\end{abstract}

\section{Introduction}

Typically, neural networks (NN) in common use include a large number of parameters and many non-linearities, resulting in a highly non-convex, high-dimensional optimization landscape. These models are usually trained with some variant of Stochastic Gradient Descent (SGD), initialized randomly, thus introducing stochasticity into the training procedure both in the form of initialization and the sampling of gradients. As a result, training the same neural architecture several times generates models with drastically different weights \citep{li2015convergent, yosinski2015understanding}, which may correspond to different local minima of the optimization landscape.

\begin{figure}[t]
  \includegraphics[width=1\linewidth]{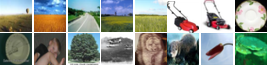}

  \includegraphics[width=1\linewidth]{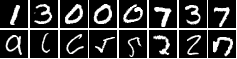}
  \caption{Images learned at the beginning (top row) and the end of the training (bottom row). Top: CIFAR-100, bottom: MNIST.}
  \label{fig:easy_and_hard_images}
  \vspace{-.2in}
\end{figure}

For most practical purposes, models trained with the same dataset, architecture and training protocols are considered similar, since they tend to have similar accuracy. Neural comparison methods \cite{li2016convergent,raghu2017svcca,morcos2018insights} tend to find them to be more similar than models trained on different datasets, based on different criteria. Nevertheless, many applications consider such models to be distinct enough to merit the use of ensemble methods \cite{le2015tiny, Kantor2019LearningTC} and average training epochs \cite{Vaswani2017AttentionIA,JunczysDowmunt2018MarianFN}. All in all, the question of how to meaningfully evaluate the similarity between trained neural models remains open.
 
Contemporary approaches tend to measure the similarity between neural models by comparing their underlying compositional properties, as done for example in \citet{lenc2015understanding,alain2016understanding,li2016convergent,raghu2017svcca,wang2018towards,cohen2019separability}. Instead, we propose to measure similarity via the direct comparison of their classification predictions per example, as formally defined in \S\ref{sec:notations}. Thus we show empirically that for a wide range of classification benchmarks, including ImageNet, CIFAR, and text classification (see \S\ref{section:robusteness}), models of the same architecture classify the data similarly (see \S\ref{section:self-consistency}) during the entire learning process. As long as the models share the same architecture, this similarity is independent of such choices as optimization and initialization methods, hyper-parameter values, the detailed architecture or the particular dataset. The similarity can be replicated for a given test set even when each model is trained on a different training set, as long as both sets of training data are sampled from the same distribution (see \S\ref{section:different_training_sets}).

The similarity between different models is not restricted to their accuracy at the end of the training, but rather can be observed throughout the entire learning process (see \S\ref{section:self-consistency}). Specifically, trained models exhibit a similar classification profile in every epoch from the beginning to the end of the training. Combined with the observation that once an example is classified correctly by some model it is rarely misclassified after further training (see \app~\ref{app:individual_learning_dynamics}), we conclude that each architecture learns to classify each benchmark dataset in a specific order, which can be seen in all its models. See examples for CIFAR-100 and MNIST in Fig.~\ref{fig:easy_and_hard_images}.

The order in which a dataset is learned seems robust across architectures. In \S\ref{section:cross-consistency} we train several commonly used architectures on the same dataset, resulting in a highly correlated learning order, in spite of there being significant differences in the final accuracy of the architectures. In fact, when tracking the training process of different architectures, we observe that the models of the stronger architecture first learn the examples which have been learned by the weaker architecture, and only then continue to learn new examples.

Is it possible that this robust similarity is an artifact of the training procedure of NNs, and specifically the use of SGD optimization? In \S\ref{sec:4-phases} we describe examples of hand-crafted image datasets where these patterns of similarity disappear, suggesting that this is not the case. In these synthetic datasets, which NNs learn successfully, even models from the same architecture seem to learn to classify examples in a different order, depending on their random initialization and mini-batch sampling. Moreover, when training NNs on a dataset with randomly shuffled labels \cite{zhang2016understanding}, in which no generalization is possible, we find that different models memorize the data in a different order\footnote{Compare with \citet{morcos2018insights}, where it is shown that NNs that generalize are more similar than those that memorize.}.

Is it possible that this robust similarity is an artifact of the structure of each dataset, and the way it is being discovered? In other words, is it all about typical points being learned before atypical points, regardless of the nature of the classifier? We show that this is not the case, and that the order by which benchmark datasets are learned is unique to NNs. Specifically, in \S\ref{section:cross-consistency-other} we analyze the order in which a benchmark dataset is learned by an AdaBoost classifier that employs weak linear classifiers. We show that this order has a low correlation with the order observed when training NNs on the same dataset.

The empirical observations described above seem to echo the interplay between how NNs learn, and the complexity of datasets which are used to evaluate these NNs. Possibly they also reflect the way NNs discover structure in a given dataset, where learning order corresponds with data complexity. In other words, some examples are consistently easier than others for NNs to learn. (We discuss the relationship with curriculum learning and hard data mining in \S\ref{sec:discussion}.) Notably, we failed to find a real dataset for which NNs differ. This may indicate the existence of a common structure in real datasets, which our synthetic datasets do not exhibit. 

\subsection{Summary of Contribution}
\begin{itemize}
    \item We propose a direct way to compare between different neural models termed TP-agreement (\S\ref{sec:notations}).
    \item We empirically show that models that share the same architecture learn real datasets in the same order (\S\ref{section:self-consistency}), and propose a measure to this effect termed accessibility score (\S\ref{sec:notations}).
    \item We argue that the learning order emerges from the coupling of neural architectures and benchmark datasets. To support shis, we show that neural architectures can learn synthetic datasets without any specific order (\S\ref{sec:4-phases}), and likewise non-neural architectures can learn benchmark datasets without any specific order (\S\ref{section:cross-consistency-other}).
    \item We show that models with different architectures can learn benchmark datasets at a different pace and performance, while still inducing a similar order (\S\ref{section:cross-consistency}). Specifically, we see that stronger architectures start off by learning the same examples that weaker networks learn, then move on to learning new examples.  
\end{itemize}

\section{Methodology and Notations}
\label{sec:notations}

Consider some architecture $f$, and a labeled dataset $\sX = \left\{(\vx_i,y_i)\right \}_{i=1}^M$ where $\vx_i\in\sR^d$ denotes a single example and $y_i\in [K]$ its corresponding label. We create a collection of $N$ models of $f$, denoted $\mathcal{F}^e=\{f_1^e,...,f_{N}^e\}$. Each model $f_i^e\in\mathcal{F}^e$ is initialized independently using the same distribution\footnote{We observe similar results with various commonly used initialization methods, see \app~\ref{app:additional_results_robust}.}, then trained with SGD on randomly sampled mini-batches for $e\in\sN$ epochs. Let $\mathcal{S}_E$ denote the set of different extents (total epochs of $\sX$) used to train $f$.

\begin{figure}[t]
\centering
  \includegraphics[width=.9\linewidth]{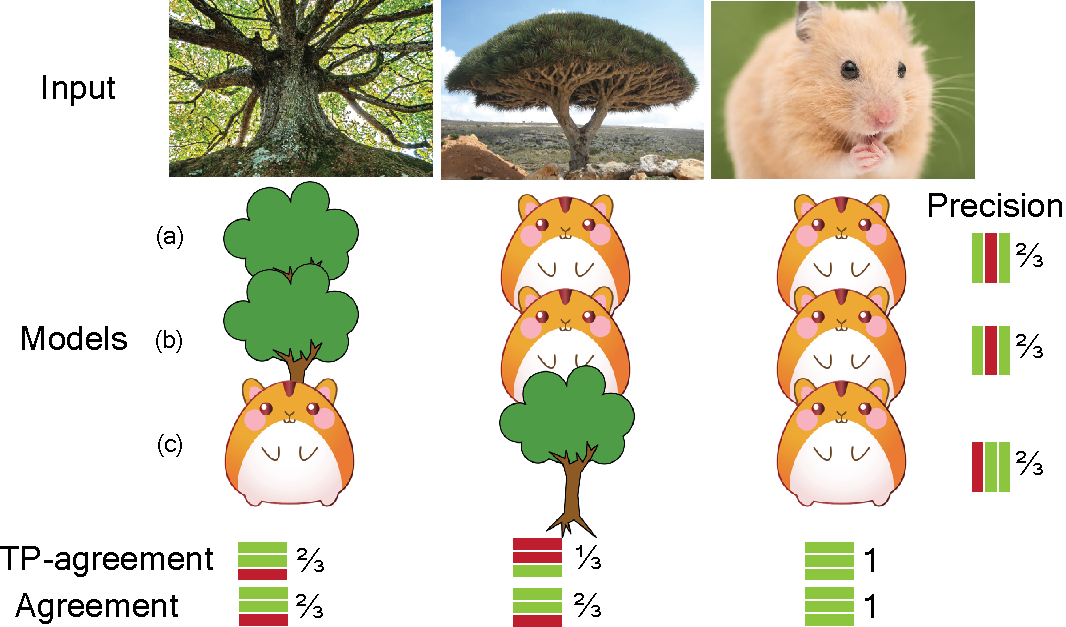}
  \caption{Illustration of TP-agreement and agreement scores. Each row (a,b,c) corresponds to a model performing a binary classification task of classifying between images of trees and hamsters. All models have $\frac{2}{3}$ precision on the task -- which can be calculated as the row-average of the correct classifications. However, examples show different TP-agreement scores -- which can be calculated as the column-average of the correct classifications.}
  \label{fig:agreement_scores_illustration}
  \vspace{-.2in}
\end{figure}

We represent each model $f_i^e\in\mathcal{F}^e$ by binary classification vectors that capture the accuracy per sample (e.g. in train or test). Each vector's dimension corresponds to the size of the dataset; each element is assigned 1 if the NN classifies the corresponding example correctly, and 0 otherwise. 
We use these vector representations to define and analyze the \emph{True-Positive agreement} (TP-agreement) of $\mathcal{F}^e$, when each model $f_i^e\in\mathcal{F}^e$ is trained on $\sX$ from scratch for $e$ epochs. We analyze the TP-agreement throughout the entire learning process, for different values of $e\in \mathcal{S}_E$.

\begin{figure*}[t]
\begin{subfigure}{.83\textwidth}
  
    \begin{subfigure}{.151\textwidth}
      \centering
      \includegraphics[width=1\linewidth]{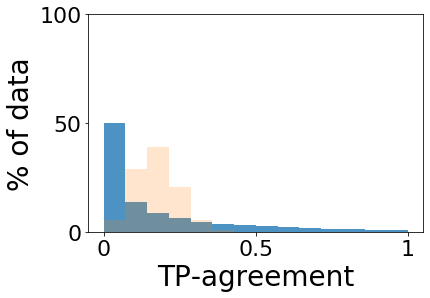}
    \end{subfigure}
    \begin{subfigure}{.151\textwidth}
      \centering
      \includegraphics[width=1\linewidth]{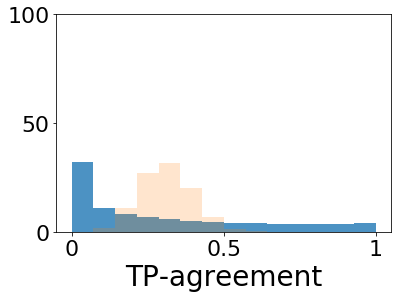}
    \end{subfigure}
    \begin{subfigure}{.151\textwidth}
      \centering
      \includegraphics[width=1\linewidth]{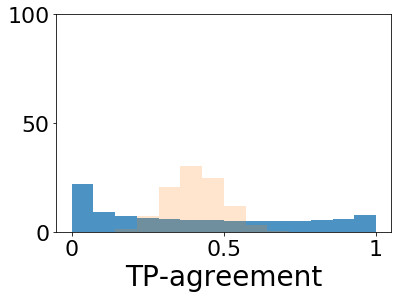}
    \end{subfigure}
    \begin{subfigure}{.151\textwidth}
      \centering
      \includegraphics[width=1\linewidth]{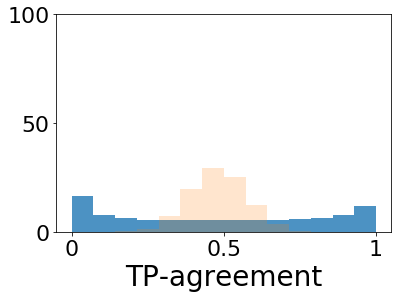}
    \end{subfigure}
    \begin{subfigure}{.151\textwidth}
      \centering
      \includegraphics[width=1\linewidth]{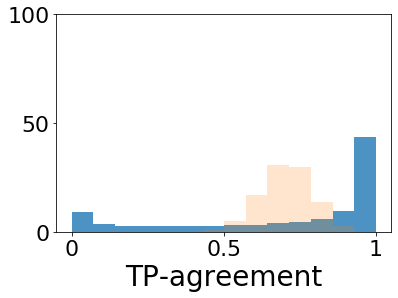}
    \end{subfigure}
    \begin{subfigure}{.151\textwidth}
      \centering
      \includegraphics[width=1\linewidth]{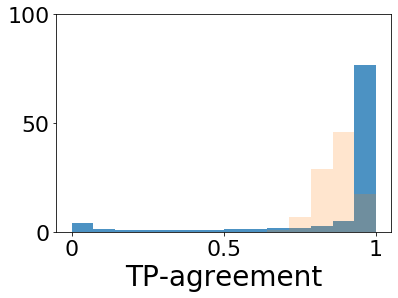}
    \end{subfigure}
    
    \begin{subfigure}{.151\textwidth}
      \centering
      \includegraphics[width=1\linewidth]{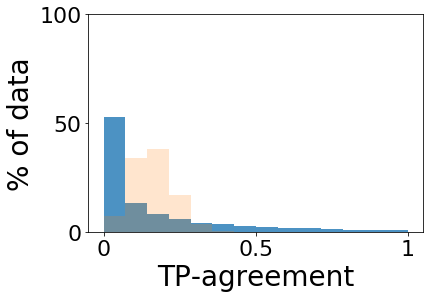}
    \end{subfigure}
    \begin{subfigure}{.151\textwidth}
      \centering
      \includegraphics[width=1\linewidth]{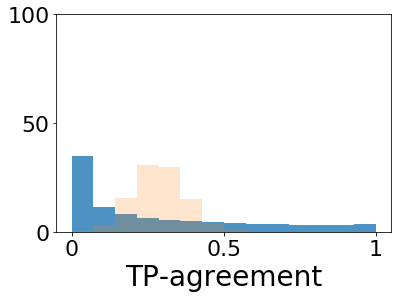}
    \end{subfigure}
    \begin{subfigure}{.151\textwidth}
      \centering
      \includegraphics[width=1\linewidth]{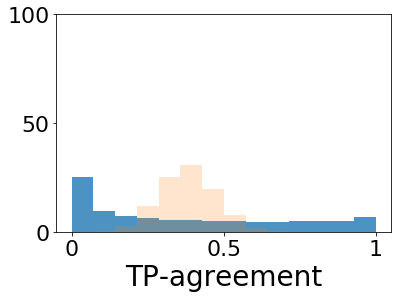}
    \end{subfigure}
    \begin{subfigure}{.151\textwidth}
      \centering
      \includegraphics[width=1\linewidth]{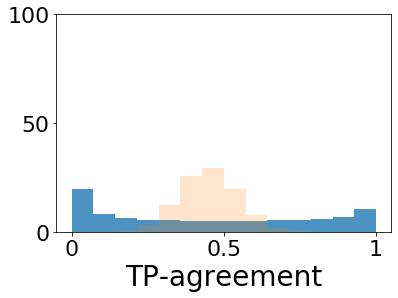}
    \end{subfigure}
    \begin{subfigure}{.151\textwidth}
      \centering
      \includegraphics[width=1\linewidth]{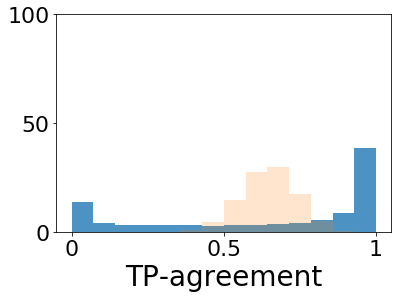}
    \end{subfigure}
    \begin{subfigure}{.151\textwidth}
      \centering
      \includegraphics[width=1\linewidth]{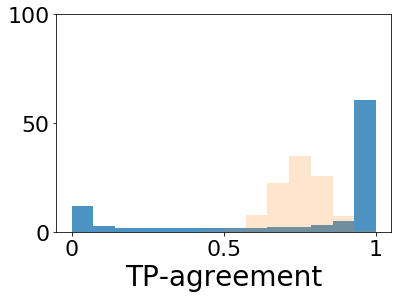}
    \end{subfigure}
\caption{TP-agreement distribution per epoch}
\label{fig:tp-agreement-distribution-imagenet-per-epoch}
\end{subfigure}
\begin{subfigure}{.16\textwidth}
    \begin{subfigure}{1\textwidth}
      \centering
      \includegraphics[width=1\linewidth]{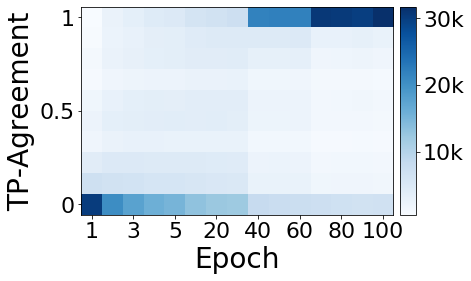}
    \end{subfigure}

    \begin{subfigure}{1\textwidth}
      \centering
      \includegraphics[width=1\linewidth]{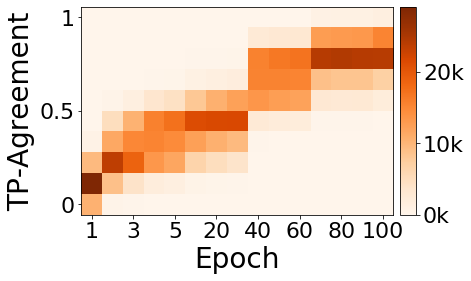}
    \end{subfigure}
\caption{All epochs}
\label{fig:tp-agreement-distribution-imagenet-total}
\end{subfigure}

\caption{The distribution of TP-agreement scores throughout the entire learning process, for $27$ models of ResNet-50 trained over ImageNet. a) The distribution at specific epochs, where epochs $1, 2, 5, 30, 40, 100$ are shown respectively from left to right. In each plot, the null hypothesis is shown in fading orange, describing the distribution of TP-agreement scores using random classification vectors with matching accuracy. Top: train data, bottom: validation data. b) Combined TP-agreement distribution during the entire learning process. The $X$-axis corresponds to epochs, while the $Y$-axis corresponds to the TP-agreement score. Intensity depicts the corresponding number of images achieving each score in each epoch. Top: combined distribution using the validation set of ImageNet, which illustrates that the models learn in a similar order (see text). Bottom: combined distribution of the null hypothesis of independent models with similar accuracy. Note the clear qualitative difference between the two cases: real classifiers (top) show a clear bi-modal behavior, independent classifiers (bottom) show a uni-modal behavior.}
  \label{fig:consistency_dynamics}
\end{figure*}

\begin{figure*}[htb]
\begin{subfigure}{.22\textwidth}
  \centering
  \includegraphics[width=1\linewidth]{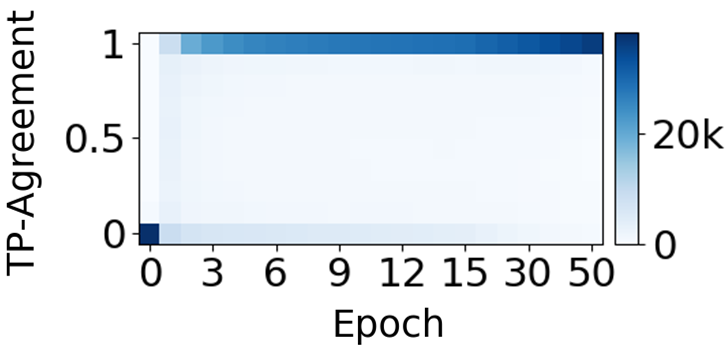}

  \includegraphics[width=1\linewidth]{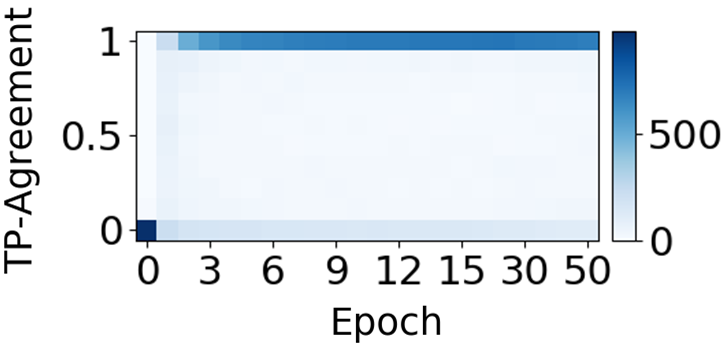}
  \caption{NLP task}
  \label{fig:nlp-task-tp-agreement}
\end{subfigure}
\begin{subfigure}{.2\textwidth}
  \centering
  \includegraphics[width=1\linewidth]{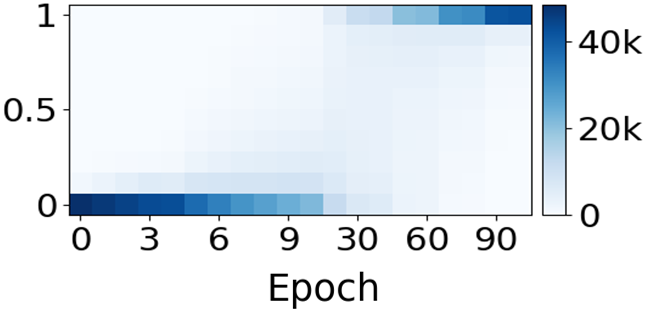}

  \includegraphics[width=1\linewidth]{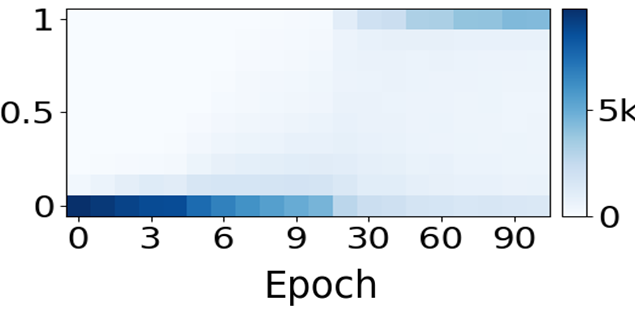}
  \caption{CIFAR-100}
  \label{fig:cifar100-tp-agreement}
\end{subfigure}
\begin{subfigure}{.2\textwidth}
  \centering
  \includegraphics[width=1\linewidth]{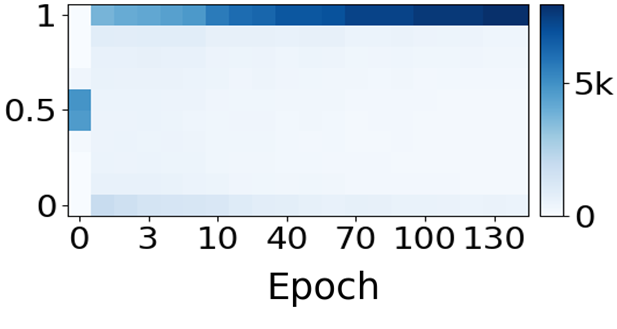}

  \includegraphics[width=1\linewidth]{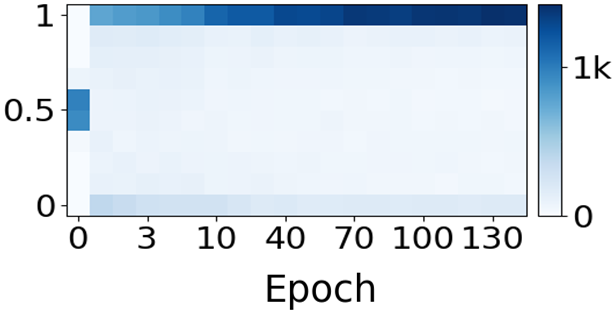}
  \caption{Cats and Dogs}
  \label{fig:cats-and-dogs-tp-agreement}
\end{subfigure}
\begin{subfigure}{.2\textwidth}
  \centering
  \includegraphics[width=1\linewidth]{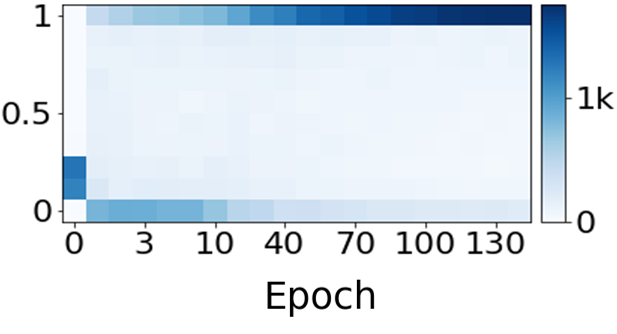}

  \includegraphics[width=1\linewidth]{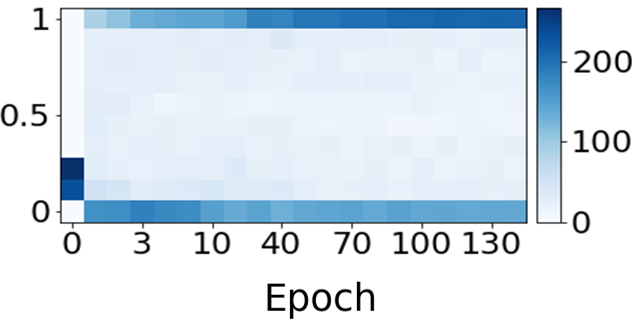}
  \caption{Small mammals}
  \label{fig:small-mammals-tp-agreement}
\end{subfigure}
\begin{subfigure}{.16\textwidth}
  \centering
  \includegraphics[width=1\linewidth]{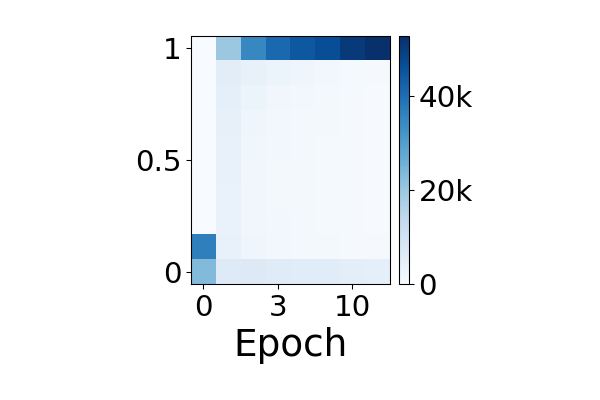}

  \includegraphics[width=1\linewidth]{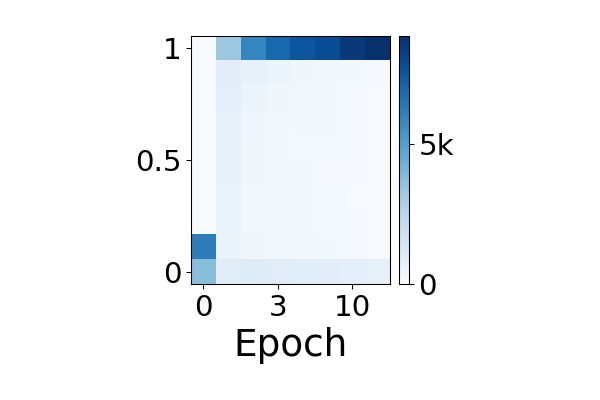}
  \caption{MNIST}
  \label{fig:mnist-tp-agreement}
\end{subfigure}

\caption{The combined distribution of TP-agreement scores during the entire learning process of both train (top) and test (bottom) datasets with several architectures and datasets. As in Fig.~\protect\ref{fig:consistency_dynamics}, the $X$-axis corresponds to epochs and the $Y$-axis corresponds to the TP-agreement score, while intensity depicts the corresponding number of images achieving each score in each epoch. Cases shown: a) $100$ attention based BiLSTM models trained on text classification; b) $20$ VGG-19 models trained on CIFAR-100; c) $100$ models of st-VGG trained on the Cats and Dogs dataset; d) $100$ models of st-VGG trained on the small-mammals dataset; e) $100$ models of a small architecture trained on MNIST (see \app~\ref{app:architectures},\ref{app:datasets} for architectures and datasets). In all cases, both the train and test datasets are learned in the same order.}
  \label{fig:full_results_multiple_datasets}
\end{figure*}

More specifically, for each epoch $e\in \mathcal{S}_E$ we define the \textbf{TP-agreement} of example $(\vx, \vy)$ as follows: 
\begin{equation*}
    TPa^e(\vx, y)=\frac{1}{N}\sum_{i=1}^{N}\1_{\left[{f^e_i(\vx)=y}\right]}
\end{equation*}
The TP-agreement $TPa^e(\vx, y)$ of example $(\vx, y)$ measures the average accuracy on $(\vx, y)$ of $N$ networks which were trained for exactly $e$ epochs each. Unlike precision, TP-agreement measures the average accuracy of a single example over multiple models, as opposed to precision which measures the average accuracy of a single model over multiple examples (see Fig.~\ref{fig:agreement_scores_illustration} for illustration). Note that TP-agreement does not take into account the classifiers' agreement when they misclassify. We, therefore, define a complementary \textbf{agreement} score that measures the agreement among the classifiers -- the largest fraction of classifiers that predict the same label for $\vx$:
\begin{equation*}
    a^e(\vx)=\max_{k\in\left[K\right]}\frac{1}{N}\sum_{i=1}^{N}\1_{\left[{f^e_i(\vx)=k}\right]}
\end{equation*}

Fig.~\ref{fig:agreement_scores_illustration} illustrates these definitions. When all the classifiers in $\mathcal{F}^e$ are identical, the TP-agreement of each example is either $0$ or $1$, and its agreement is $1$. This results in a perfect bi-modal distribution of the TP-agreement scores over the examples. On the other hand, 
if the identity of the correctly classified examples per classifier is independent,
the distribution of both scores is expected to resemble a Gaussian\footnote{For large $N$ this follows from the central limit theorem.\label{note:central-limit}}. Its center is the average accuracy of the classifiers in $\mathcal{F}^e$, in the case of the TP-agreement, and a slightly higher value for the Agreement score. It follows that the higher the mean agreement is, and the more bi-modal-like the distribution of TP-agreement is around $0$ and $1$, the more similar the set of classifiers is. We measure the bi-modality of random variable $X$ using the following score:  $kurtosis(X)-skewness^2(X)-1$ \cite{pearson1894contributions}; the lower the score is, the more bi-modal $X$ is.

As shown in the next section, the distribution of TP-agreement over examples is mostly bi-modal during the entire learning process. Specifically, for most examples the TP-agreement is $0$ at the beginning of learning, and then rapidly changes to $1$ at some point during learning (see \app~\ref{app:additional_results}). This property suggests that data is learned in a specific order. To measure how fast an example is learned by some architecture $f$, we note that the faster the example is learned, the higher its TP-agreement for all epochs $e\in\mathcal{S}_E$ must be. Therefore, we define the \textbf{accessibility score} of an example to be its averaged TP-agreement over all epochs, formally given by $\mathbb{E}_{e\in \mathcal{S}_E}\left[TPa^e(\vx, y)\right]$.

\section{Diversity in a Single Architecture}
\label{section:self-consistency}

In this section, we investigate collections of classifiers obtained from a single architecture $f$.

\subsection{Same Training Set}
\label{section:same_training_set}

We start with the simplest condition, where all models in collection $\mathcal{F}$ are obtained by training with the same training set $\sX$, with different initial conditions and with independently sampled mini-batches. When using common benchmark datasets, the TP-agreement distributions over both train and test sets are bi-modal, see Figs.~\ref{fig:consistency_dynamics}, \ref{fig:full_results_multiple_datasets} and \app~\ref{app:additional_results_robust}. 

\begin{figure*}[htb]
\begin{subfigure}{.158\textwidth}
  \centering
  \includegraphics[width=1\linewidth]{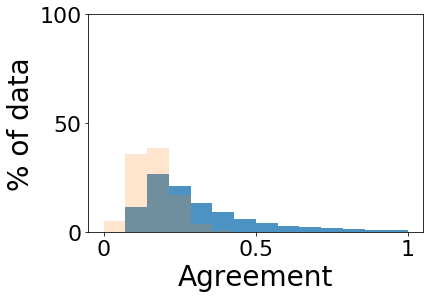}
\end{subfigure}
\begin{subfigure}{.158\textwidth}
  \centering
  \includegraphics[width=1\linewidth]{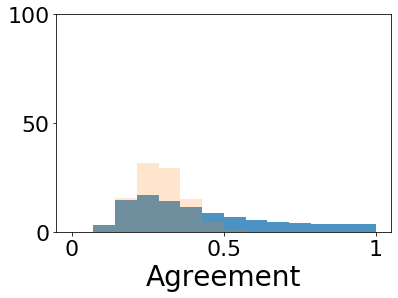}
\end{subfigure}
\begin{subfigure}{.158\textwidth}
  \centering
  \includegraphics[width=1\linewidth]{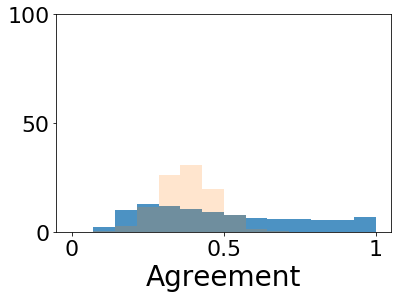}
\end{subfigure}
\begin{subfigure}{.158\textwidth}
  \centering
  \includegraphics[width=1\linewidth]{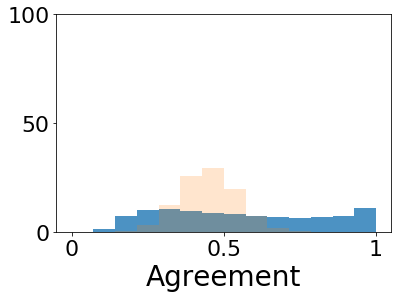}
\end{subfigure}
\begin{subfigure}{.158\textwidth}
  \centering
  \includegraphics[width=1\linewidth]{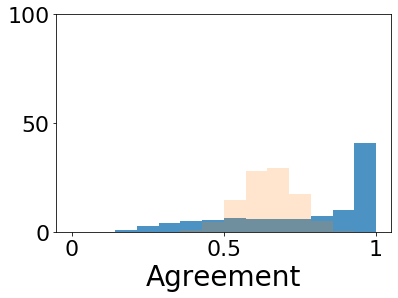}
\end{subfigure}
\begin{subfigure}{.158\textwidth}
  \centering
  \includegraphics[width=1\linewidth]{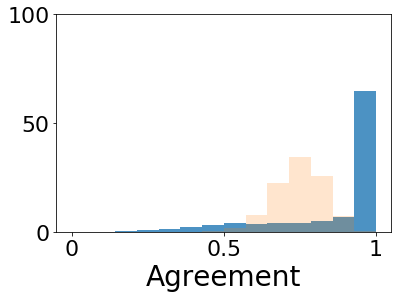}
\end{subfigure}

\caption{The distribution of Agreement scores of $27$ models of ResNet-50 models trained on ImageNet, in corresponding epochs as in Fig.~\ref{fig:tp-agreement-distribution-imagenet-per-epoch}. 
In each plot, the distribution of the null hypothesis is shown in fading orange, modeling the expected results for models that do not learn in a similar order (see text for details).
}
  \label{fig:agreement_testics}
\end{figure*}

Upon initialization, all models are effectively i.i.d random variables, and therefore the distribution of TP-agreement scores is approximately Gaussian around random chance. After a few epochs (in many cases a single epoch, see \app~\ref{app:additional_results}), the TP-agreement distribution changes dramatically, becoming bi-modal with peaks around 0 and 1. This abrupt distribution change is robust, and rather striking: from a state where most of the examples are being classified correctly by $\frac{1}{K}$ of the models, now most of the examples are being misclassified by all the models, while a small fraction is being correctly classified by all the models. When learning proceeds, the improvement in accuracy affects a shift of points from the peak at $0$ to the peak at $1$ while the distribution remains bi-modal, see Figs.~\ref{fig:consistency_dynamics}, \ref{fig:full_results_multiple_datasets} and \app~\ref{app:additional_results_robust}.

The data is learned in a specific order which is insensitive to the initialization and the sampling of the mini-batches. This is true for both the train and test sets. It indicates that the models capture similar functions in corresponding epochs: they classify correctly the same examples, and also consistently misclassify the same examples. Had the learning of the different models progressed independently, we would have seen a different dynamic. To rule the independence assumption out, we evaluate the null hypothesis - the independent progress of learning - by calculating the TP-agreement over a set of $N$ independent random classification vectors, with specific accuracy as a baseline. As seen in Fig.~\ref{fig:consistency_dynamics} (in fading orange), the distribution of TP-agreement in this case remains Gaussian in all epochs, where the Gaussian's mean slowly shifts while tracking the improved accuracy.

TP-agreement is not affected by the consensus among the misclassifications. To this end we have the Agreement score, which measures agreement regardless of whether the label is true or false: an agreement of 1 indicates that all models have classified the datapoint in the same way, regardless of whether it is correct or incorrect. Fig.~\ref{fig:agreement_testics} (blue) shows the distribution of Agreement scores for the cases depicted in Fig.~\ref{fig:consistency_dynamics}, showing that indeed all the models classify examples in almost the same way, even when they misclassify. Had the learning of the different models progressed independently, the dynamic would resemble a moving Gaussian, as can be seen in Fig.~\ref{fig:agreement_testics}, plotted in faded orange.

Similar results are seen when analyzing a classification problem that involves text, see Fig.~\ref{fig:nlp-task-tp-agreement}.
We applied a BiLSTM with attention using Glove \cite{Pennington2014GloveGV} over 39K training and 1K test questions from stack overflow \cite{stack}. Labels consist of 20 mutually exclusive programming language tags assigned by users.

\begin{figure}[htb]
\includegraphics[width=1\linewidth]{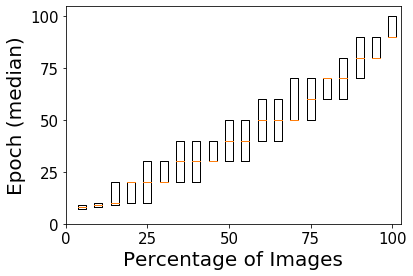}
\caption{Analysis of $20$ models of VGG-19 trained on CIFAR-100. Examples are sorted by the accessibility score (see \S\ref{sec:notations}), and each box represents $5\%$ of the examples with similar score. The orange bar represents the median epoch in which each set of images is learned, and each box represents the median's confidence intervals.}
  \label{fig:full_results_vggcifar100}
\end{figure}

The robust order of learning when training models from a single architecture allows us to measure for each example the epoch in which it is effectively learned. Specifically, for each example and each model we define the epoch in which an example is learned as the last epoch after which it is being correctly classified. In Fig.~\ref{fig:full_results_vggcifar100} we plot the median value (over all models) of this measure for small sets of examples, where the examples are sorted based on their accessibility score. We note the relatively low variance in learning epoch between different models in the collection. This result illustrates the robustness of the order in which examples are learned and the similarity between models from the same architecture. These results were reproduced for many datasets and architectures (see following section, and Fig.~\ref{fig:full_results_multiple_datasets}).

\subsection{Robustness}
\label{section:robusteness}

\begin{figure*}[htb]
 \begin{subfigure}{.193\textwidth}
   \centering
   \includegraphics[width=1\linewidth]{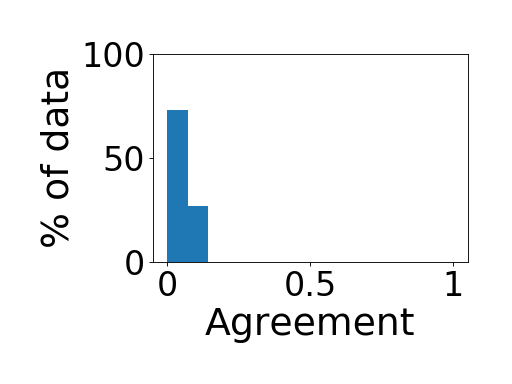}\vspace{-0.2cm}
  \caption{Random classification}
   \label{fig:random_network_out_of_sample_agreement}
 \end{subfigure}
 \begin{subfigure}{.193\textwidth}
   \centering
   \includegraphics[width=1\linewidth]{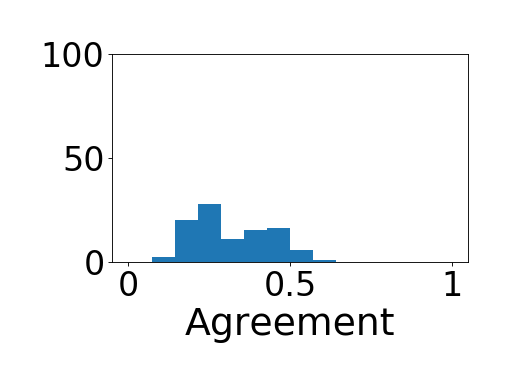}\vspace{-0.2cm}
  \caption{Noise images}
   \label{fig:resnet_out_of_sample_on_random}
 \end{subfigure}
 \begin{subfigure}{.193\textwidth}
   \centering
   \includegraphics[width=1\linewidth]{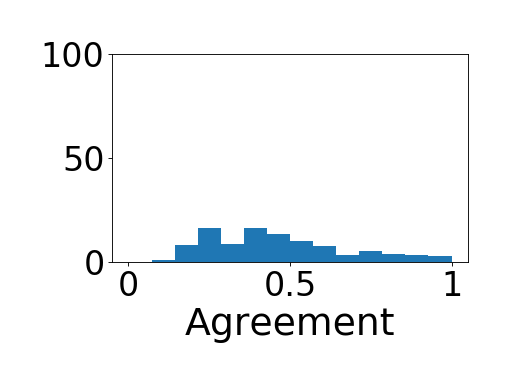}\vspace{-0.2cm}
  \caption{StyleGan images}
   \label{fig:resnet_out_of_sample_on_stylegan}
 \end{subfigure}
 \begin{subfigure}{.193\textwidth}
   \centering
   \includegraphics[width=1\linewidth]{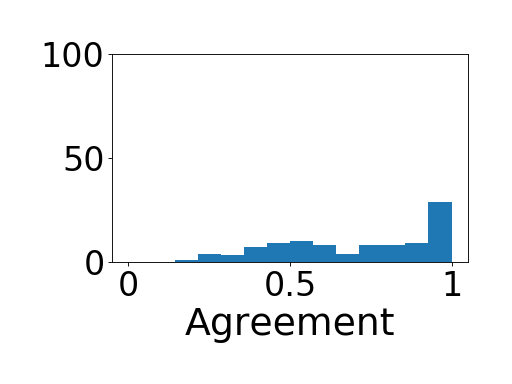}\vspace{-0.2cm}
  \caption{Different dataset}
   \label{fig:resnet_out_of_sample_on_indoor}
 \end{subfigure}
 \begin{subfigure}{.193\textwidth}
   \centering
   \includegraphics[width=1\linewidth]{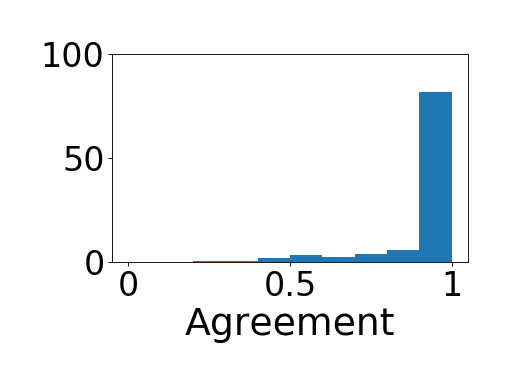}\vspace{-0.2cm}
  \caption{Test set}
   \label{fig:resnet_out_of_sample_on_in_sample}
 \end{subfigure}
\caption{The distribution of the Agreement score for $27$ models of ResNet-50 trained on ImageNet. a) The Agreement distribution of independent classifiers. (b-e) The agreement distribution when the 27 models are used to classify the following test datasets: (b) images generated by the random sampling of pixels from a normal distribution; (c) images generated by StyleGAN \cite{karras2019style} trained on ImageNet; (d) natural images from a different dataset -- Indoor Scene Recognition \cite{quattoni2009recognizing}; (e) the imageNet test set. Clearly, test datasets whose image statistics is more similar to the train data show a higher Agreement score.}
  \label{fig:out_of_sample}
\end{figure*}
The results reported above are extremely robust, seen in all the datasets and architectures that we have investigated, except for the synthetic datasets we artificially created specifically for this purpose (see \S\ref{sec:4-phases}). In addition to the results shown above on ImageNet \cite{deng2009imagenet}, CIFAR-100 \cite{krizhevsky2009learning} and the text classification task, similar results are obtained for a wide range of additional image datasets as shown in \app~\ref{app:additional_results_robust}, including: MNIST \cite{lecun1998gradient} -- Figs.~\ref{fig:mnist-tp-agreement},\ref{appendix:full_results_mnist}, Fashion-MNIST \cite{xiao2017fashion} -- Fig.~\ref{appendix:full_results_fashion_mnist}, CIFAR-10 \cite{krizhevsky2009learning} and CIFAR-100 -- Figs.~\ref{fig:cifar100-tp-agreement},\ref{fig:full_results_vggcifar100},\ref{appendix:full_results_vggcifar10},\ref{appendix:full_results_st-vgg-cifar10}, tiny ImageNet \cite{tinyImageNet} -- Fig.~\ref{appendix:full_results_tiny_imagenet}, ImageNet --  Figs.~\ref{fig:consistency_dynamics},\ref{appendix:full_results_imagenet_consistency_architectures},
VGGfaces2 \cite{Cao18} -- Fig.~\ref{appendix:full_results_vggfaces2}, and subsets of these datasets -- Figs.~\ref{fig:cats-and-dogs-tp-agreement},\ref{fig:small-mammals-tp-agreement},\ref{appendix:full_results_subset1},\ref{appendix:full_results_cats_dogs},\ref{appendix:full_results_split_fashion_mnist}. 

We investigated a variety of architectures, including AlexNet \cite{krizhevsky2012imagenet}, DenseNet \cite{huang2017densely} and ResNet-50 \cite{he2016deep} for ImageNet, VGG19 \cite{simonyan2014very} and a stripped version of VGG (denoted st-VGG) for CIFAR-10 and CIFAR-100, and several different handcrafted architectures for other data sets (see details in \app~\ref{app:hand_crafted_architecture}). The results can be replicated when changing the following hyper-parameters: learning rate, optimizer, batch size, dropout, weight decay, width, length and depth of layers, number of layers, kernel size, initialization, and activation functions. These hyper-parameters differ across the experiments detailed both in the main paper and in \app~\ref{app:additional_results_robust}.

\subsection{Different Training Sets}
\label{section:different_training_sets}

The observed pattern of similarity, and the order in which data is learned do not depend on the specific train set, rather on the distribution the train set is sampled from. To see this, we randomly split the train set into several partitions. We train a different collection of $N$ NNs on each of the random partitions, and compute the distribution of TP-agreement scores according to each partition for both the train set and the unmodified test set. Once again, all the NNs trained with the same partition show a bi-modal distribution of TP-agreement scores during the entire learning process, over their training partition and the common test set. More interestingly, the order in which these partitions learn the common test set is similar, depicting an almost perfect correlation between accessibility scores calculated according to different partitions. See \app~\ref{app:additional_results} and Fig.~\ref{appendix:full_results_split_fashion_mnist} for details.

\subsection{Out of Sample Test Sets}
\label{section:out_of_sample_test_sets}
Using the collection of ResNet-50 models whose analysis is shown in Figs.~\ref{fig:consistency_dynamics},\ref{fig:agreement_testics}, we further examined the agreement\footnote{Since the classes in the test sets are not present in the train, only the agreement remains relevant.\label{note1}} of the collection on out of sample test sets as shown in Fig.~\ref{fig:out_of_sample}. We see that the agreement is always higher than the agreement that would be achieved by a random assignment of labels. Interestingly, the more natural the images are and the more similar the distributions of the train and test images are, the higher the agreement is and the further away it gets from the agreement of random assignment of labels. This property may be used as a tool for novelty detection, and was left for future work.

\section{Cross Architectures Diversity}
\label{section:cross-consistency}

We now extend the analysis of the previous section to include NN instances of different architectures. In \S\ref{section:cross-consistency-other} we discuss comparisons with other learning paradigms. 

\subsection{Comparing Different Public Domain CNNs}
\label{section:compare_cross_architectures}

\begin{figure*}[ht]

\begin{subfigure}{.24\textwidth}
  \centering
  \includegraphics[width=1\linewidth]{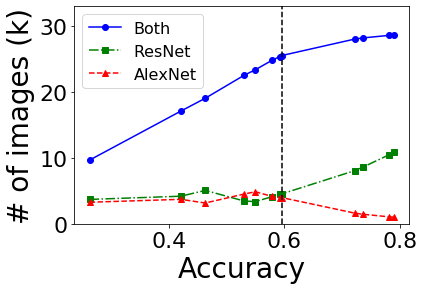}
  \caption{Correctly predicted images}
  \label{fig:order_learning_resnet_and_alexnet_common_classifications}
\end{subfigure}
\begin{subfigure}{.24\textwidth}
  \centering
  \includegraphics[width=1\linewidth]{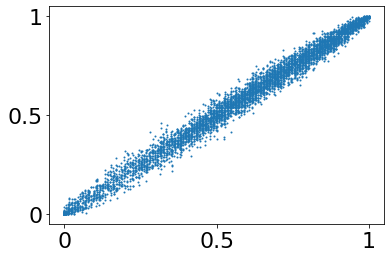}
  \caption{ResNet-ResNet}
  \label{fig:imagenet_order_learning_resnet_vs_resnet}
\end{subfigure}
\begin{subfigure}{.24\textwidth}
  \centering
  \includegraphics[width=1\linewidth]{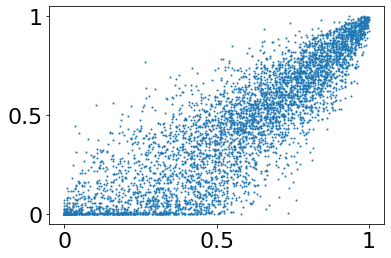}
  \caption{ResNet-AlexNet}
  \label{fig:imagenet_order_learning_alexnet_vs_resnet}
\end{subfigure}
\begin{subfigure}{.24\textwidth}
  \centering
  \includegraphics[width=1\linewidth]{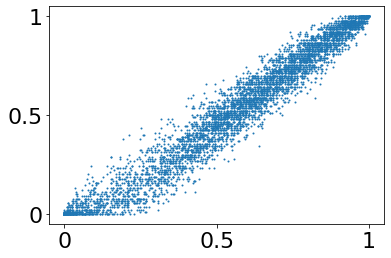}
  \caption{ResNet-DenseNet}
  \label{fig:imagenet_order_learning_resnet_vs_sensenet}
\end{subfigure}

\caption{Comparison of the learning order induced by models with different architectures -- $27$ models of ResNet-50, $22$ of AlexNet, and $6$ of DenseNet, all trained on ImageNet. a) The collections of ResNet and AlexNet are used to consturct a single ensemble classifier for ResNet-50 and AlexNet respectively, using the majority vote. We calculate the number of examples from the validation set which are classified correctly by either both ensembles (blue solid line), or only by a single ensemble (dotted green for ResNet and dashed red for AlexNet). Comparison is done during learning in corresponding epochs, where both ensembles reach the same accuracy. Since ResNet-50 reaches $80\%$ accuracy while AlexNet only reaches $60\%$ accuracy, there are no corresponding epochs beyond $60\%$, an event marked by a vertical dashed black line. Beyond this line, we compare ResNet-50 to the converged AlexNet. Before the convergence of AlexNet, both architectures learn the same examples during training. After convergence, ResNet first learns those examples classified correctly only by AlexNet, and then carries on to learn new stuff.
(b-d) Correlation between the \emph{accessibility score} of ImageNet train set when considering three pair of architectures. The $X$-axis corresonds to the \emph{accessibility score} of the first architecture in the pair, while the $Y$-axis corresponds to the second architecture in the pair. We considered the following pairs: (b) two disjoint collections of $13$ ResNet-50 models; (c) $27$ ResNet-50 models and $22$ AlexNet models; (d) $27$ ResNet-50 models and $6$ DenseNet models. Overall, all the pairs show high correlation between the \emph{accessibility scores} of the corresponding collections in the pair, see \S\ref{section:compare_cross_architectures}.}
  \label{fig:alexnet_vs_resnet}
\end{figure*}

\begin{figure*}[htb]
\begin{subfigure}{.73\textwidth}
  \includegraphics[width=0.19\linewidth]{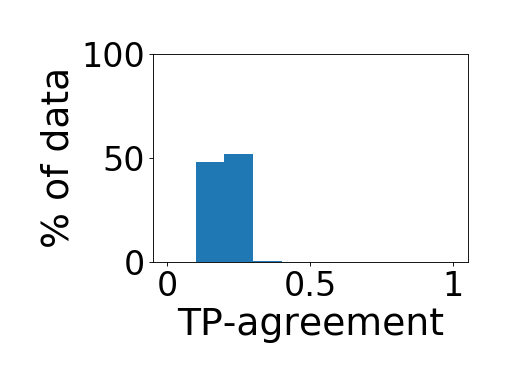}
  \includegraphics[width=0.19\linewidth]{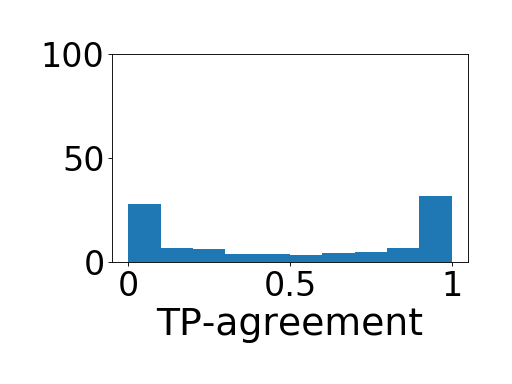}
  \includegraphics[width=0.19\linewidth]{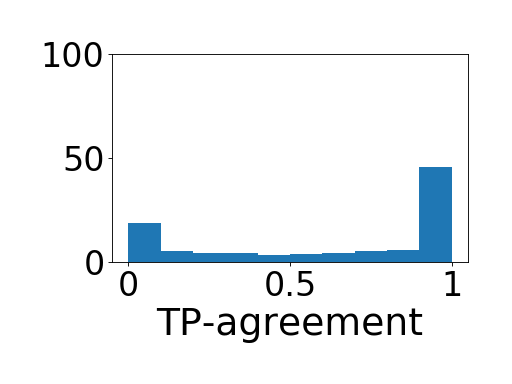}
  \includegraphics[width=0.19\linewidth]{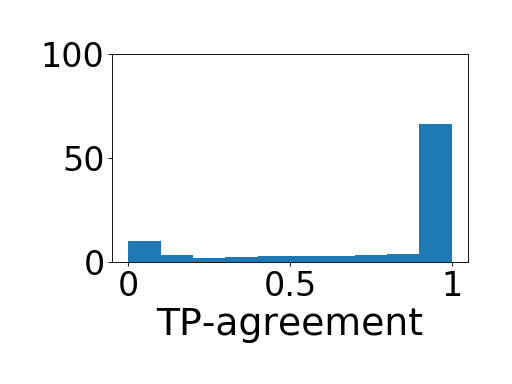}
  \includegraphics[width=0.19\linewidth]{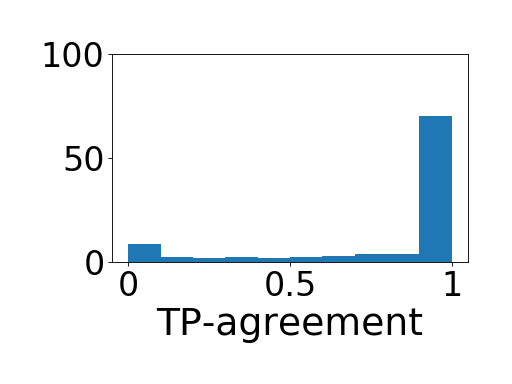}

  \includegraphics[width=0.19\linewidth]{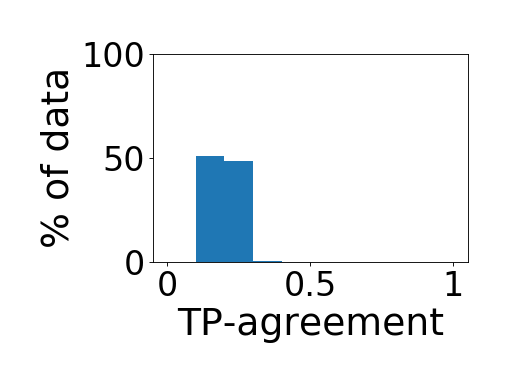}
  \includegraphics[width=0.19\linewidth]{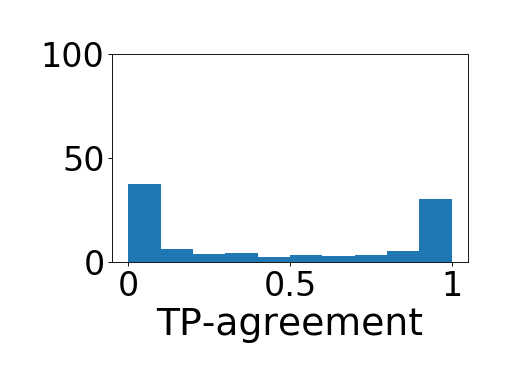}
  \includegraphics[width=0.19\linewidth]{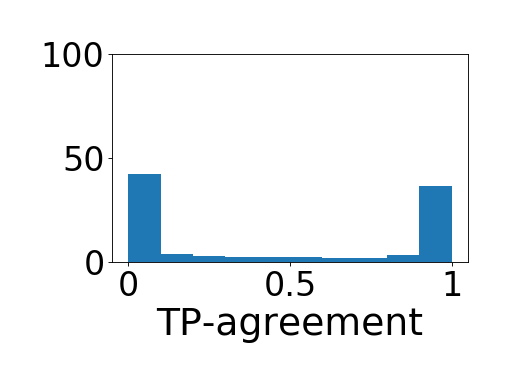}
  \includegraphics[width=0.19\linewidth]{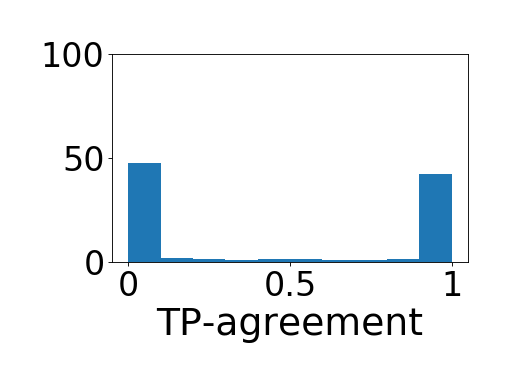}
  \includegraphics[width=0.19\linewidth]{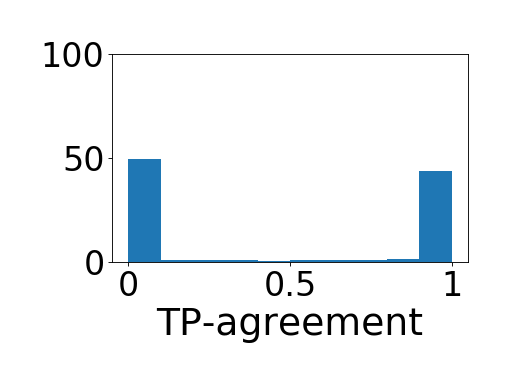}
  \caption{TP-agreement dynamics}
  \label{fig:consistency_linear_net}
\end{subfigure}
\begin{subfigure}{.26\textwidth}
\includegraphics[width=1\linewidth]{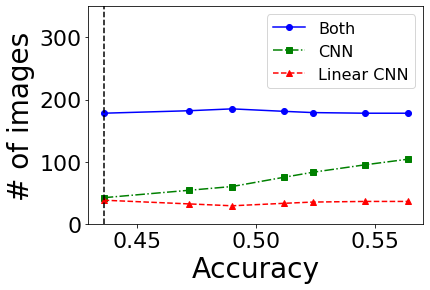}
\caption{performance comparison}
\label{fig:linear_net_comparison_bars}
\end{subfigure}
  \caption{Comparison of linear and non-linear networks trained on the small-mammals dataset. (a) 
  The distribution of TP-agreement scores at specific epochs, where epochs $0, 10, 30, 90, 140$ are shown respectively from left to right. Top -- st-VGG, bottom -- linear st-VGG. (b) Similar to Fig.~\ref{fig:order_learning_resnet_and_alexnet_common_classifications}. Solid blue -- the number of examples classified correctly by both architectures. Dotted green -- classified correctly only by st-VGG. Dashed red -- classified correctly only by linear st-VGG. Linear networks converge very fast, while non-linear networks learn the same points that are learned by linear networks, then carry on to learn more examples.}

  \label{fig:linear_net_comparison}
\end{figure*}

We start by directly extending the previous analysis to two collections generated by two different architectures. Each architecture is characterized by its own learning pace, therefore it makes little sense to compare TP-agreement epoch-wise. Instead, we compare collections in epochs with equivalent accuracy\footnote{Equivalence is determined up to a tolerance of $\pm1\%$; results are not sensitive to this value.}. Given specific accuracy, we compare the number of examples that are classified correctly by both architectures, to the number of examples which are classified correctly only by a single architecture. In Fig.~\ref{fig:order_learning_resnet_and_alexnet_common_classifications}, we compare between a collection of $27$ models of ResNet-50 and $22$ models of AlexNet trained on ImageNet. We see that as the accuracy improves, the number of examples that both models classify correctly increases, while the number of examples only a single model classifies remains constant and low. As ResNet-50 reaches higher final accuracy on ImageNet, we continue to compare these collections after AlexNet converged, comparing ResNet-50 to the final performance of AlexNet (all points after the dashed line in Fig.~\ref{fig:order_learning_resnet_and_alexnet_common_classifications}). We see that ResNet-50 first learns all the examples AlexNet does, then continues to learn new examples. 

The order in which different architectures learn the data is similar. To this end, we correlate the \emph{accessibility score} (see \S\ref{sec:notations}) of the two collections generated by the same architecture. When comparing two collections of ResNet-50, the correlation is almost $1$ ($r=0.99, ~p\leq10^{-50}$, Fig.~\ref{fig:imagenet_order_learning_resnet_vs_resnet}). The correlation remains high when comparing two collections of two different architectures: ResNet-50 and AlexNet ($r=0.87, ~p\leq10^{-50}$, Fig.~\ref{fig:imagenet_order_learning_alexnet_vs_resnet}) or ResNet-50 and DenseNet ($r=0.97, ~p\leq10^{-50}$, Fig.~\ref{fig:imagenet_order_learning_resnet_vs_sensenet}). These results are quite surprising given how the error rates of the three architectures differ: AlexNet Top-1 error: $0.45$, ResNet-50 Top-1 error: $0.24$, DenseNet Top-1 error: $0.27$. The results of comparing additional pairs of competitive ImageNet architectures are shown in \app~\ref{app:additional_results_robust},  Figs.~\ref{appendix:full_results_imagenet_dynamics},\ref{appendix:full_results_imagenet_correlations}. The results have been replicated for other settings, including: VGG19 and st-VGG on CIFAR-10 and CIFAR-100 (Fig.~\ref{appendix:full_results_correlations_comparison}).

\subsection{Linear Networks}
\label{section:linear_networks}

Convolutional Neural Networks (CNN) where the internal operations are limited to linear operators \cite{oja1992principal} define an important class of CNNs, as their linearity is often exploited in the theoretical investigation of deep learning. We report that the bi-modal behavior observed in general CNNs also occurs in this case.

\begin{figure*}[htb]
\begin{subfigure}{.33\textwidth}
  \centering
  \includegraphics[width=1\linewidth]{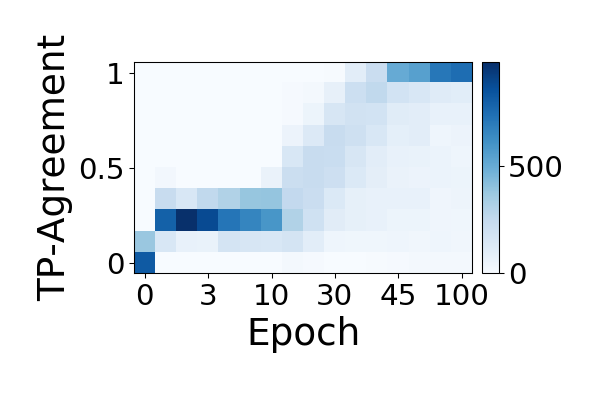}

  \includegraphics[width=1\linewidth]{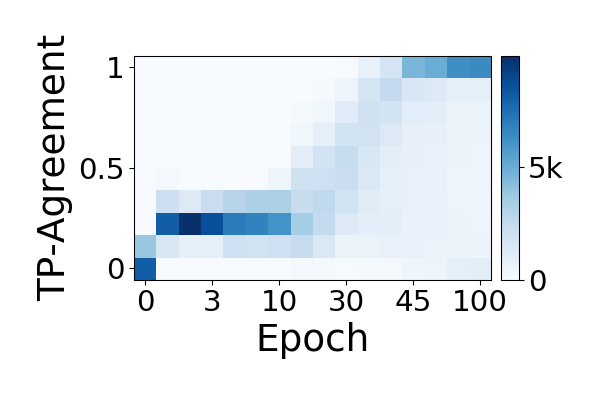}
  \caption{Gabor patches}
  \label{appendix:full_results_gabor12classes}
\end{subfigure}
\begin{subfigure}{.33\textwidth}
  \centering
  \includegraphics[width=1\linewidth]{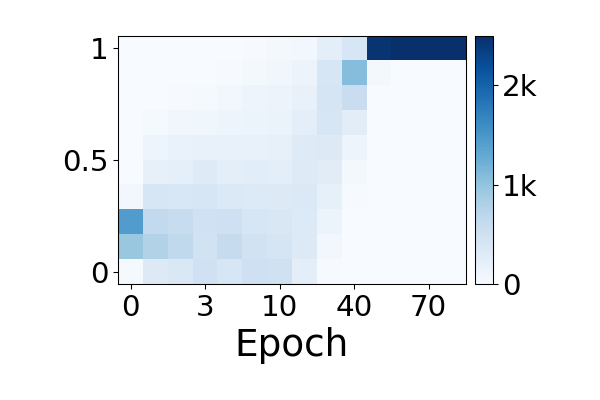}

  \includegraphics[width=1\linewidth]{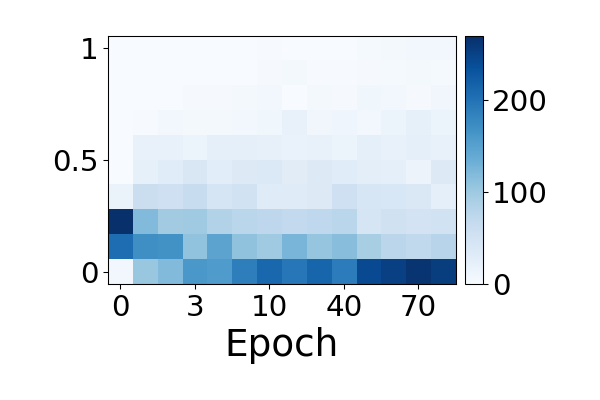}
  \caption{Small mammals -- Random labels}
  \label{fig:consistency_demonstration_random}
\end{subfigure}
\begin{subfigure}{.33\textwidth}
  \centering
  \includegraphics[width=1\linewidth]{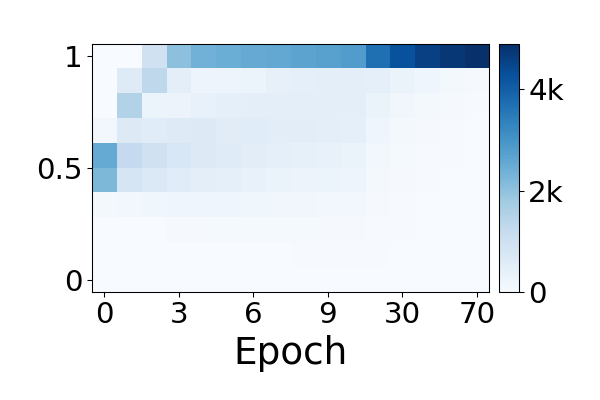}

  \includegraphics[width=1\linewidth]{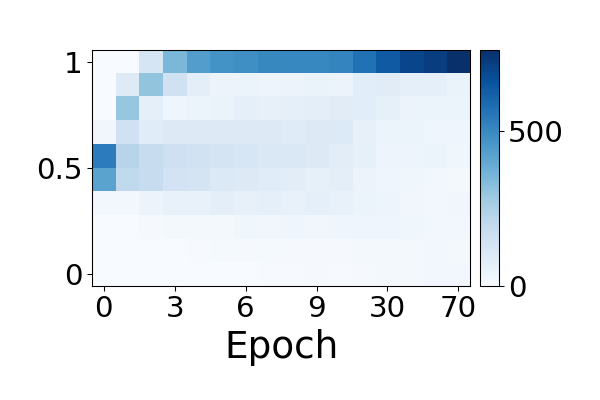}
  \caption{Gaussian classification}
  \label{fig:fully_connected_networks_consistency}
\end{subfigure}

\caption{Similarly to Fig.~\ref{fig:full_results_multiple_datasets}, we plot the distribution of TP-agreement scores during the entire learning process of both train (top) and test (bottom) datasets for several architectures and synthetic datasets. All the cases analyzed here show no common learning order, and models seem to learn the data independently of each other. The cases include: a) $100$ st-VGG models trained on the artificial Gabor patches dataset (see \S\ref{sec:gabor}); b) $100$ st-VGG models trained on the small mammals dataset with shuffled labels \citep{zhang2016understanding} (see \S\ref{sec:random_labels}); c) $100$ fully connected networks trained on a Gaussian classification task  (see \S\ref{sec:fully_connected_networks}).}
  \label{fig:non-bimodal-tp-agreement}
\end{figure*}

We define a linear st-VGG by replacing all the non-linear layers of st-VGG with their linear equivalents (see \app~\ref{app:architectures}). We train $100$ linear st-VGG on the small-mammals dataset (see \app~\ref{app:datasets}). The performance of these linear networks is weaker ($0.43$ average accuracy) than the original non-linear networks ($0.56$ average accuracy), and they converge faster. Still, the distribution of the TP-agreement throughout the entire learning process is bi-modal (maximum Pearson bi-modality: $0.06$), and this bi-modality is even more pronounced than the bi-modality in the non-linear case (maximum Pearson bi-modality: $0.22$). The bi-modal dynamics of st-VGG can be seen in the top row of Fig.~\ref{fig:consistency_linear_net}, compared to the dynamics of linear st-VGG in similar epochs at the bottom row of Fig.~\ref{fig:consistency_linear_net}.

Linear networks converge in just a few epochs, which is too fast for meaningful evaluation of accessibility score. Nevertheless, we still observe that non-linear networks learn first the examples linear networks do, then continue to learn new examples. In Fig.~\ref{fig:linear_net_comparison_bars}, we plot for each accuracy the number of images that were classified correctly by both linear and non-linear models, and the number of images that were learned by a single model only. In the beginning, the linear and non-linear variants learn roughly the same examples, while in more advanced epochs the non-linear networks continue to learn examples that remain hard for the linear networks.

These results show that the order of learning is not a direct result of using non-linear operations in neural architectures.

\section{When TP-Agreement is Not Bi-Modal}
\label{sec:4-phases}

In \S\ref{section:self-consistency} we discussed the characteristic bi-modal distribution of TP-agreement scores, illustrated in Figs.~\ref{fig:consistency_dynamics},\ref{fig:full_results_multiple_datasets}, which has appeared in all the experiments presented until now, in both the train and test sets. In this section, we investigate the circumstances under which the bi-modal distribution of TP-agreement is no longer seen.

\subsection{Synthetic Datasets}
\label{sec:fully_connected_networks}
\label{sec:gabor}
The bi-modal distribution of TP-agreement scores through all stages of learning is not an inherent property of NNs. We demonstrate this point using a dataset of artificial images, consisting of Gabor patches: the dataset contains 12 overlapping classes that differ from each other in orientation and color (see \app~\ref{app:gabor-dataset}). We trained a collection of $100$ st-VGG models on this data. The distribution of TP-agreement scores, shown in Fig.~\ref{appendix:full_results_gabor12classes}, is no longer bi-modal. Rather, the distribution is approximately normal. As learning proceeds, the mean of the distribution slowly shifts towards $1$, and the width of the distribution seems to expand. At convergence, the models have reached similar performance, and the bi-modal characteristics partially re-appears on the test data. These results suggest that networks in the collection have learned the data in different orders.

For some datasets, the order in which NNs learn may be completely independent. We train $100$ models of a fully connected architecture (see \app~\ref{app:architectures}). The NNs are trained to discriminate points sampled from two largely overlapping Gaussian distributions in high dimension. The dynamic distribution of TP-agreement scores is shown in Fig.~\ref{fig:fully_connected_networks_consistency}, and resembles the distribution of TP-agreement of the null hypothesis of random classification vectors. These results suggest that the order of learning in this case is independent across different models.

\subsection{Random Labels}
\label{sec:random_labels}

The bi-modality of TP-agreement seems to be associated with successful generalization. To see this, we take the small-mammals dataset, and reshuffle the labels such that every image is assigned a random label \cite{zhang2016understanding}. In this case, training accuracy reaches $100\%$ while test accuracy remains at chance level, which indicates that the NNs can memorize the data. Interestingly, the distribution of TP-agreement scores is no longer bi-modal, with a minimum Pearson bi-modality score of $1.07$ on train set and $1.35$ on the test set during the entire learning process. Rather, the distribution in each epoch resembles a Gaussian centered around the mean accuracy of the NNs, see Fig.~\ref{fig:consistency_demonstration_random}.
These results are in agreement with the results of \citet{arpit2017closer,morcos2018insights}, which show different NN dynamics when NNs perform memorization or generalization.

\section{Learning Order With AdaBoost Classifier}
\label{section:cross-consistency-other}

\begin{figure}[t!]
	\begin{subfigure}{.23\textwidth}
		\centering
		\includegraphics[width=1\linewidth]{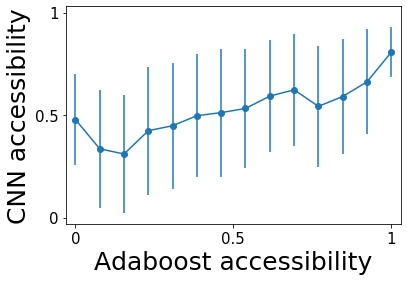}
	\end{subfigure}
	\begin{subfigure}{.23\textwidth}
		\centering
		\includegraphics[width=1\linewidth]{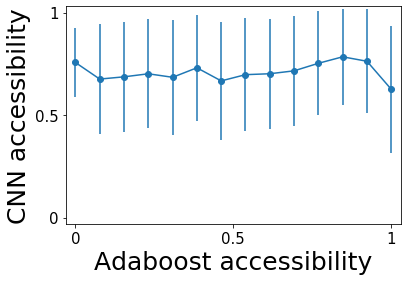}
	\end{subfigure}

	\caption{Correlations between the learning order of AdaBoost with $100$ linear classifiers and $100$ models of st-VGG both trained on CIFAR-10. Each value of AdaBoost accessibility (see \S\ref{section:cross-consistency-other}) is matched with the average over the corresponding st-VGG accessibility scores (see \S\ref{sec:notations}). Error bars show standard error. Left: AdaBoost trained on pixel values. Right: AdaBoost trained on the embedding obtained from the penultimate layer of Inception-V3 trained on ImageNet.}
	\label{fig:adaboost_cifar_10}
	\vspace{-.15in}
\end{figure}

Up to now, we investigated a variety of architectures, revealing a common learning order on benchmark datasets. This order may be fully determined by the benchmark dataset, in which case it should be replicated when inspecting other learning paradigms. In this section, we show that this is not the case. To this end, we consider boosting based on linear classifiers as weak learners, since the training of both neural models and AdaBoost share a dynamic aspect: in NN training accuracy increases with time due to the use of SGD, while in AdaBoost accuracy increases over time due to the accumulation of weak learners.

Thus we trained AdaBoost with up to $100$ linear classifiers on the CIFAR-10 dataset. Each channel in each image is normalized to $0$ mean and $1$ standard deviation (similarly to the normalization for NNs). The image tensor is then flattened to a vector. As can be seen in \app~\ref{app:other_learning_paradigms}, the AdaBoost accuracy is increasing as a function of the number of linear classifiers, and as expected, its final accuracy is significantly lower than neural models.

While most of the examples successfully learned by AdaBoost were also learned by the NN, the order in which the examples were learned was different. 
Similarly to Fig.~\ref{fig:alexnet_vs_resnet}, we compare the correlation between both learning orders (see Fig.~\ref{fig:adaboost_cifar_10}), showing a low correlation ($r=0.35,p\leq10^{-20}$) between the orders. This result demonstrates that non-neural paradigms may learn data in a different order.

Additionally, we trained AdaBoost using as features the penultimate layer of Inception-V3 \cite{szegedy2015going} trained on ImageNet. In this case, the AdaBoost accuracy increases dramatically (see \app~\ref{app:other_learning_paradigms}), while the correlation between the learning order gets even lower ($r=0.05,p\leq10^{-20}$), see Fig.~\ref{fig:adaboost_cifar_10}. This result shows that the learning order of NNs is not correlated with the learning order induced by AdaBoost, even when Adaboost uses a representation based on transfer learning, which shows that benchmark datasets can be learned in different ways. These results were replicated for other datasets, including subsets of CIFAR-100 and ImageNet, see \app~\ref{app:other_learning_paradigms}.

\section{Summary and Discussion}
\label{sec:discussion}

We empirically show that different neural models learn similar classification functions. We also show that the learning dynamics are similar, as they learn similar functions in all intermediate stages of learning. This is true for a variety of architecture, including different commonly used CNN architectures and LSTMs, trained on public domain datasets, and irrespective of size and other hyper-parameters.
This pattern of similarity crosses architectural boundaries: while different architectures may learn at a different speed, the data is learned in the same order.
Finally, we discuss cases where this similarity breaks down, indicating that the observed similarity is not an artifact of learning using SGD-based algorithms.

During this work, we see that some examples are being learned faster than others. However, we do not explain why are specific examples are "easier" than others, and leave this question for future work -- as this question is not trivial. The property that makes an example "easy" is not local, but context-based: we found datasets for which the same example can be easy for one and hard for the other.

Changing the sampling of train points during training, based on some notion of point difficulty, has been investigated in the context of curriculum learning \cite{bengio2009curriculum}, self-paced learning \cite{kumar2010self}, hard example mining \cite{shrivastava2016training}, and boosting \cite{hastie2009multi}. In contrast to this work, our measure of difficulty depends on what a network learns most robustly, as opposed to a measure determined by a teacher or the network's immediate accuracy. \citet{hacohen2019power} suggested a way to any ordering of the data as a base for a curriculum. Using this method with the order of learning depict above did not seem to benefit the learning.

\section*{Acknowledgements}
This work was supported in part by a grant from the Israel Science Foundation (ISF) and by the Gatsby Charitable Foundations. We thank Reviewer \#2 for proposing a more intuitive graphical representation of the results than we originally had, which feature prominently throughout the paper.

\bibliography{icml2020}

\begin{thebibliography}{48}
\providecommand{\natexlab}[1]{#1}
\providecommand{\url}[1]{\texttt{#1}}
\expandafter\ifx\csname urlstyle\endcsname\relax
  \providecommand{\doi}[1]{doi: #1}\else
  \providecommand{\doi}{doi: \begingroup \urlstyle{rm}\Url}\fi

\bibitem[Alain \& Bengio(2016)Alain and Bengio]{alain2016understanding}
Alain, G. and Bengio, Y.
\newblock Understanding intermediate layers using linear classifier probes.
\newblock \emph{arXiv preprint arXiv:1610.01644}, 2016.

\bibitem[Arpit et~al.(2017)Arpit, Jastrz{\k{e}}bski, Ballas, Krueger, Bengio,
  Kanwal, Maharaj, Fischer, Courville, Bengio, et~al.]{arpit2017closer}
Arpit, D., Jastrz{\k{e}}bski, S., Ballas, N., Krueger, D., Bengio, E., Kanwal,
  M.~S., Maharaj, T., Fischer, A., Courville, A., Bengio, Y., et~al.
\newblock A closer look at memorization in deep networks.
\newblock In \emph{Proceedings of the 34th International Conference on Machine
  Learning-Volume 70}, pp.\  233--242. JMLR. org, 2017.

\bibitem[Bengio et~al.(2009)Bengio, Louradour, Collobert, and
  Weston]{bengio2009curriculum}
Bengio, Y., Louradour, J., Collobert, R., and Weston, J.
\newblock Curriculum learning.
\newblock In \emph{Proceedings of the 26th annual international conference on
  machine learning}, pp.\  41--48. ACM, 2009.

\bibitem[Cao et~al.(2018)Cao, Shen, Xie, Parkhi, and Zisserman]{Cao18}
Cao, Q., Shen, L., Xie, W., Parkhi, O.~M., and Zisserman, A.
\newblock Vggface2: A dataset for recognising faces across pose and age.
\newblock In \emph{International Conference on Automatic Face and Gesture
  Recognition}, 2018.

\bibitem[Cohen et~al.(2019)Cohen, Chung, Lee, and
  Sompolinsky]{cohen2019separability}
Cohen, U., Chung, S., Lee, D.~D., and Sompolinsky, H.
\newblock Separability and geometry of object manifolds in deep neural
  networks.
\newblock \emph{bioRxiv}, pp.\  644658, 2019.

\bibitem[Cybenko(1989)]{cybenko1989approximation}
Cybenko, G.
\newblock Approximation by superpositions of a sigmoidal function.
\newblock \emph{Mathematics of control, signals and systems}, 2\penalty0
  (4):\penalty0 303--314, 1989.

\bibitem[Deng et~al.(2009)Deng, Dong, Socher, Li, Li, and
  Fei-Fei]{deng2009imagenet}
Deng, J., Dong, W., Socher, R., Li, L.-J., Li, K., and Fei-Fei, L.
\newblock Imagenet: A large-scale hierarchical image database.
\newblock In \emph{2009 IEEE conference on computer vision and pattern
  recognition}, pp.\  248--255. Ieee, 2009.

\bibitem[Erhan et~al.(2010)Erhan, Bengio, Courville, Manzagol, Vincent, and
  Bengio]{erhan2010does}
Erhan, D., Bengio, Y., Courville, A., Manzagol, P.-A., Vincent, P., and Bengio,
  S.
\newblock Why does unsupervised pre-training help deep learning?
\newblock \emph{Journal of Machine Learning Research}, 11\penalty0
  (Feb):\penalty0 625--660, 2010.

\bibitem[Glorot \& Bengio(2010)Glorot and Bengio]{glorot2010understanding}
Glorot, X. and Bengio, Y.
\newblock Understanding the difficulty of training deep feedforward neural
  networks.
\newblock In \emph{Proceedings of the thirteenth international conference on
  artificial intelligence and statistics}, pp.\  249--256, 2010.

\bibitem[Hacohen \& Weinshall(2019)Hacohen and Weinshall]{hacohen2019power}
Hacohen, G. and Weinshall, D.
\newblock On the power of curriculum learning in training deep networks.
\newblock \emph{arXiv preprint arXiv:1904.03626}, 2019.

\bibitem[Hastie et~al.(2009)Hastie, Rosset, Zhu, and Zou]{hastie2009multi}
Hastie, T., Rosset, S., Zhu, J., and Zou, H.
\newblock Multi-class adaboost.
\newblock \emph{Statistics and its Interface}, 2\penalty0 (3):\penalty0
  349--360, 2009.

\bibitem[He et~al.(2016)He, Zhang, Ren, and Sun]{he2016deep}
He, K., Zhang, X., Ren, S., and Sun, J.
\newblock Deep residual learning for image recognition.
\newblock In \emph{Proceedings of the IEEE conference on computer vision and
  pattern recognition}, pp.\  770--778, 2016.

\bibitem[Hornik et~al.(1989)Hornik, Stinchcombe, and
  White]{hornik1989multilayer}
Hornik, K., Stinchcombe, M., and White, H.
\newblock Multilayer feedforward networks are universal approximators.
\newblock \emph{Neural networks}, 2\penalty0 (5):\penalty0 359--366, 1989.

\bibitem[Huang et~al.(2017)Huang, Liu, Van Der~Maaten, and
  Weinberger]{huang2017densely}
Huang, G., Liu, Z., Van Der~Maaten, L., and Weinberger, K.~Q.
\newblock Densely connected convolutional networks.
\newblock In \emph{Proceedings of the IEEE conference on computer vision and
  pattern recognition}, pp.\  4700--4708, 2017.

\bibitem[Jiang et~al.(2017)Jiang, Zhou, Leung, Li, and
  Fei-Fei]{jiang2017mentornet}
Jiang, L., Zhou, Z., Leung, T., Li, L.-J., and Fei-Fei, L.
\newblock Mentornet: Learning data-driven curriculum for very deep neural
  networks on corrupted labels.
\newblock \emph{arXiv preprint arXiv:1712.05055}, 2017.

\bibitem[Junczys-Dowmunt et~al.(2018)Junczys-Dowmunt, Grundkiewicz, Dwojak,
  Hoang, Heafield, Neckermann, Seide, Germann, Aji, Bogoychev, Martins, and
  Birch]{JunczysDowmunt2018MarianFN}
Junczys-Dowmunt, M., Grundkiewicz, R., Dwojak, T., Hoang, H., Heafield, K.,
  Neckermann, T., Seide, F., Germann, U., Aji, A.~F., Bogoychev, N., Martins,
  A. F.~T., and Birch, A.
\newblock Marian: Fast neural machine translation in c++.
\newblock In \emph{ACL}, 2018.

\bibitem[Kantor et~al.(2019)Kantor, Katz, Choshen, Cohen-Karlik, Liberman,
  Toledo, Menczel, and Slonim]{Kantor2019LearningTC}
Kantor, Y., Katz, Y., Choshen, L., Cohen-Karlik, E., Liberman, N., Toledo, A.,
  Menczel, A., and Slonim, N.
\newblock Learning to combine grammatical error corrections.
\newblock In \emph{BEA@ACL}, 2019.

\bibitem[Karras et~al.(2019)Karras, Laine, and Aila]{karras2019style}
Karras, T., Laine, S., and Aila, T.
\newblock A style-based generator architecture for generative adversarial
  networks.
\newblock In \emph{Proceedings of the IEEE Conference on Computer Vision and
  Pattern Recognition}, pp.\  4401--4410, 2019.

\bibitem[Kawaguchi et~al.(2017)Kawaguchi, Kaelbling, and
  Bengio]{kawaguchi2017generalization}
Kawaguchi, K., Kaelbling, L.~P., and Bengio, Y.
\newblock Generalization in deep learning.
\newblock \emph{arXiv preprint arXiv:1710.05468}, 2017.

\bibitem[Krizhevsky \& Hinton(2009)Krizhevsky and
  Hinton]{krizhevsky2009learning}
Krizhevsky, A. and Hinton, G.
\newblock Learning multiple layers of features from tiny images.
\newblock Technical report, Citeseer, 2009.

\bibitem[Krizhevsky et~al.(2012)Krizhevsky, Sutskever, and
  Hinton]{krizhevsky2012imagenet}
Krizhevsky, A., Sutskever, I., and Hinton, G.~E.
\newblock Imagenet classification with deep convolutional neural networks.
\newblock In \emph{Advances in neural information processing systems}, pp.\
  1097--1105, 2012.

\bibitem[Krueger et~al.(2017)Krueger, Ballas, Jastrzebski, Arpit, Kanwal,
  Maharaj, Bengio, Fischer, and Courville]{krueger2017deep}
Krueger, D., Ballas, N., Jastrzebski, S., Arpit, D., Kanwal, M.~S., Maharaj,
  T., Bengio, E., Fischer, A., and Courville, A.
\newblock Deep nets don't learn via memorization.
\newblock 2017.

\bibitem[Kumar et~al.(2010)Kumar, Packer, and Koller]{kumar2010self}
Kumar, M.~P., Packer, B., and Koller, D.
\newblock Self-paced learning for latent variable models.
\newblock In \emph{Advances in Neural Information Processing Systems}, pp.\
  1189--1197, 2010.

\bibitem[Le \& Yang(2015)Le and Yang]{le2015tiny}
Le, Y. and Yang, X.
\newblock Tiny imagenet visual recognition challenge.
\newblock \emph{CS 231N}, 2015.

\bibitem[LeCun et~al.(1998)LeCun, Bottou, Bengio, Haffner,
  et~al.]{lecun1998gradient}
LeCun, Y., Bottou, L., Bengio, Y., Haffner, P., et~al.
\newblock Gradient-based learning applied to document recognition.
\newblock \emph{Proceedings of the IEEE}, 86\penalty0 (11):\penalty0
  2278--2324, 1998.

\bibitem[Lenc \& Vedaldi(2015)Lenc and Vedaldi]{lenc2015understanding}
Lenc, K. and Vedaldi, A.
\newblock Understanding image representations by measuring their equivariance
  and equivalence.
\newblock In \emph{Proceedings of the IEEE conference on computer vision and
  pattern recognition}, pp.\  991--999, 2015.

\bibitem[Li et~al.(2015)Li, Yosinski, Clune, Lipson, and
  Hopcroft]{li2015convergent}
Li, Y., Yosinski, J., Clune, J., Lipson, H., and Hopcroft, J.~E.
\newblock Convergent learning: Do different neural networks learn the same
  representations?
\newblock In \emph{FE@ NIPS}, pp.\  196--212, 2015.

\bibitem[Li et~al.(2016)Li, Yosinski, Clune, Lipson, and
  Hopcroft]{li2016convergent}
Li, Y., Yosinski, J., Clune, J., Lipson, H., and Hopcroft, J.~E.
\newblock Convergent learning: Do different neural networks learn the same
  representations?
\newblock In \emph{Iclr}, 2016.

\bibitem[Morcos et~al.(2018)Morcos, Raghu, and Bengio]{morcos2018insights}
Morcos, A., Raghu, M., and Bengio, S.
\newblock Insights on representational similarity in neural networks with
  canonical correlation.
\newblock In \emph{Advances in Neural Information Processing Systems}, pp.\
  5727--5736, 2018.

\bibitem[Oja(1992)]{oja1992principal}
Oja, E.
\newblock Principal components, minor components, and linear neural networks.
\newblock \emph{Neural networks}, 5\penalty0 (6):\penalty0 927--935, 1992.

\bibitem[Pearson(1894)]{pearson1894contributions}
Pearson, K.
\newblock Contributions to the mathematical theory of evolution.
\newblock \emph{Philosophical Transactions of the Royal Society of London. A},
  185:\penalty0 71--110, 1894.

\bibitem[Pennington et~al.(2014)Pennington, Socher, and
  Manning]{Pennington2014GloveGV}
Pennington, J., Socher, R., and Manning, C.~D.
\newblock Glove: Global vectors for word representation.
\newblock In \emph{EMNLP}, 2014.

\bibitem[Public domain dataset a()]{stack}
Public domain dataset a.
\newblock Stack overflow bigquery program language dataset.
\newblock Online:
  \url{https://storage.googleapis.com/tensorflow-workshop-examples/stack-overflow-data.csv},
  2019.
\newblock Accessed: 2019-09-24.

\bibitem[Public domain dataset b()]{tinyImageNet}
Public domain dataset b.
\newblock Tiny imagenet challenge.
\newblock Online: \url{https://tinyimagenet.herokuapp.com}, 2019.
\newblock Accessed: 2019-05-22.

\bibitem[Quattoni \& Torralba(2009)Quattoni and
  Torralba]{quattoni2009recognizing}
Quattoni, A. and Torralba, A.
\newblock Recognizing indoor scenes.
\newblock In \emph{2009 IEEE Conference on Computer Vision and Pattern
  Recognition}, pp.\  413--420. IEEE, 2009.

\bibitem[Raghu et~al.(2017)Raghu, Gilmer, Yosinski, and
  Sohl-Dickstein]{raghu2017svcca}
Raghu, M., Gilmer, J., Yosinski, J., and Sohl-Dickstein, J.
\newblock Svcca: Singular vector canonical correlation analysis for deep
  learning dynamics and interpretability.
\newblock In \emph{Advances in Neural Information Processing Systems}, pp.\
  6076--6085, 2017.

\bibitem[Rumelhart et~al.(1988)Rumelhart, Hinton, Williams,
  et~al.]{rumelhart1988learning}
Rumelhart, D.~E., Hinton, G.~E., Williams, R.~J., et~al.
\newblock Learning representations by back-propagating errors.
\newblock \emph{Cognitive modeling}, 5\penalty0 (3):\penalty0 1, 1988.

\bibitem[Saxe et~al.(2018)Saxe, McClelland, and Ganguli]{Saxe2018AMT}
Saxe, A.~M., McClelland, J.~L., and Ganguli, S.
\newblock A mathematical theory of semantic development in deep neural
  networks.
\newblock \emph{Proceedings of the National Academy of Sciences of the United
  States of America}, 116 23:\penalty0 11537--11546, 2018.

\bibitem[Schein \& Ungar(2007)Schein and Ungar]{schein2007active}
Schein, A.~I. and Ungar, L.~H.
\newblock Active learning for logistic regression: an evaluation.
\newblock \emph{Machine Learning}, 68\penalty0 (3):\penalty0 235--265, 2007.

\bibitem[Shrivastava et~al.(2016)Shrivastava, Gupta, and
  Girshick]{shrivastava2016training}
Shrivastava, A., Gupta, A., and Girshick, R.
\newblock Training region-based object detectors with online hard example
  mining.
\newblock In \emph{Proceedings of the IEEE conference on computer vision and
  pattern recognition}, pp.\  761--769, 2016.

\bibitem[Simonyan \& Zisserman(2014)Simonyan and Zisserman]{simonyan2014very}
Simonyan, K. and Zisserman, A.
\newblock Very deep convolutional networks for large-scale image recognition.
\newblock \emph{arXiv preprint arXiv:1409.1556}, 2014.

\bibitem[Szegedy et~al.(2015)Szegedy, Liu, Jia, Sermanet, Reed, Anguelov,
  Erhan, Vanhoucke, and Rabinovich]{szegedy2015going}
Szegedy, C., Liu, W., Jia, Y., Sermanet, P., Reed, S., Anguelov, D., Erhan, D.,
  Vanhoucke, V., and Rabinovich, A.
\newblock Going deeper with convolutions.
\newblock In \emph{Proceedings of the IEEE conference on computer vision and
  pattern recognition}, pp.\  1--9, 2015.

\bibitem[Vaswani et~al.(2017)Vaswani, Shazeer, Parmar, Uszkoreit, Jones, Gomez,
  Kaiser, and Polosukhin]{Vaswani2017AttentionIA}
Vaswani, A., Shazeer, N., Parmar, N., Uszkoreit, J., Jones, L., Gomez, A.~N.,
  Kaiser, L., and Polosukhin, I.
\newblock Attention is all you need.
\newblock In \emph{NIPS}, 2017.

\bibitem[Wang et~al.(2018)Wang, Hu, Gu, Hu, Wu, He, and
  Hopcroft]{wang2018towards}
Wang, L., Hu, L., Gu, J., Hu, Z., Wu, Y., He, K., and Hopcroft, J.
\newblock Towards understanding learning representations: To what extent do
  different neural networks learn the same representation.
\newblock In \emph{Advances in Neural Information Processing Systems}, pp.\
  9584--9593, 2018.

\bibitem[Xiao et~al.(2017)Xiao, Rasul, and Vollgraf]{xiao2017fashion}
Xiao, H., Rasul, K., and Vollgraf, R.
\newblock Fashion-mnist: a novel image dataset for benchmarking machine
  learning algorithms.
\newblock \emph{arXiv preprint arXiv:1708.07747}, 2017.

\bibitem[Yosinski et~al.(2015)Yosinski, Clune, Nguyen, Fuchs, and
  Lipson]{yosinski2015understanding}
Yosinski, J., Clune, J., Nguyen, A., Fuchs, T., and Lipson, H.
\newblock Understanding neural networks through deep visualization.
\newblock \emph{arXiv preprint arXiv:1506.06579}, 2015.

\bibitem[Zhang et~al.(2016)Zhang, Bengio, Hardt, Recht, and
  Vinyals]{zhang2016understanding}
Zhang, C., Bengio, S., Hardt, M., Recht, B., and Vinyals, O.
\newblock Understanding deep learning requires rethinking generalization.
\newblock \emph{arXiv preprint arXiv:1611.03530}, 2016.

\bibitem[Zhang et~al.(2018)Zhang, Isola, Efros, Shechtman, and
  Wang]{Zhang_2018_CVPR}
Zhang, R., Isola, P., Efros, A.~A., Shechtman, E., and Wang, O.
\newblock The unreasonable effectiveness of deep features as a perceptual
  metric.
\newblock In \emph{The IEEE Conference on Computer Vision and Pattern
  Recognition (CVPR)}, June 2018.

\end{thebibliography}
\bibliographystyle{icml2020}

\pagenumbering{arabic}
\setcounter{page}{1}

\appendix
\part*{Supplementary}

\section{Related work}
\label{app:related-work}
How deep neural networks generalize is an open problem \cite{kawaguchi2017generalization}. The expressiveness of NNs is broad \cite{cybenko1989approximation}, and they can learn any arbitrary complex function \cite{hornik1989multilayer}. This extended capacity can indeed be reached, and neural networks can memorize datasets with randomly assigned labels \cite{zhang2016understanding}. Nevertheless, the dominant hypothesis today is that in natural datasets they "prefer" to learn an easier hypothesis that fits the data rather than memorize it all \cite{zhang2016understanding, arpit2017closer}. Our work is consistent with a hypothesis which requires fewer assumptions, see Section~\ref{sec:discussion}.

The direct comparison of neural representations is regarded to be a hard problem, due to a large number of parameters and the many underlying symmetries. Many non-direct approaches are available in the literature: \citet{li2016convergent, wang2018towards} compare subsets of similar features across multiple networks, which span similar low dimensional spaces, and show that while single neurons can vary drastically, some features are reliably learned across networks. \citet{raghu2017svcca} proposed the SVCCA method, which can compare layers and networks efficiently, with an amalgamation of SVD and CCA. They showed that multiple instances of the same converged network are similar to each other and that networks converge in a bottom-up way, from earlier layers to deeper ones. \citet{morcos2018insights} builds off the results of \citet{raghu2017svcca}, further showing that networks which generalize are more similar than ones which memorize, and that similarity grows with the width of the network. Other works has suggested various aspects of NN similarities \cite{Zhang_2018_CVPR,Saxe2018AMT}.

In various machine learning methods such as curriculum learning \cite{bengio2009curriculum}, self-paced learning \cite{kumar2010self} and active learning \cite{schein2007active}, examples are presented to the learner in a specific order \cite{hacohen2019power, jiang2017mentornet}. 
Although conceptually similar, here we analyze the order in which examples are learned, while the aforementioned methods seek ways to alter it. Likewise, the design of effective initialization methods is a striving research area \cite{erhan2010does,glorot2010understanding,rumelhart1988learning}. Here we do not seek to improve these methods, but rather analyze the properties of a collection of network instances generated by the same initialization methodology.


\section{Architectures}
\label{app:architectures}
\label{app:hand_crafted_architecture}
In addition to the public domain architectures described in \S\ref{section:robusteness}, we also experimented with some handcrafted networks. Such networks are simpler and faster to train, and are typically used to investigate the learning of less commonly used datasets, such as the small-mammals dataset and tiny ImageNet. Below we list all the architectures used in this paper.

\paragraph{st-VGG.}
A stripped version of VGG which we used in many experiments. It is a convolutional neural network, containing 8 convolutional layers with 32, 32, 64, 64, 128, 128, 256, 256 filters respectively. The first 6 layers have filters of size $3\times3$, and the last 2 layers have filters of size $2\times2$. Every second layer there is followed by $2\times2$ max-pooling layer and a $0.25$ dropout layer. After the convolutional layers, the units are flattened, and there is a fully-connected layer with 512 units followed by $0.5$ dropout.  When training with random labels, we removed both dropout layers to enable proper training, as suggested in \citet{krueger2017deep}. The batch size we used was $100$. The output layer is a fully connected layer with output units matching the number of classes in the dataset, followed by a softmax layer. We trained the network using the SGD optimizer, with cross-entropy loss. When training st-VGG, we used a learning rate of $0.05$ which decayed by a factor of $1.8$ every 20 epochs. 

\paragraph{Linear st-VGG.}
A linear version of st-VGG presented above. In linear st-VGG, we change the activation function to the id function, and replace all the max-pooling with average pooling with similar stride. We used the same hyper-parameters to train models of this architecture, although a wide range of hyper-parameters will reach similar results.

\paragraph{Small st-VGG.}
To compare st-VGG with another architecture, we created a smaller version of it: we used another convolutional neural network, containing 4 convolutional layers with 32, 32, 64, 64 filters respectively, with filters of size $3\times3$. Every second layer there is followed by $2\times2$ max-pooling and a $0.25$ dropout layer. After the convolutional layers, the units are flattened, and there is a fully-connected layer with 128 units followed by $0.5$ dropout. The output layer is a fully connected layer with output units matching the number of classes in the dataset, followed by a softmax layer. We trained the network using the SGD optimizer, with cross-entropy loss. We trained this network with the same learning rate and batch size as st-VGG.

\paragraph{MNIST architecture.}
When experimenting with the MNIST dataset, we used some arbitrary small architecture for simplicity, as most architectures are able to reach over $0.99$ accuracy. The architecture we used had 2 convolutional layers, with 32 and 64 filters respectively of size $3\times 3$. After the convolutions, we used $2\times 2$ max-pooling, followed by $0.25$ dropout. Finally, we used a fully connected layer of size $128$ followed by $0.5$ dropout and Softmax. We used a learning rate of $1$ for $12$ epochs, using AdaDelta optimizer and a batch size of $100$.

\paragraph{Fully connected architecture.}
When experimenting with fully connected networks, we used a 4 layers network, which simply flattened the data, followed by 2 fully connected layers with $1024$ units, followed byn an output layer with softmax. We used $0.5$ dropout after each fully connected layer. Since these networks converge fast, a wide range of learning rates can be used. Specifically, we used $0.04$. We experimented with a wide range of numbers of fully connected layers, reaching similar results.

\paragraph{BiLSTM with Attention.} When experimenting on textual data we used a GloVe embeddings, a layer of BiLSTM of size $300$, $0.25$ dropout and recurrent dropout, an attention layer, a fully connected layer of size $256$ with $0.25$ dropout and a last fully connected layer to extract output. The networks were optimized using Adam optimization with a learning rate of $0.005$ and a batch size of $256$. 

\section{Datasets}
\label{app:datasets}

\paragraph{Small Mammals.}
The small-mammals dataset used in the paper is the relevant super-class of the CIFAR-100 dataset. It contains $2500$ train images divided into 5 classes equally, and $500$ test images. Each image is of size $32\times 32\times 3$. This dataset was chosen due to its small size, which allowed for efficient experimentation. All the results observed in this dataset were reproduced on large, public domain datasets, such as CIFAR-100, CIFAR-10, and ImageNet.

\paragraph{Insect.}
Similarly to the small mammals dataset, the relevant super-class of CIFAR-100.

\paragraph{Fish.}
Similarly to the small mammals dataset, the relevant super-class of CIFAR-100.

\paragraph{Cats and Dogs.}
The cats and dogs dataset is a subset of CIFAR-10. It uses only the 2 relevant classes, to create a binary problem. Each image is of size $32\times 32\times 3$. The dataset is divided to $20000$ train images ($10000$ per class) and $2000$ test images ($1000$ per class).

\begin{figure}[hbt]
a)
  \includegraphics[width=1\linewidth]{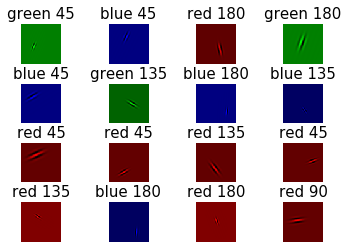}
b)
  \includegraphics[width=1\linewidth]{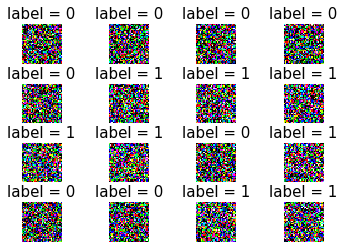}
\caption{Visualization of synthetic datasets used thoughout the paper. a) Gabor. b) Gaussian.}
  \label{appendix:datasets_visualization}
\end{figure}

\paragraph{Gabor.}
\label{app:gabor-dataset}
The Gabor dataset used in the paper, is a dataset we created which contains 12 classes of Gabor patches. Each class contains 100 images of Gabor patches which vary in size and orientation. Classes differ from each other in 2 parameters: 1) RGB channel -- each class depicts the Gabor patch in a single RGB channel. 2) Orientation -- each class can have one of the following base orientations: $45^{\circ}, 90^{\circ}, 135^{\circ}, 180^{\circ}$. The orientation of each class varies by $\pm 30^{\circ}$, making some of the classes non-separable, while some classes are. The images within each class vary from each other in the size of the Gabor patch, and its spatial location across the images. See attached code for generating the dataset. See Fig.~\ref{appendix:datasets_visualization} for visualization.

\paragraph{Gaussian.}
\label{app:gaussian_dataset}
The Gaussian dataset used in the fully connected case, is a 2-classes dataset. One class is sampled from a multivariate Gaussian with mean $0$ and $\Sigma=I$, while the other class is sampled from a multivariate Gaussian with mean $0.1$ and $\Sigma=I$. Other choices for the mean and variance yield similar results. Each sampled vector was of dimension $3072$, and then reshaped to $32\times 32\times 3$ to resemble the shape of CIFAR images. Each class contained $2500$ train images and $500$ test images. See attached code for generating the dataset. See Fig.~\ref{appendix:datasets_visualization} for visualization.

\paragraph{VGGFace2 subset.}
\label{app:vgg_face_dataset}
We created a classification task for face recognition, using a subset of $10$ classes from VGGFace2. We chose the classes containing the largest number of images. We chose $600$ images from each class arbitrarily to be the train set, while the remaining points (between $89$ and $243$) served as the test set. Each image was resized to $64\times 64\times 3$, using center cropping while maintaining aspect ratio. 

\paragraph{Stack Overflow.} The data from Stack Overflow is publicly shared and used for tutorials. It contains 39K training samples and 1K test samples, each tagged with one of 20 programming languages as the language the question asks about. Each question must be regarded more as a paragraph than a sentence. Many words, terms and symbols are expected to be domain-dependent, and therefore under-represented in the embeddings.

\paragraph{ImageNet cats.}
This dataset is a subset of ImageNet dataset ILSVRC 2012. We used 7 classes of cats, which obtained all the hyponyms of the cat synset that appeared in the data, following the work of \citet{hacohen2019power}. The labels of the classes included wre 'Egyptian cat', 'Persian cat', 'Cougar, puma, catamount, mountaion lion, painter, panther, Felis concolor', 'Tiger cat', 'Siamese cat, Siamese', 'Tabby, tabby cat', 'Lynx, catamount'. Images were resized to $56\times 56$ for faster preformance. Data was normalized to 0 mean and standard variation of 1 in each channel separately.

\section{Robustness of results}
\label{app:additional_results_robust}

Similar qualitative results were obtained in all the experiments with natural datasets. To maintain a fair comparison across epochs, the results for each shown epoch $e$ (effectively epoch extent) were obtained by independently training a different set of $N$ networks from scratch for $e$ epochs. The specific set of epochs $\mathcal{S}_E$, where $\vert\mathcal{S}_E\vert=7$, that was used in each plot was determined arbitrarily, to evenly span all sections of learning. 
All the networks in all test cases converged before the final epoch plotted. 

\begin{figure*}[hbt]
a)
\begin{subfigure}{.135\textwidth}
  \centering
  \includegraphics[width=1\linewidth]{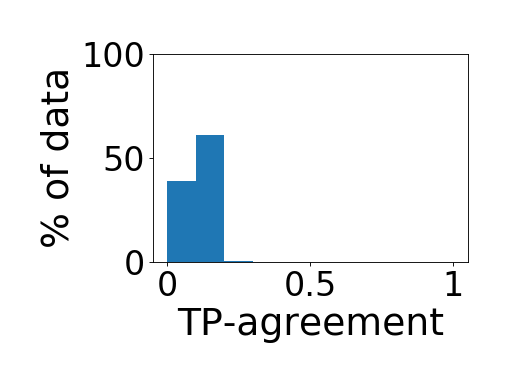}
\end{subfigure}
\begin{subfigure}{.135\textwidth}
  \centering
  \includegraphics[width=1\linewidth]{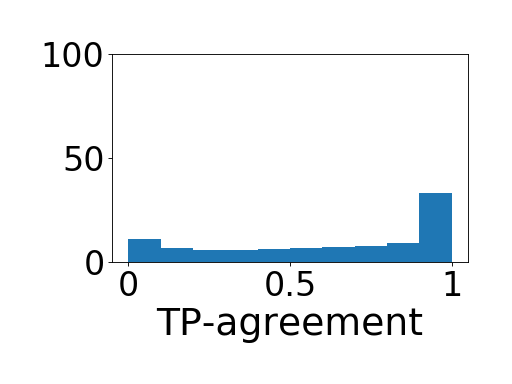}
\end{subfigure}
\begin{subfigure}{.135\textwidth}
  \centering
  \includegraphics[width=1\linewidth]{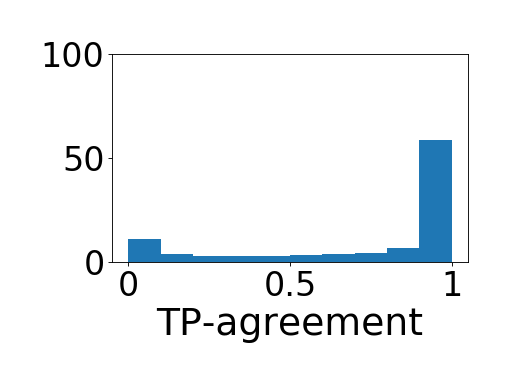}
\end{subfigure}
\begin{subfigure}{.135\textwidth}
  \centering
  \includegraphics[width=1\linewidth]{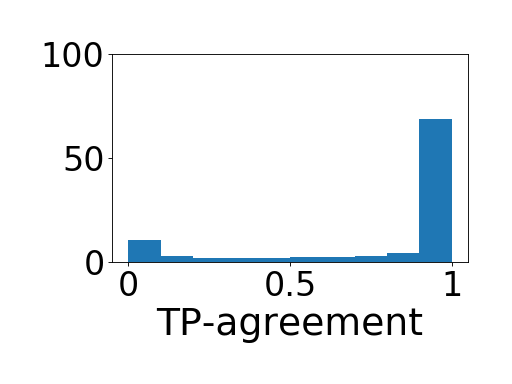}
\end{subfigure}
\begin{subfigure}{.135\textwidth}
  \centering
  \includegraphics[width=1\linewidth]{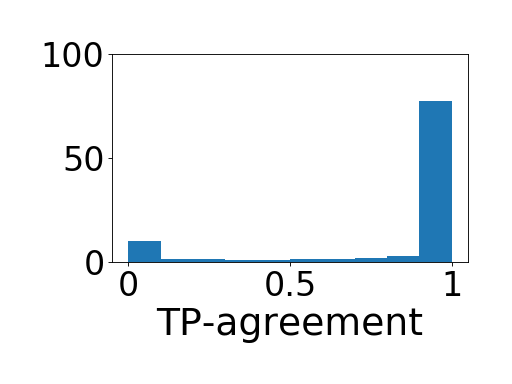}
\end{subfigure}
\begin{subfigure}{.135\textwidth}
  \centering
  \includegraphics[width=1\linewidth]{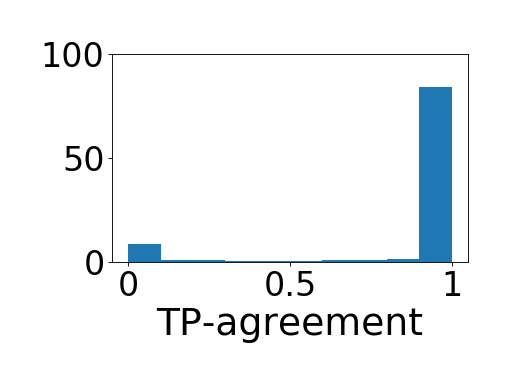}
\end{subfigure}
\begin{subfigure}{.135\textwidth}
  \centering
  \includegraphics[width=1\linewidth]{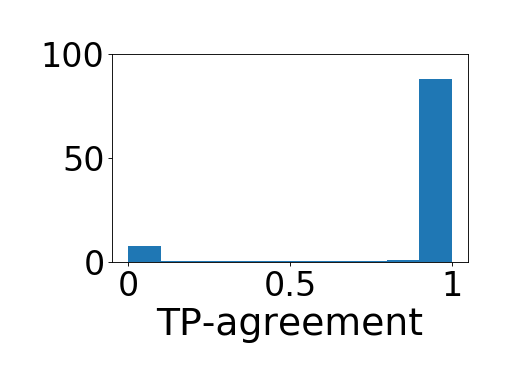}
\end{subfigure}

b)
\begin{subfigure}{.135\textwidth}
  \centering
  \includegraphics[width=1\linewidth]{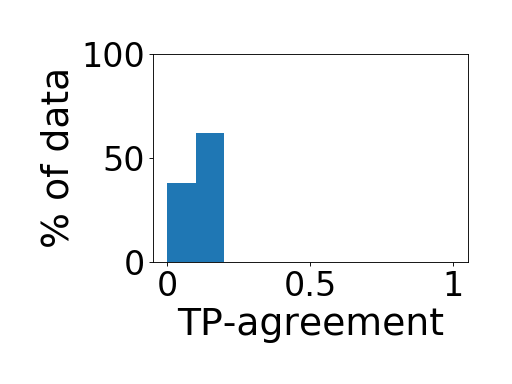}
\end{subfigure}
\begin{subfigure}{.135\textwidth}
  \centering
  \includegraphics[width=1\linewidth]{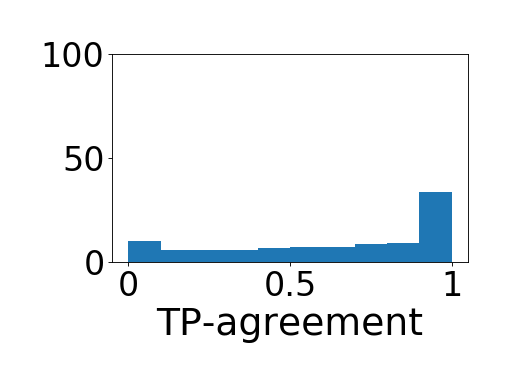}
\end{subfigure}
\begin{subfigure}{.135\textwidth}
  \centering
  \includegraphics[width=1\linewidth]{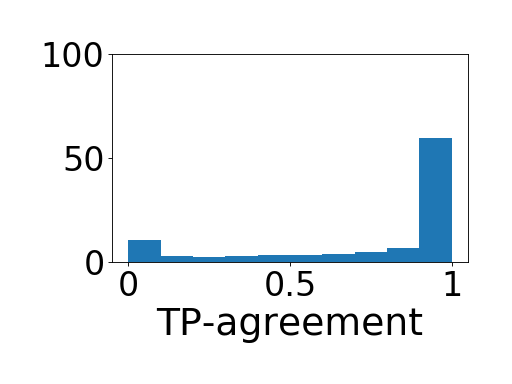}
\end{subfigure}
\begin{subfigure}{.135\textwidth}
  \centering
  \includegraphics[width=1\linewidth]{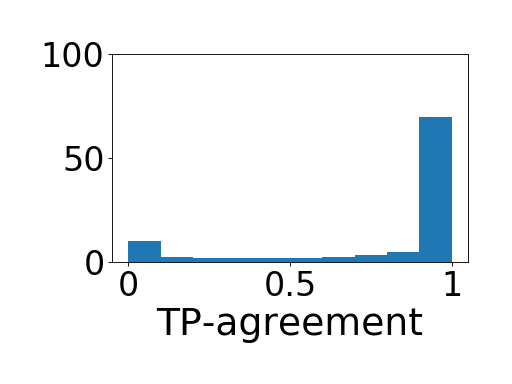}
\end{subfigure}
\begin{subfigure}{.135\textwidth}
  \centering
  \includegraphics[width=1\linewidth]{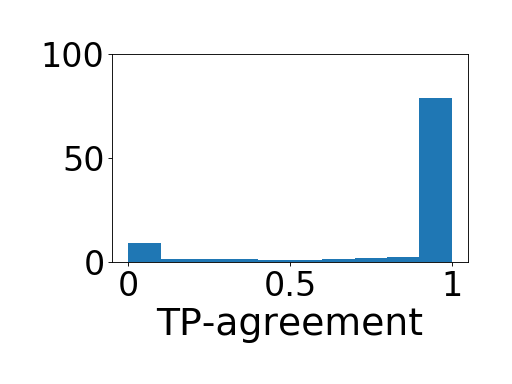}
\end{subfigure}
\begin{subfigure}{.135\textwidth}
  \centering
  \includegraphics[width=1\linewidth]{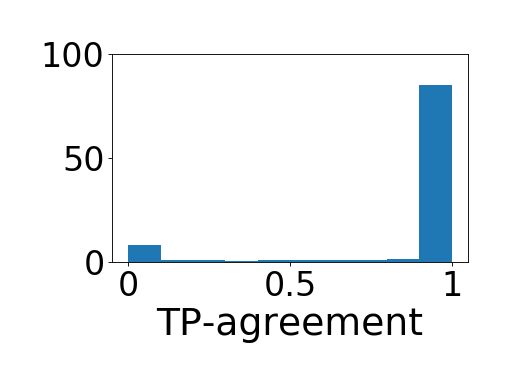}
\end{subfigure}
\begin{subfigure}{.135\textwidth}
  \centering
  \includegraphics[width=1\linewidth]{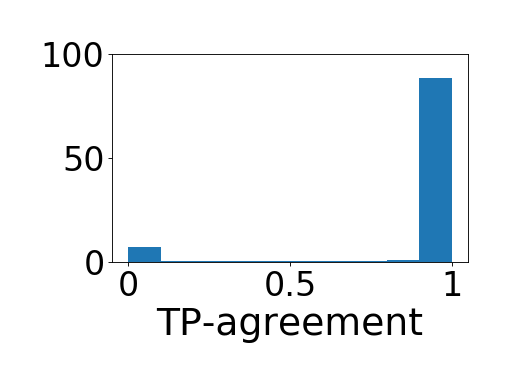}
\end{subfigure}
\caption{The distribution of TP-agreement scores during the learning process of $100$ instances of small architecture (see \app~\ref{app:architectures}) trained on MNIST. Epochs shown: $0, 1, 2, 3, 5, 10, 20$. We used a low learning rate ($0.001$) to avoid convergence after one epoch. a) Train set; b) Test set.}
  \label{appendix:full_results_mnist}
\end{figure*}

\begin{figure*}[hbt]
a)
\begin{subfigure}{.135\textwidth}
  \centering
  \includegraphics[width=1\linewidth]{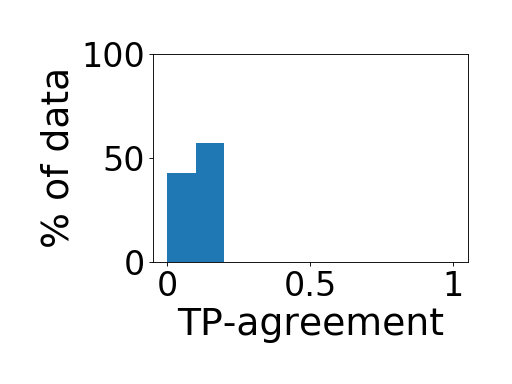}
\end{subfigure}
\begin{subfigure}{.135\textwidth}
  \centering
  \includegraphics[width=1\linewidth]{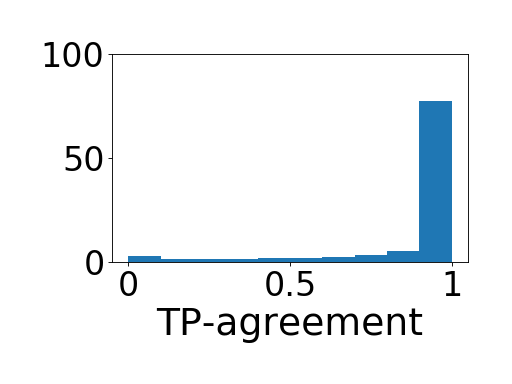}
\end{subfigure}
\begin{subfigure}{.135\textwidth}
  \centering
  \includegraphics[width=1\linewidth]{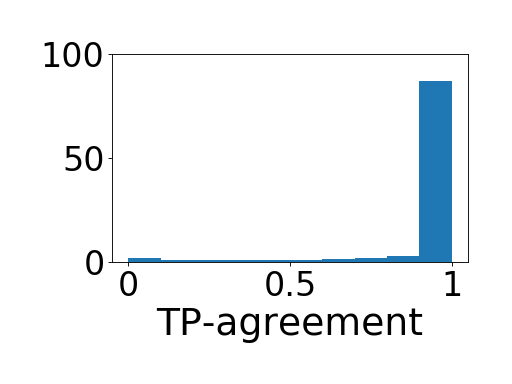}
\end{subfigure}
\begin{subfigure}{.135\textwidth}
  \centering
  \includegraphics[width=1\linewidth]{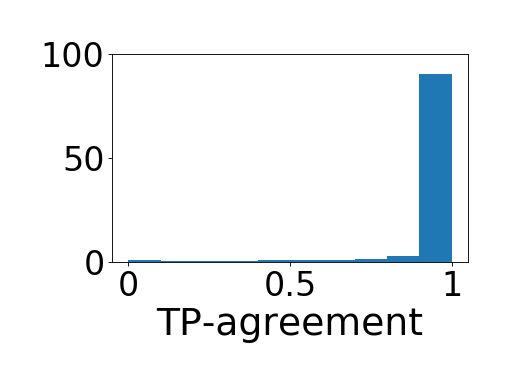}
\end{subfigure}
\begin{subfigure}{.135\textwidth}
  \centering
  \includegraphics[width=1\linewidth]{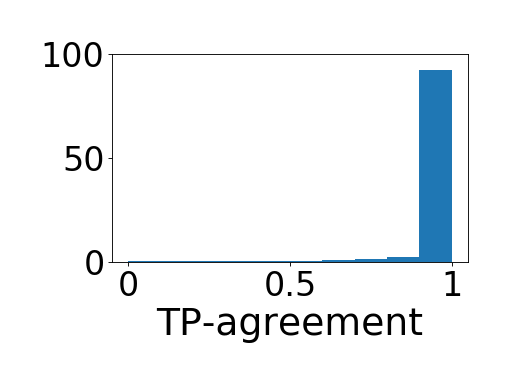}
\end{subfigure}
\begin{subfigure}{.135\textwidth}
  \centering
  \includegraphics[width=1\linewidth]{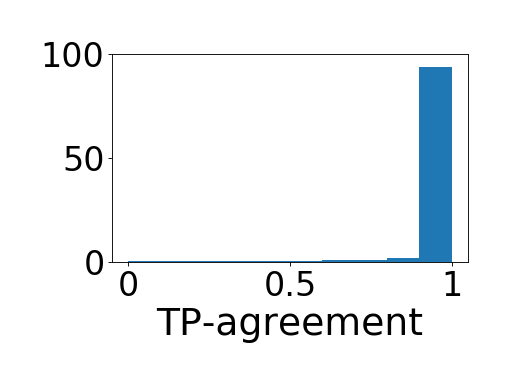}
\end{subfigure}
\begin{subfigure}{.135\textwidth}
  \centering
  \includegraphics[width=1\linewidth]{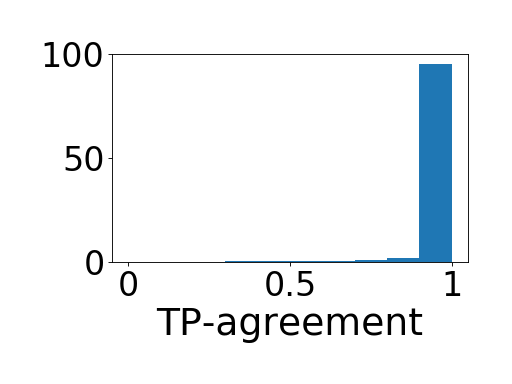}
\end{subfigure}

b)
\begin{subfigure}{.135\textwidth}
  \centering
  \includegraphics[width=1\linewidth]{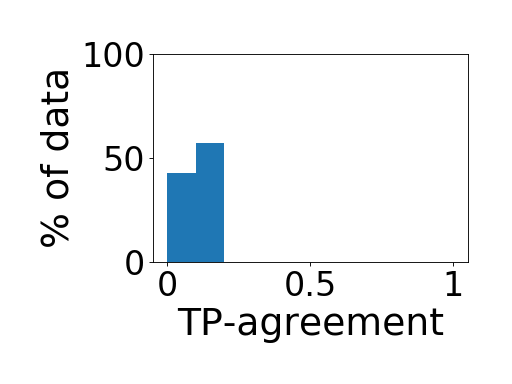}
\end{subfigure}
\begin{subfigure}{.135\textwidth}
  \centering
  \includegraphics[width=1\linewidth]{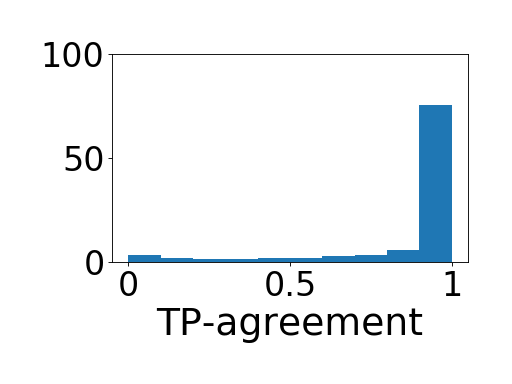}
\end{subfigure}
\begin{subfigure}{.135\textwidth}
  \centering
  \includegraphics[width=1\linewidth]{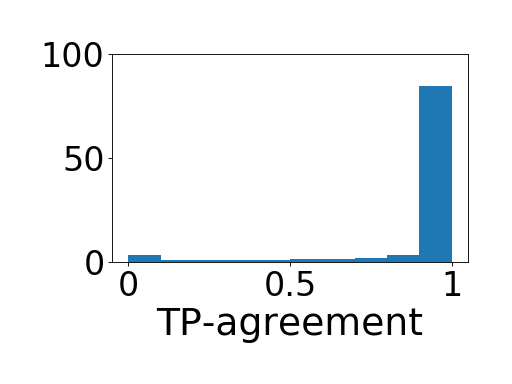}
\end{subfigure}
\begin{subfigure}{.135\textwidth}
  \centering
  \includegraphics[width=1\linewidth]{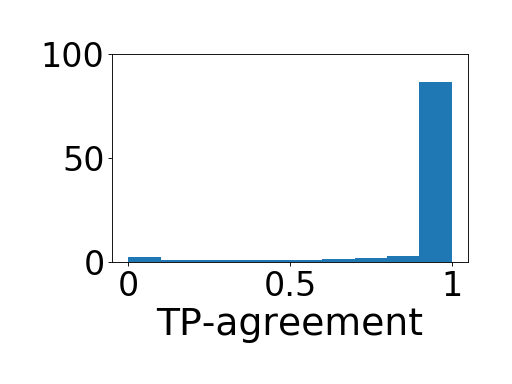}
\end{subfigure}
\begin{subfigure}{.135\textwidth}
  \centering
  \includegraphics[width=1\linewidth]{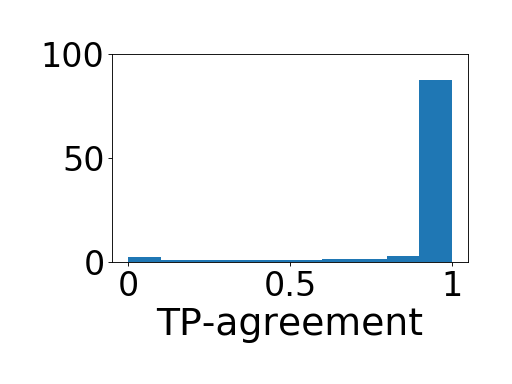}
\end{subfigure}
\begin{subfigure}{.135\textwidth}
  \centering
  \includegraphics[width=1\linewidth]{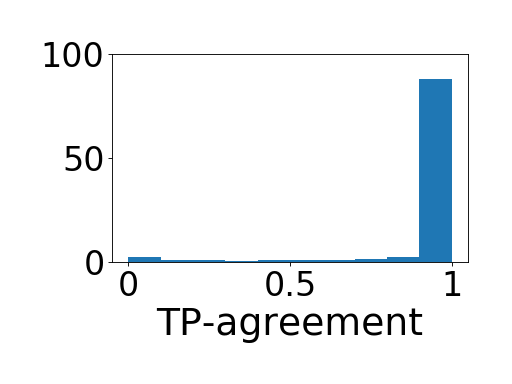}
\end{subfigure}
\begin{subfigure}{.135\textwidth}
  \centering
  \includegraphics[width=1\linewidth]{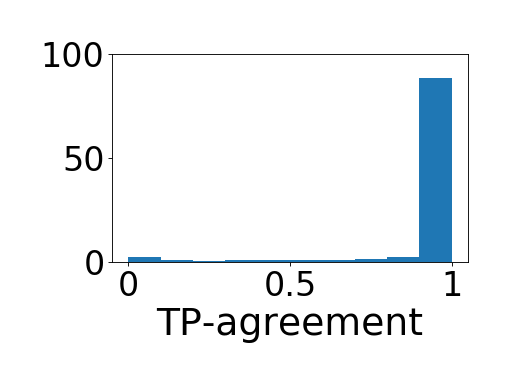}
\end{subfigure}
\caption{The distribution of TP-agreement scores during the learning process of $100$ instances of st-VGG (see \app~\ref{app:architectures}) trained on Fashion-MNIST. Epochs shown: $0, 1, 5, 10, 15, 20, 25$. a) Train set; b) Test set.}
  \label{appendix:full_results_fashion_mnist}
\end{figure*}

\begin{figure*}[hbt]
a)
\begin{subfigure}{.135\textwidth}
  \centering
  \includegraphics[width=1\linewidth]{st-vgg-subset16/histogram_consistency_train_epoch0.png}
\end{subfigure}
\begin{subfigure}{.135\textwidth}
  \centering
  \includegraphics[width=1\linewidth]{st-vgg-subset16/histogram_consistency_train_epoch10.png}
\end{subfigure}
\begin{subfigure}{.135\textwidth}
  \centering
  \includegraphics[width=1\linewidth]{st-vgg-subset16/histogram_consistency_train_epoch30.png}
\end{subfigure}
\begin{subfigure}{.135\textwidth}
  \centering
  \includegraphics[width=1\linewidth]{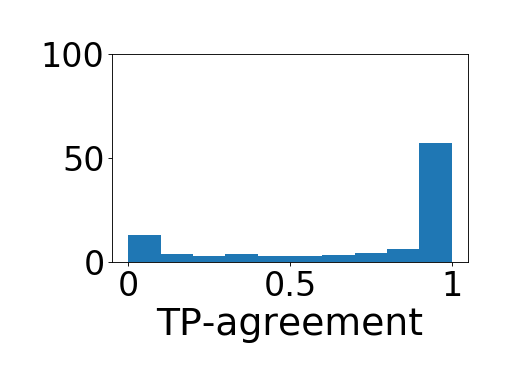}
\end{subfigure}
\begin{subfigure}{.135\textwidth}
  \centering
  \includegraphics[width=1\linewidth]{st-vgg-subset16/histogram_consistency_train_epoch90.png}
\end{subfigure}
\begin{subfigure}{.135\textwidth}
  \centering
  \includegraphics[width=1\linewidth]{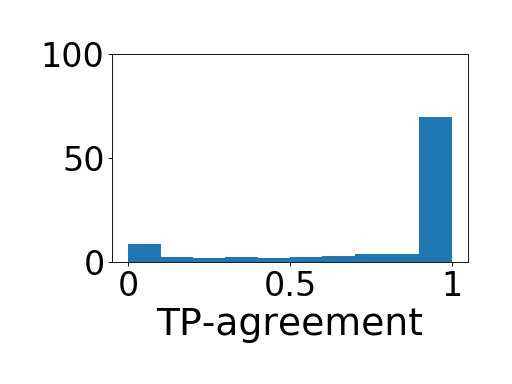}
\end{subfigure}
\begin{subfigure}{.135\textwidth}
  \centering
  \includegraphics[width=1\linewidth]{st-vgg-subset16/histogram_consistency_train_epoch140.png}
\end{subfigure}

b)
\begin{subfigure}{.135\textwidth}
  \centering
  \includegraphics[width=1\linewidth]{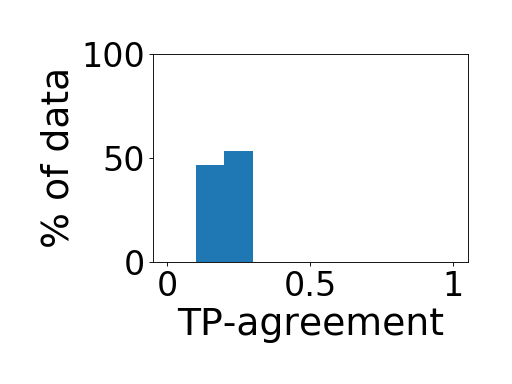}
\end{subfigure}
\begin{subfigure}{.135\textwidth}
  \centering
  \includegraphics[width=1\linewidth]{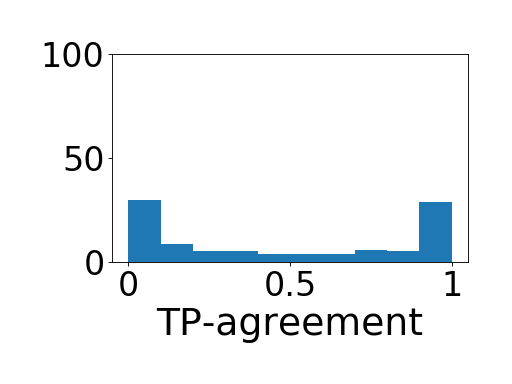}
\end{subfigure}
\begin{subfigure}{.135\textwidth}
  \centering
  \includegraphics[width=1\linewidth]{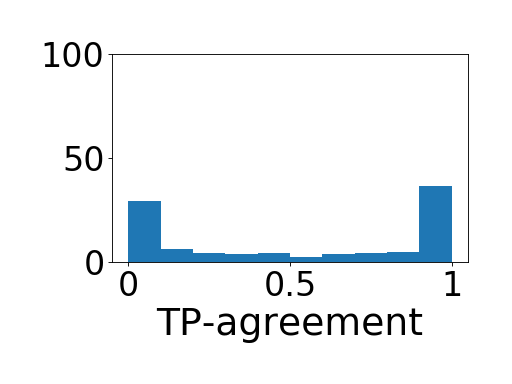}
\end{subfigure}
\begin{subfigure}{.135\textwidth}
  \centering
  \includegraphics[width=1\linewidth]{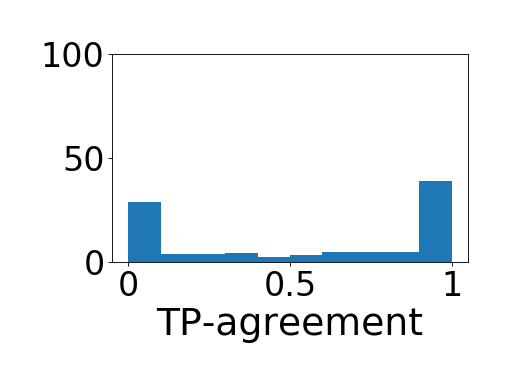}
\end{subfigure}
\begin{subfigure}{.135\textwidth}
  \centering
  \includegraphics[width=1\linewidth]{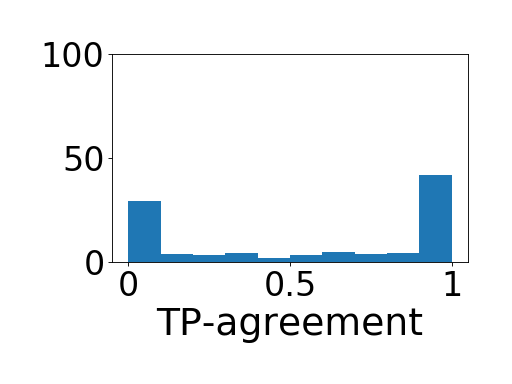}
\end{subfigure}
\begin{subfigure}{.135\textwidth}
  \centering
  \includegraphics[width=1\linewidth]{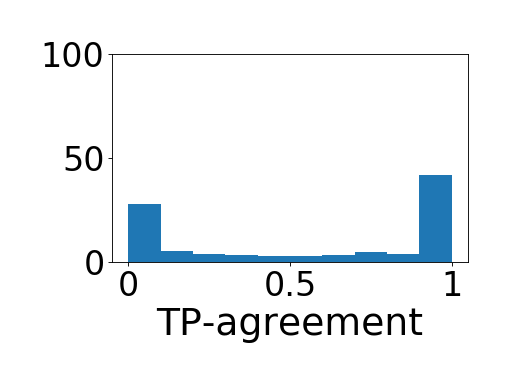}
\end{subfigure}
\begin{subfigure}{.135\textwidth}
  \centering
  \includegraphics[width=1\linewidth]{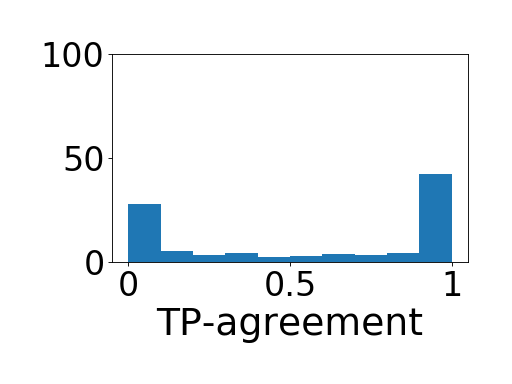}
\end{subfigure}

c)
\begin{subfigure}{.135\textwidth}
  \centering
  \includegraphics[width=1\linewidth]{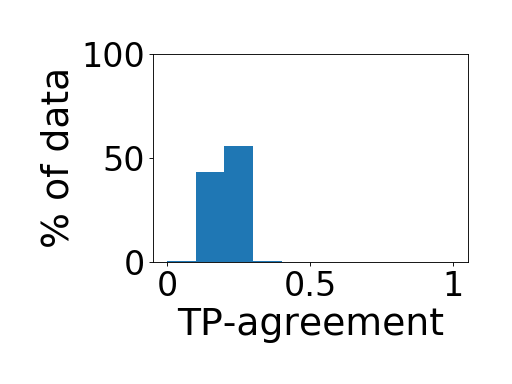}
\end{subfigure}
\begin{subfigure}{.135\textwidth}
  \centering
  \includegraphics[width=1\linewidth]{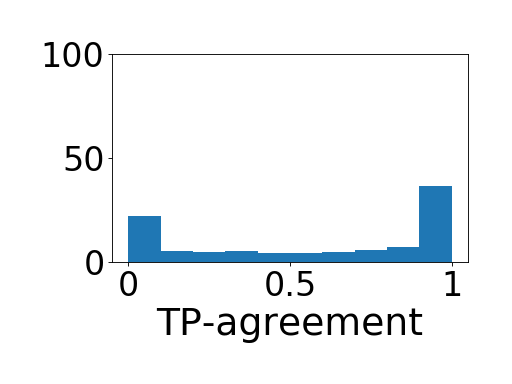}
\end{subfigure}
\begin{subfigure}{.135\textwidth}
  \centering
  \includegraphics[width=1\linewidth]{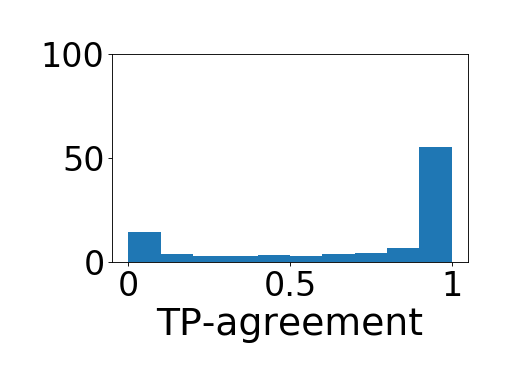}
\end{subfigure}
\begin{subfigure}{.135\textwidth}
  \centering
  \includegraphics[width=1\linewidth]{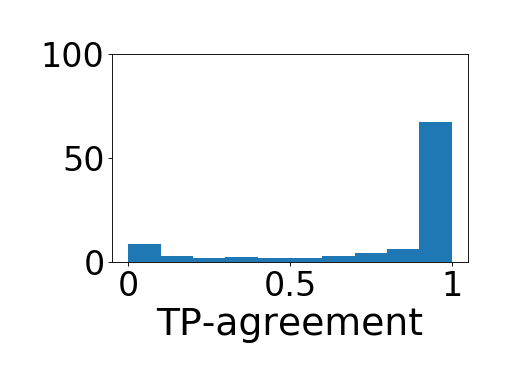}
\end{subfigure}
\begin{subfigure}{.135\textwidth}
  \centering
  \includegraphics[width=1\linewidth]{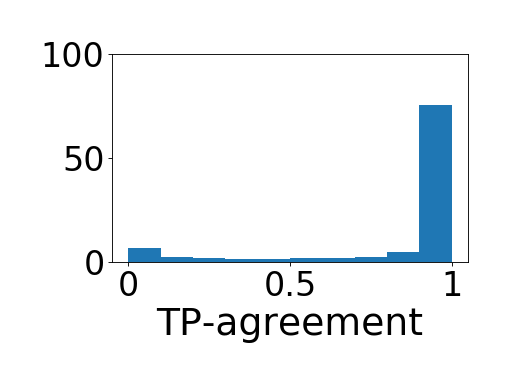}
\end{subfigure}
\begin{subfigure}{.135\textwidth}
  \centering
  \includegraphics[width=1\linewidth]{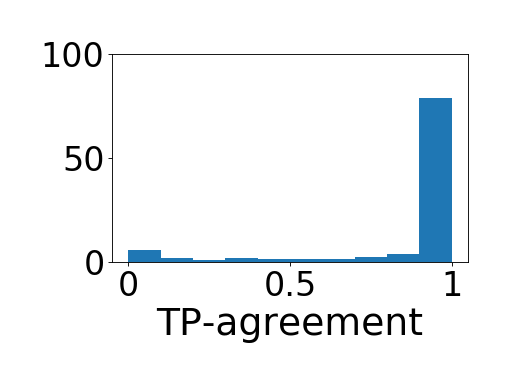}
\end{subfigure}
\begin{subfigure}{.135\textwidth}
  \centering
  \includegraphics[width=1\linewidth]{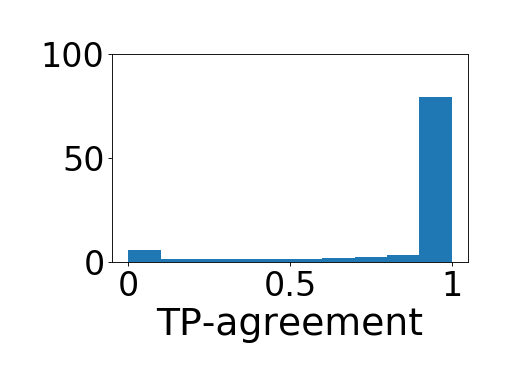}
\end{subfigure}

d)
\begin{subfigure}{.135\textwidth}
  \centering
  \includegraphics[width=1\linewidth]{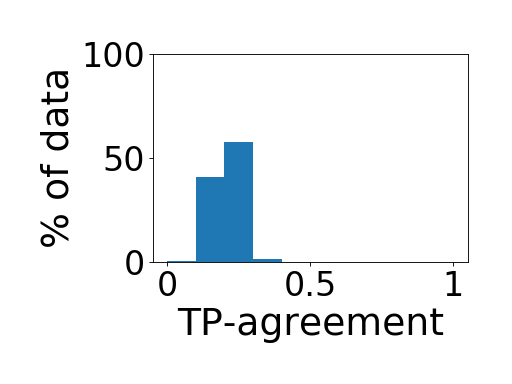}
\end{subfigure}
\begin{subfigure}{.135\textwidth}
  \centering
  \includegraphics[width=1\linewidth]{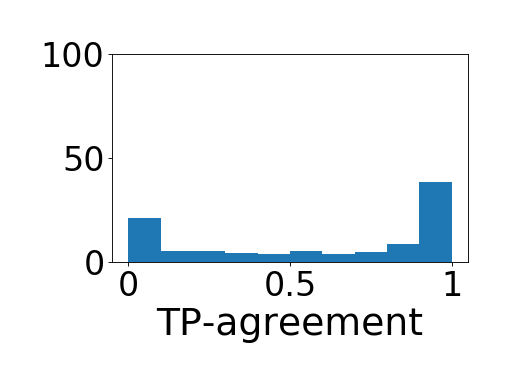}
\end{subfigure}
\begin{subfigure}{.135\textwidth}
  \centering
  \includegraphics[width=1\linewidth]{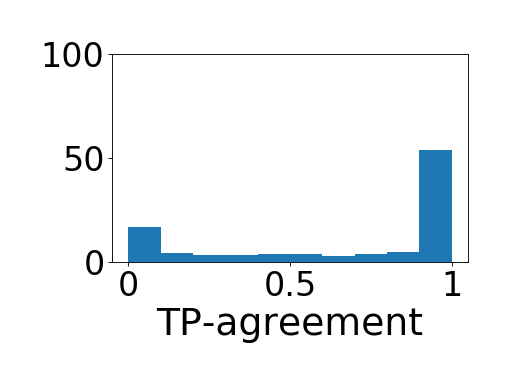}
\end{subfigure}
\begin{subfigure}{.135\textwidth}
  \centering
  \includegraphics[width=1\linewidth]{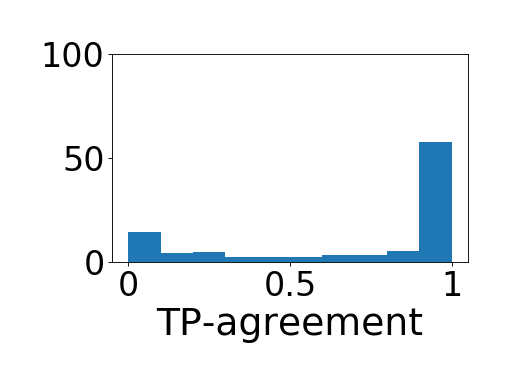}
\end{subfigure}
\begin{subfigure}{.135\textwidth}
  \centering
  \includegraphics[width=1\linewidth]{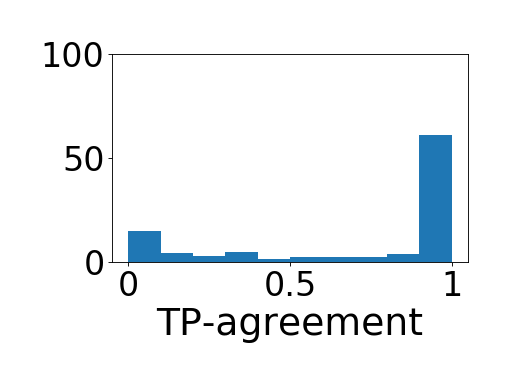}
\end{subfigure}
\begin{subfigure}{.135\textwidth}
  \centering
  \includegraphics[width=1\linewidth]{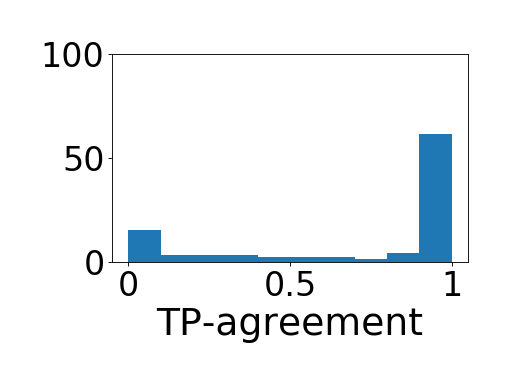}
\end{subfigure}
\begin{subfigure}{.135\textwidth}
  \centering
  \includegraphics[width=1\linewidth]{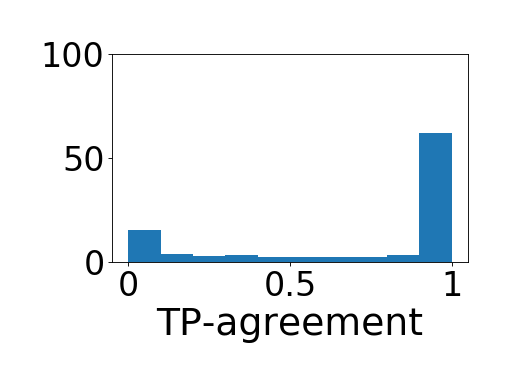}
\end{subfigure}

e)
\begin{subfigure}{.135\textwidth}
  \centering
  \includegraphics[width=1\linewidth]{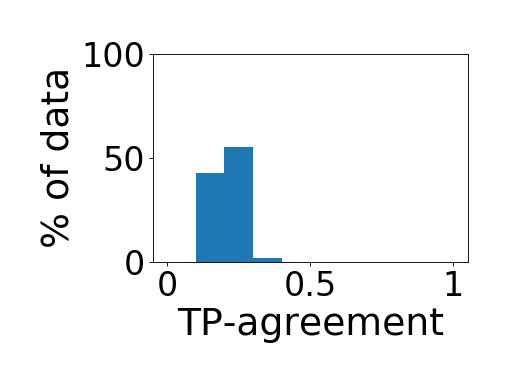}
\end{subfigure}
\begin{subfigure}{.135\textwidth}
  \centering
  \includegraphics[width=1\linewidth]{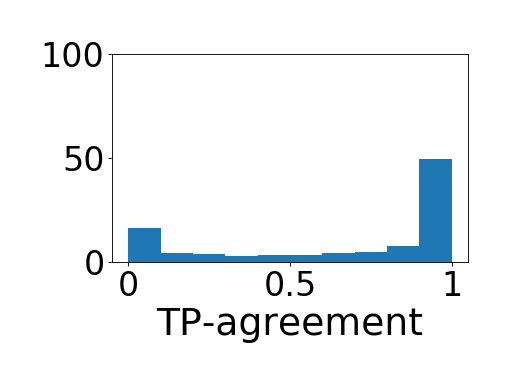}
\end{subfigure}
\begin{subfigure}{.135\textwidth}
  \centering
  \includegraphics[width=1\linewidth]{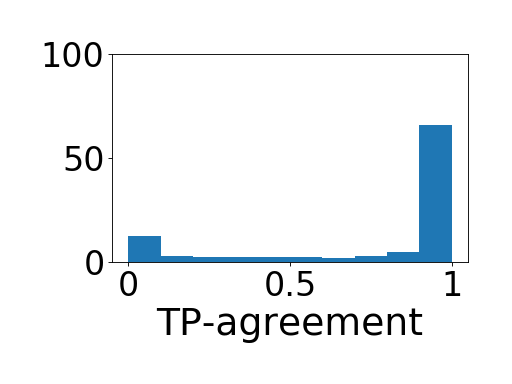}
\end{subfigure}
\begin{subfigure}{.135\textwidth}
  \centering
  \includegraphics[width=1\linewidth]{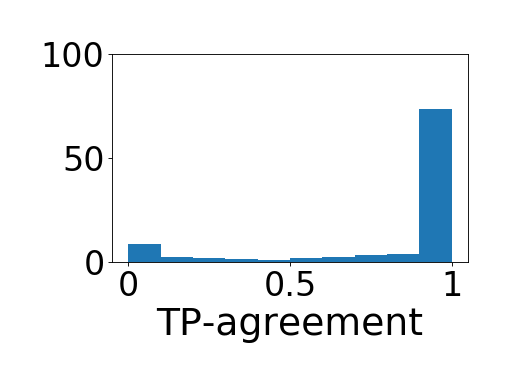}
\end{subfigure}
\begin{subfigure}{.135\textwidth}
  \centering
  \includegraphics[width=1\linewidth]{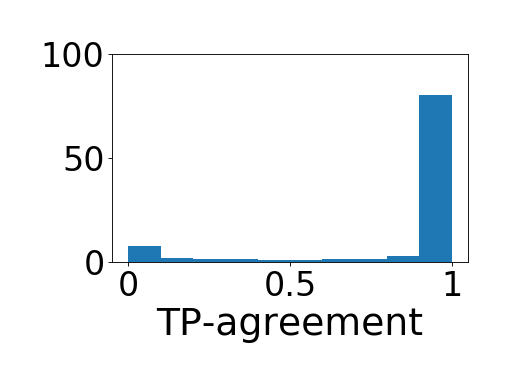}
\end{subfigure}
\begin{subfigure}{.135\textwidth}
  \centering
  \includegraphics[width=1\linewidth]{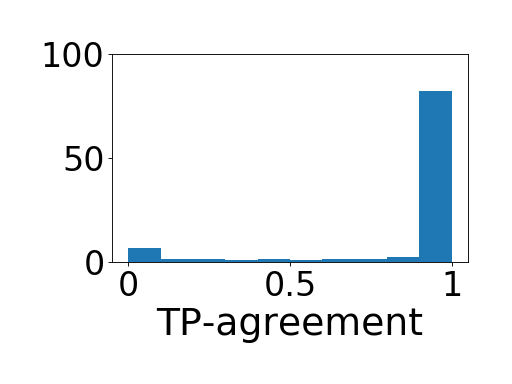}
\end{subfigure}
\begin{subfigure}{.135\textwidth}
  \centering
  \includegraphics[width=1\linewidth]{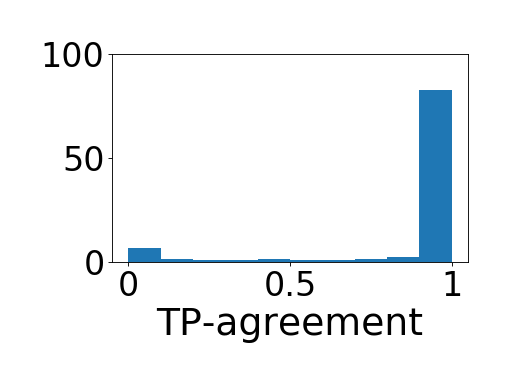}
\end{subfigure}

f)
\begin{subfigure}{.135\textwidth}
  \centering
  \includegraphics[width=1\linewidth]{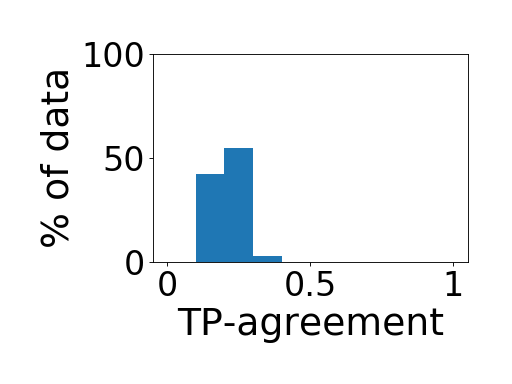}
\end{subfigure}
\begin{subfigure}{.135\textwidth}
  \centering
  \includegraphics[width=1\linewidth]{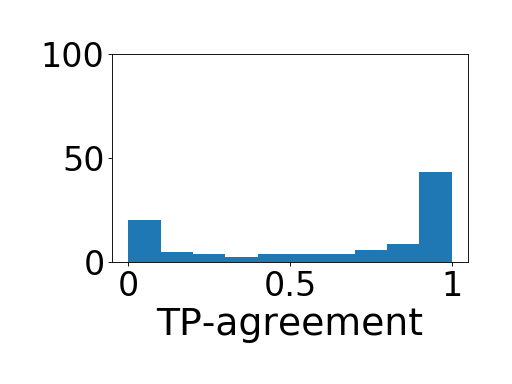}
\end{subfigure}
\begin{subfigure}{.135\textwidth}
  \centering
  \includegraphics[width=1\linewidth]{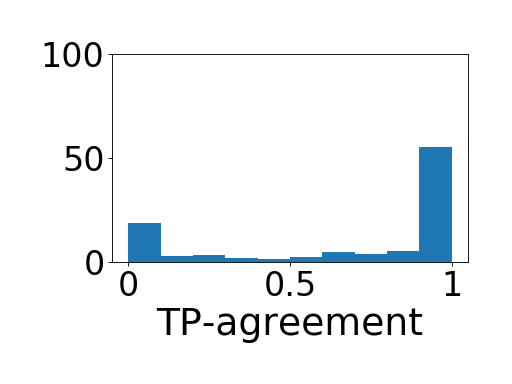}
\end{subfigure}
\begin{subfigure}{.135\textwidth}
  \centering
  \includegraphics[width=1\linewidth]{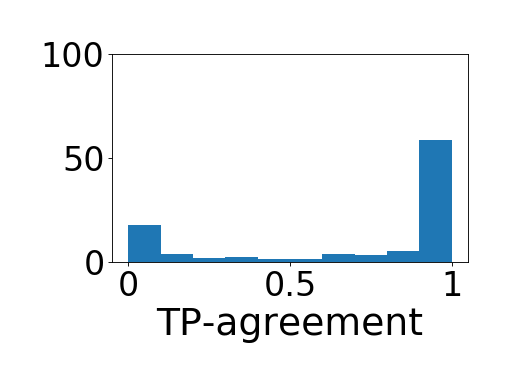}
\end{subfigure}
\begin{subfigure}{.135\textwidth}
  \centering
  \includegraphics[width=1\linewidth]{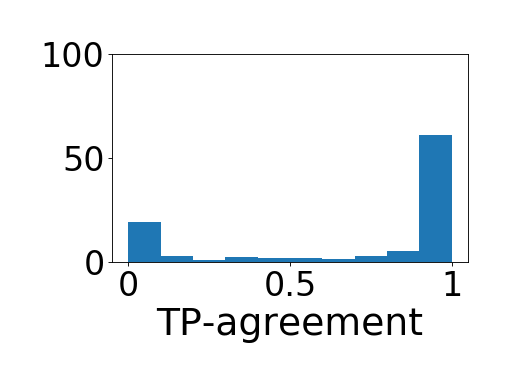}
\end{subfigure}
\begin{subfigure}{.135\textwidth}
  \centering
  \includegraphics[width=1\linewidth]{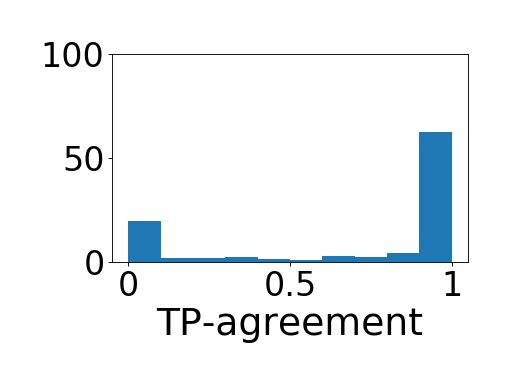}
\end{subfigure}
\begin{subfigure}{.135\textwidth}
  \centering
  \includegraphics[width=1\linewidth]{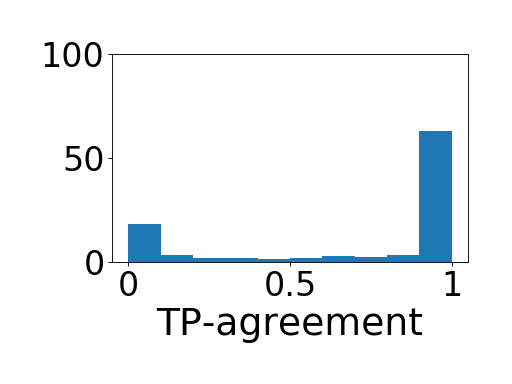}
\end{subfigure}
\caption{The distribution of TP-agreement scores during the learning process of $100$ instances of st-VGG (see \app~\ref{app:architectures}) on super-classes of CIFAR-100. Epochs shown: $0, 10, 30, 60, 90, 120, 140$. a-b) Train and test sets of the fish dataset respectively; c-d) train and test sets of the insect dataset respectively; e-f) train and test sets of the small mammals dataset respectively.}
  \label{appendix:full_results_subset1}
  \label{appendix:full_results_subset7}
  \label{appendix:full_results_subset16}
\end{figure*}

\begin{figure*}[hbt]
a)
\begin{subfigure}{.135\textwidth}
  \centering
  \includegraphics[width=1\linewidth]{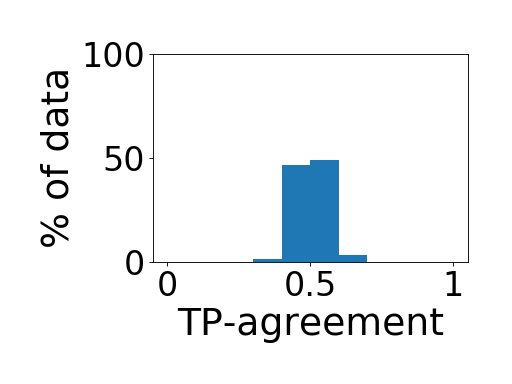}
\end{subfigure}
\begin{subfigure}{.135\textwidth}
  \centering
  \includegraphics[width=1\linewidth]{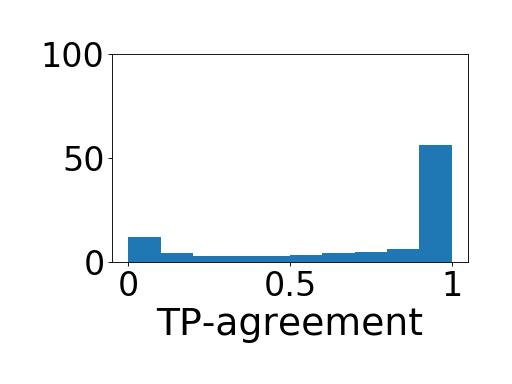}
\end{subfigure}
\begin{subfigure}{.135\textwidth}
  \centering
  \includegraphics[width=1\linewidth]{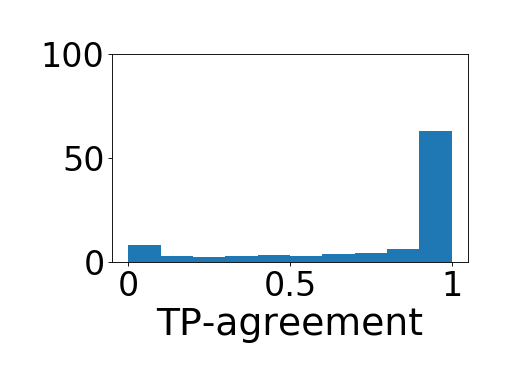}
\end{subfigure}
\begin{subfigure}{.135\textwidth}
  \centering
  \includegraphics[width=1\linewidth]{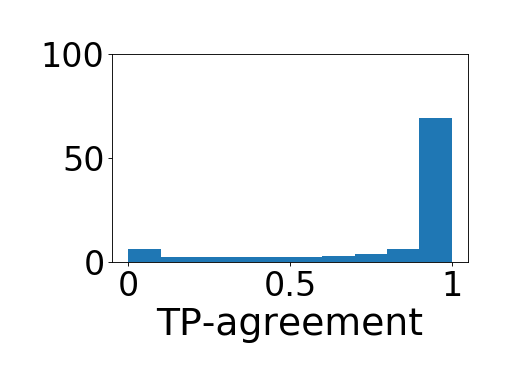}
\end{subfigure}
\begin{subfigure}{.135\textwidth}
  \centering
  \includegraphics[width=1\linewidth]{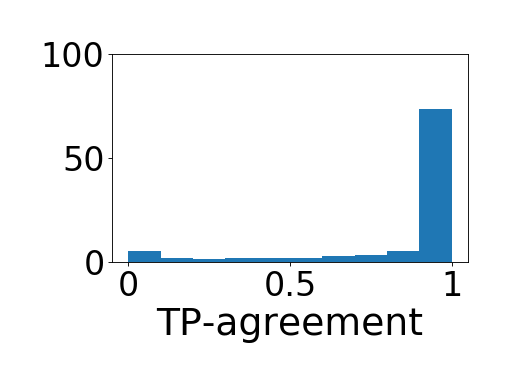}
\end{subfigure}
\begin{subfigure}{.135\textwidth}
  \centering
  \includegraphics[width=1\linewidth]{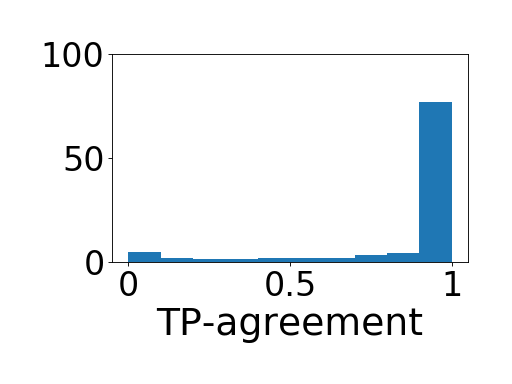}
\end{subfigure}
\begin{subfigure}{.135\textwidth}
  \centering
  \includegraphics[width=1\linewidth]{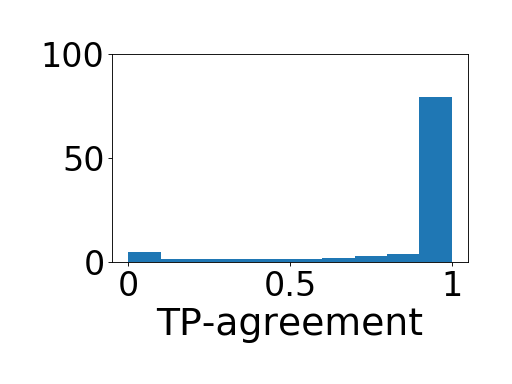}
\end{subfigure}

b)
\begin{subfigure}{.135\textwidth}
  \centering
  \includegraphics[width=1\linewidth]{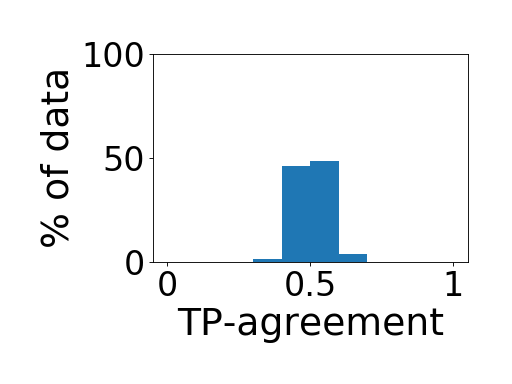}
\end{subfigure}
\begin{subfigure}{.135\textwidth}
  \centering
  \includegraphics[width=1\linewidth]{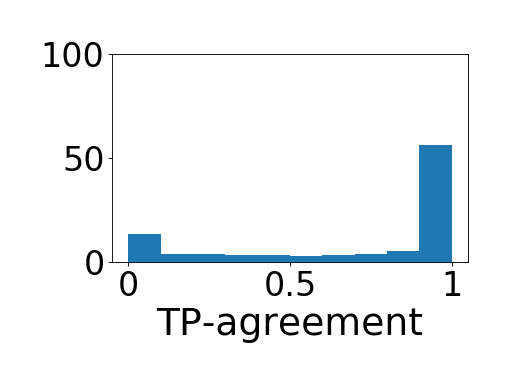}
\end{subfigure}
\begin{subfigure}{.135\textwidth}
  \centering
  \includegraphics[width=1\linewidth]{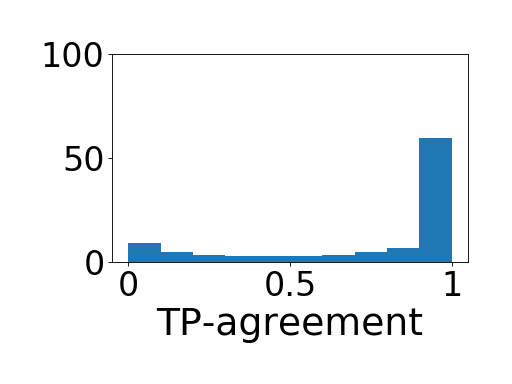}
\end{subfigure}
\begin{subfigure}{.135\textwidth}
  \centering
  \includegraphics[width=1\linewidth]{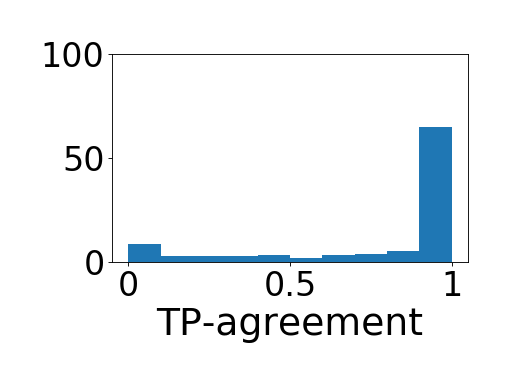}
\end{subfigure}
\begin{subfigure}{.135\textwidth}
  \centering
  \includegraphics[width=1\linewidth]{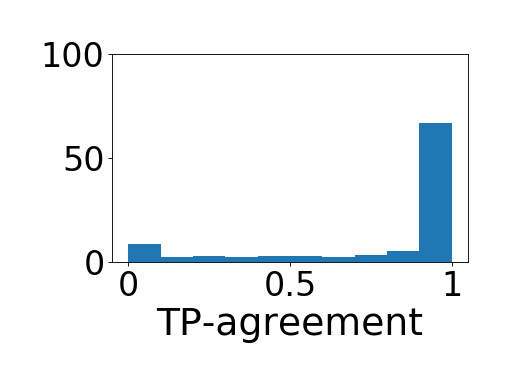}
\end{subfigure}
\begin{subfigure}{.135\textwidth}
  \centering
  \includegraphics[width=1\linewidth]{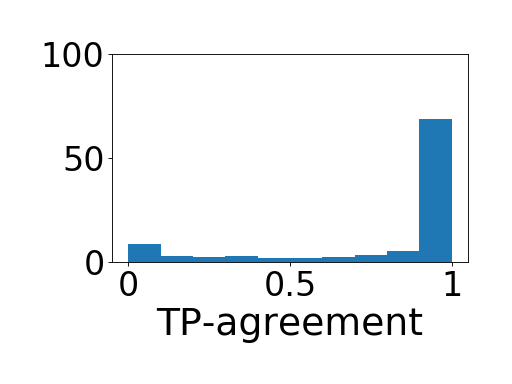}
\end{subfigure}
\begin{subfigure}{.135\textwidth}
  \centering
  \includegraphics[width=1\linewidth]{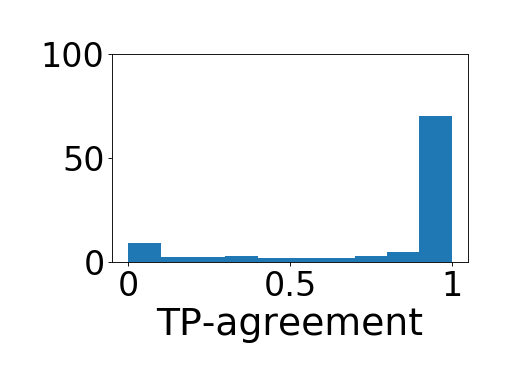}
\end{subfigure}
\caption{The distribution of TP-agreement scores during the learning process of $100$ instances of st-VGG (see \app~\ref{app:architectures}) trained on the cats and dogs binary dataset. Epochs shown: $0, 10, 30, 60, 90, 120, 140$. a) Train set; b) Test set.}
  \label{appendix:full_results_cats_dogs}
\end{figure*}

\begin{figure*}[hbt]
a)
\begin{subfigure}{.135\textwidth}
  \centering
  \includegraphics[width=1\linewidth]{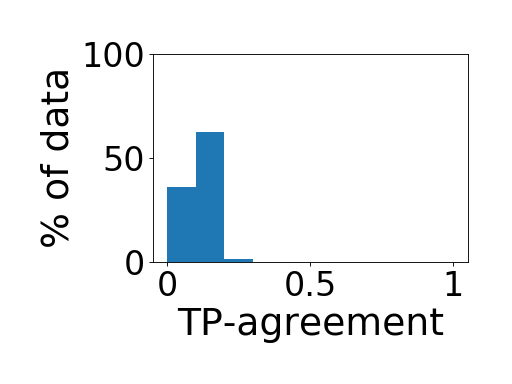}
\end{subfigure}
\begin{subfigure}{.135\textwidth}
  \centering
  \includegraphics[width=1\linewidth]{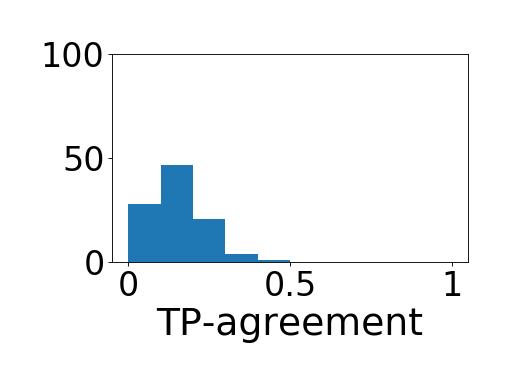}
\end{subfigure}
\begin{subfigure}{.135\textwidth}
  \centering
  \includegraphics[width=1\linewidth]{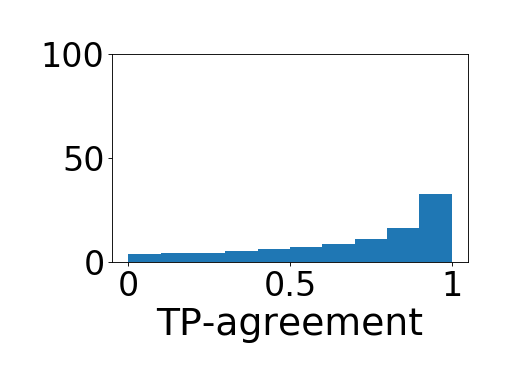}
\end{subfigure}
\begin{subfigure}{.135\textwidth}
  \centering
  \includegraphics[width=1\linewidth]{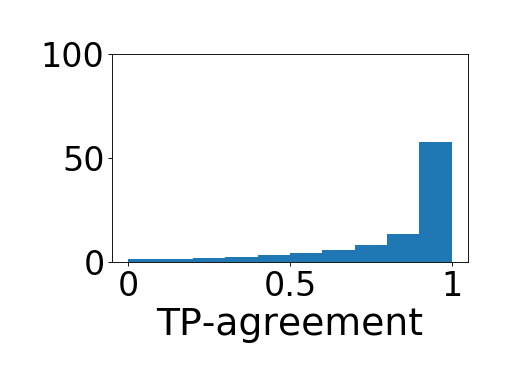}
\end{subfigure}
\begin{subfigure}{.135\textwidth}
  \centering
  \includegraphics[width=1\linewidth]{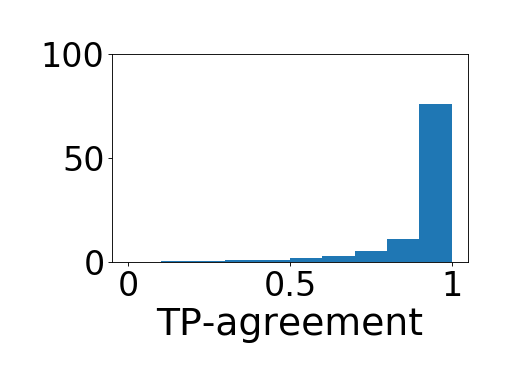}
\end{subfigure}
\begin{subfigure}{.135\textwidth}
  \centering
  \includegraphics[width=1\linewidth]{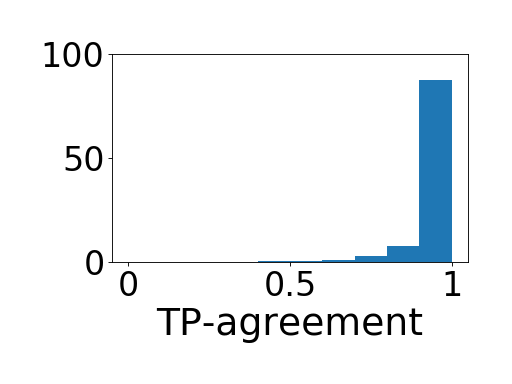}
\end{subfigure}
\begin{subfigure}{.135\textwidth}
  \centering
  \includegraphics[width=1\linewidth]{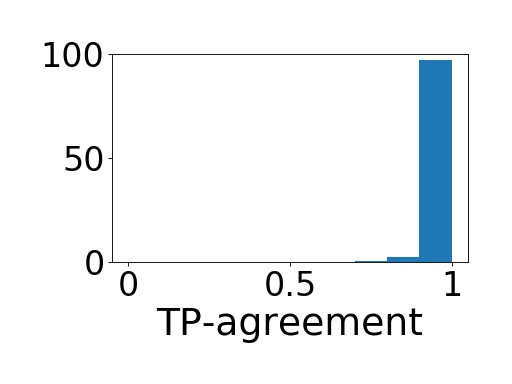}
\end{subfigure}

b)
\begin{subfigure}{.135\textwidth}
  \centering
  \includegraphics[width=1\linewidth]{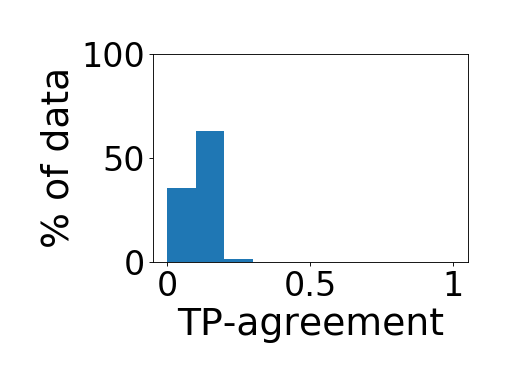}
\end{subfigure}
\begin{subfigure}{.135\textwidth}
  \centering
  \includegraphics[width=1\linewidth]{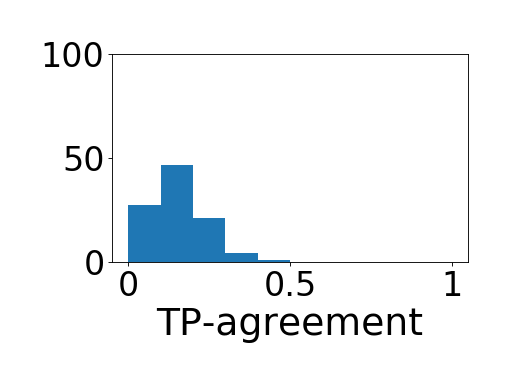}
\end{subfigure}
\begin{subfigure}{.135\textwidth}
  \centering
  \includegraphics[width=1\linewidth]{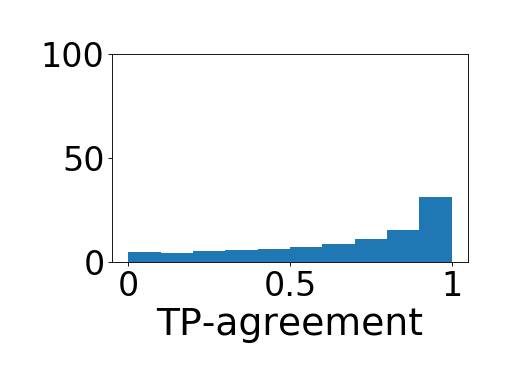}
\end{subfigure}
\begin{subfigure}{.135\textwidth}
  \centering
  \includegraphics[width=1\linewidth]{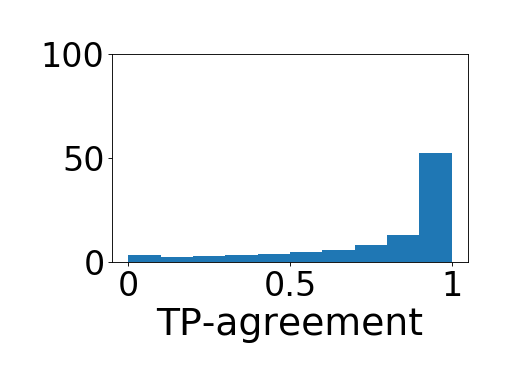}
\end{subfigure}
\begin{subfigure}{.135\textwidth}
  \centering
  \includegraphics[width=1\linewidth]{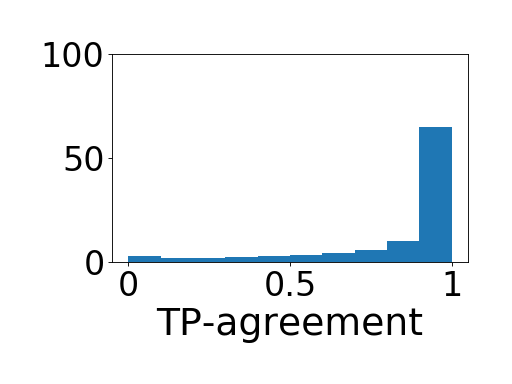}
\end{subfigure}
\begin{subfigure}{.135\textwidth}
  \centering
  \includegraphics[width=1\linewidth]{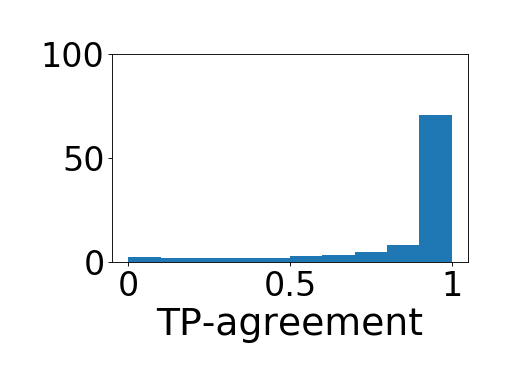}
\end{subfigure}
\begin{subfigure}{.135\textwidth}
  \centering
  \includegraphics[width=1\linewidth]{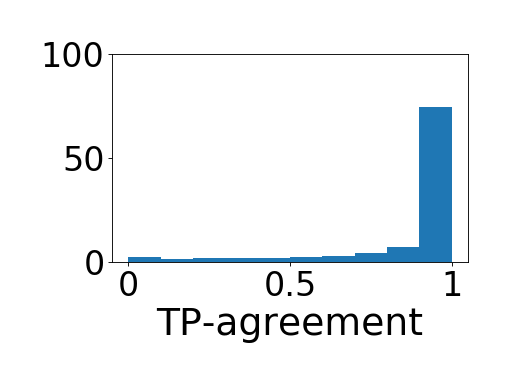}
\end{subfigure}
\caption{The distribution of TP-agreement scores during the learning process of $19$ instances of VGG19, trained on CIFAR-10. Epochs shown: $0, 1, 10, 30, 60, 80, 100$. a) Train set; b) Test set.}
  \label{appendix:full_results_vggcifar10}
\end{figure*}

\begin{figure*}[hbt]
a)
\begin{subfigure}{.135\textwidth}
  \centering
  \includegraphics[width=1\linewidth]{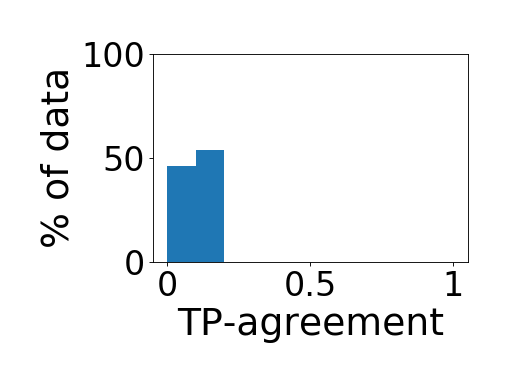}
\end{subfigure}
\begin{subfigure}{.135\textwidth}
  \centering
  \includegraphics[width=1\linewidth]{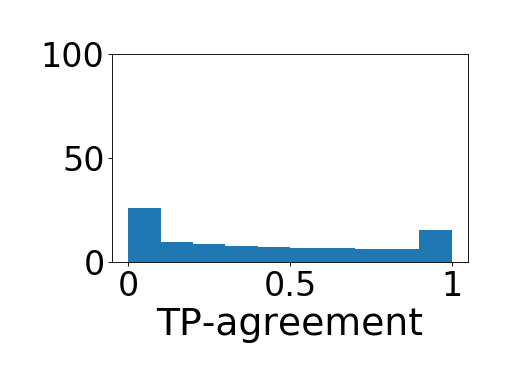}
\end{subfigure}
\begin{subfigure}{.135\textwidth}
  \centering
  \includegraphics[width=1\linewidth]{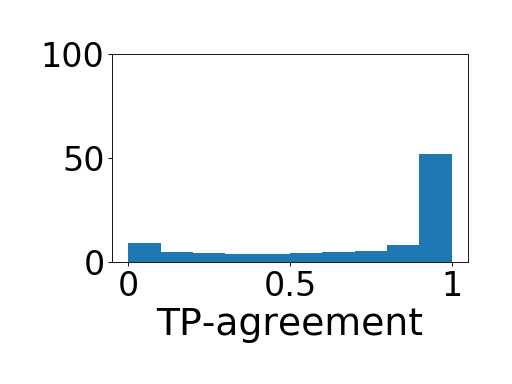}
\end{subfigure}
\begin{subfigure}{.135\textwidth}
  \centering
  \includegraphics[width=1\linewidth]{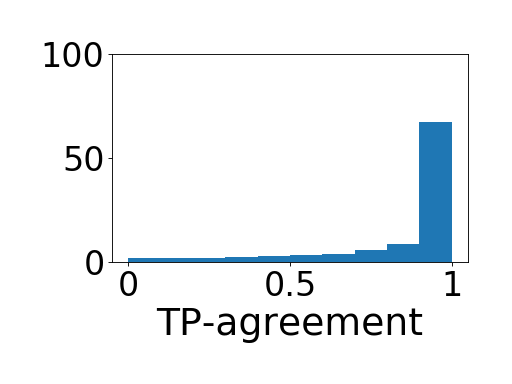}
\end{subfigure}
\begin{subfigure}{.135\textwidth}
  \centering
  \includegraphics[width=1\linewidth]{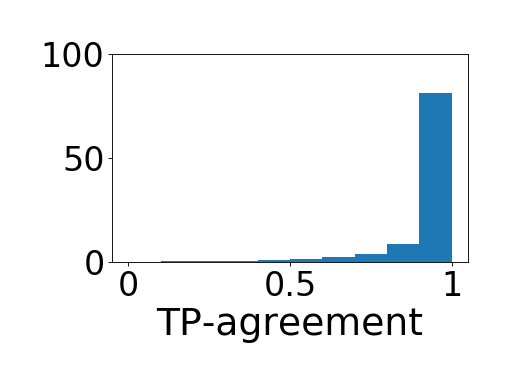}
\end{subfigure}
\begin{subfigure}{.135\textwidth}
  \centering
  \includegraphics[width=1\linewidth]{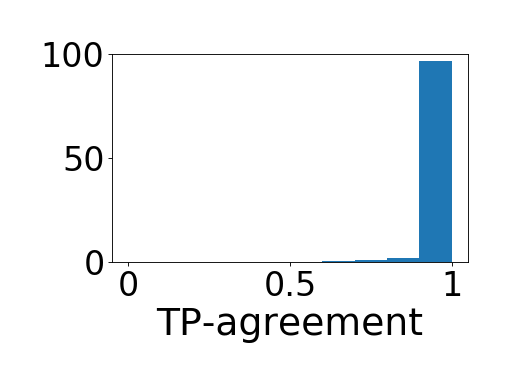}
\end{subfigure}
\begin{subfigure}{.135\textwidth}
  \centering
  \includegraphics[width=1\linewidth]{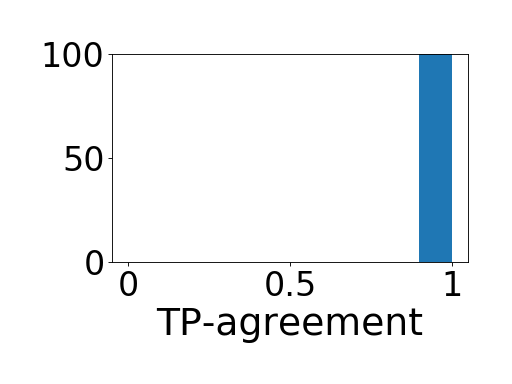}
\end{subfigure}

b)
\begin{subfigure}{.135\textwidth}
  \centering
  \includegraphics[width=1\linewidth]{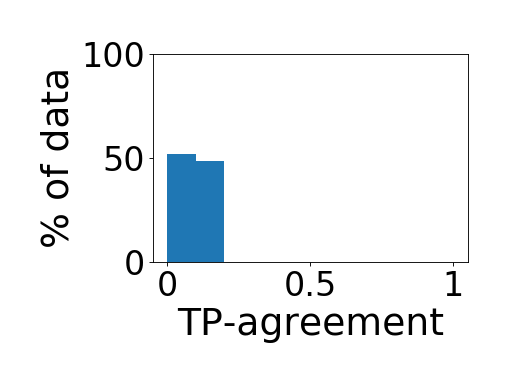}
\end{subfigure}
\begin{subfigure}{.135\textwidth}
  \centering
  \includegraphics[width=1\linewidth]{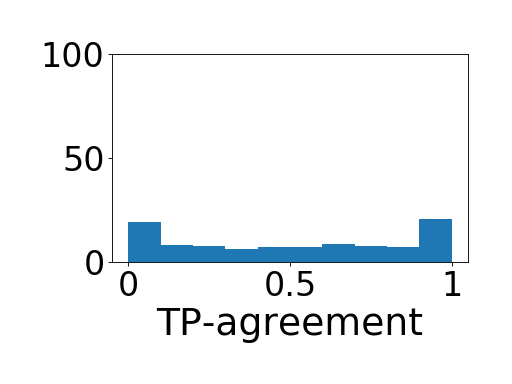}
\end{subfigure}
\begin{subfigure}{.135\textwidth}
  \centering
  \includegraphics[width=1\linewidth]{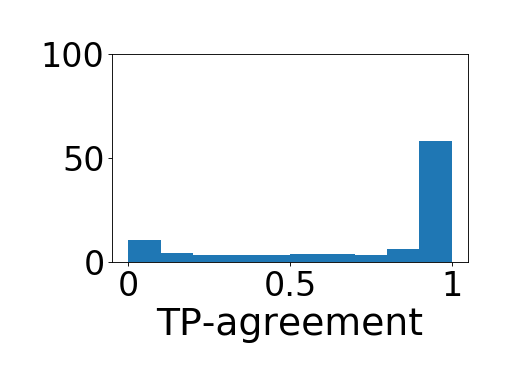}
\end{subfigure}
\begin{subfigure}{.135\textwidth}
  \centering
  \includegraphics[width=1\linewidth]{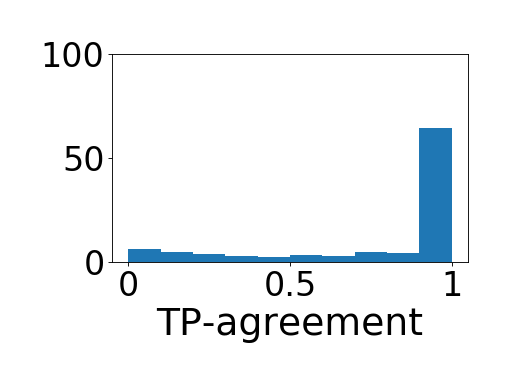}
\end{subfigure}
\begin{subfigure}{.135\textwidth}
  \centering
  \includegraphics[width=1\linewidth]{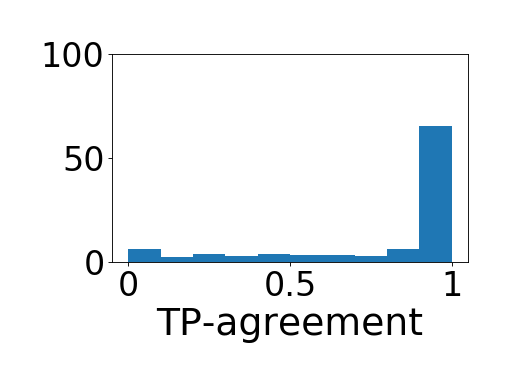}
\end{subfigure}
\begin{subfigure}{.135\textwidth}
  \centering
  \includegraphics[width=1\linewidth]{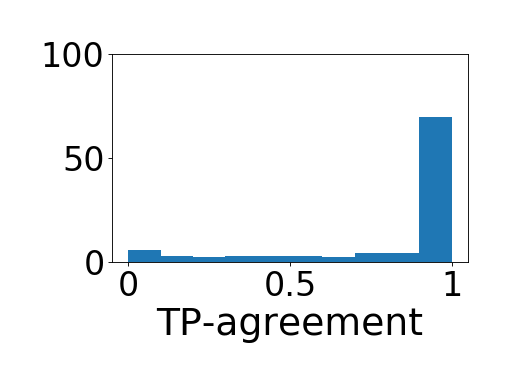}
\end{subfigure}
\begin{subfigure}{.135\textwidth}
  \centering
  \includegraphics[width=1\linewidth]{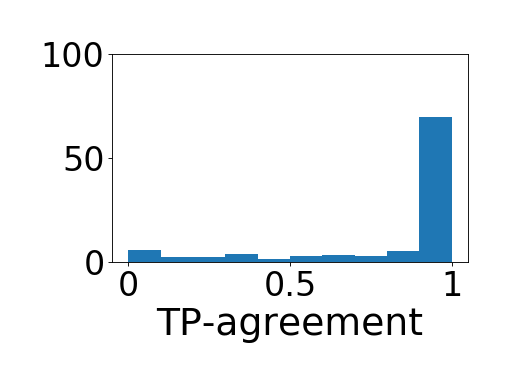}
\end{subfigure}
\caption{The distribution of TP-agreement scores during the learning process of $20$ instances of st-VGG (see \app~\ref{app:architectures}) trained on the face classification task (see \app~\ref{app:datasets}). Epochs shown: $0, 1, 10, 20, 30, 40, 60$. a) Train set; b) Test set.}
  \label{appendix:full_results_vggfaces2}
\end{figure*}

\begin{figure*}[hbt]
a)
\begin{subfigure}{.135\textwidth}
  \centering
  \includegraphics[width=1\linewidth]{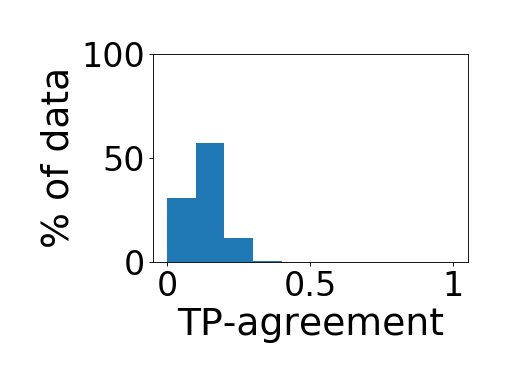}
\end{subfigure}
\begin{subfigure}{.135\textwidth}
  \centering
  \includegraphics[width=1\linewidth]{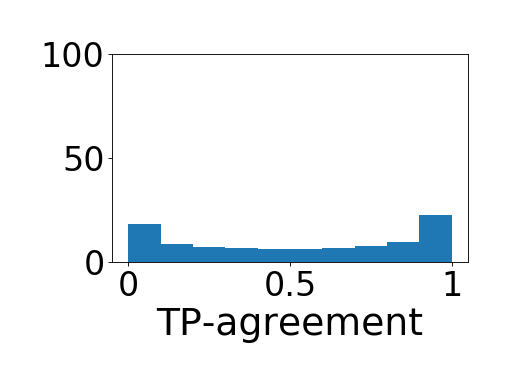}
\end{subfigure}
\begin{subfigure}{.135\textwidth}
  \centering
  \includegraphics[width=1\linewidth]{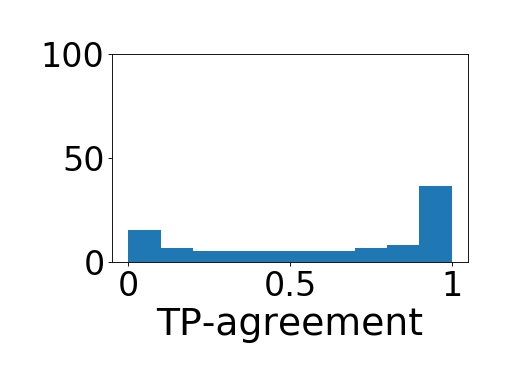}
\end{subfigure}
\begin{subfigure}{.135\textwidth}
  \centering
  \includegraphics[width=1\linewidth]{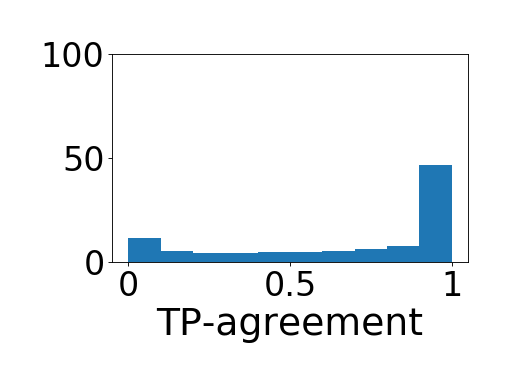}
\end{subfigure}
\begin{subfigure}{.135\textwidth}
  \centering
  \includegraphics[width=1\linewidth]{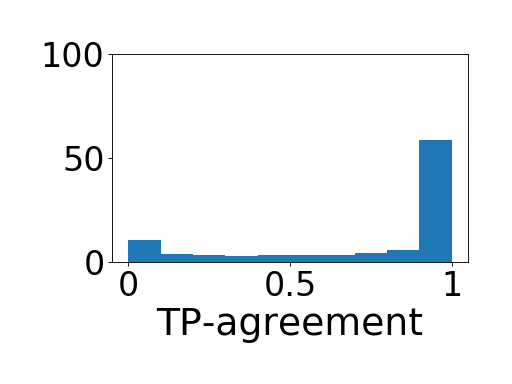}
\end{subfigure}
\begin{subfigure}{.135\textwidth}
  \centering
  \includegraphics[width=1\linewidth]{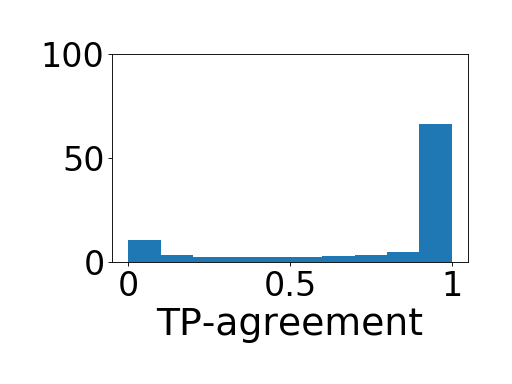}
\end{subfigure}
\begin{subfigure}{.135\textwidth}
  \centering
  \includegraphics[width=1\linewidth]{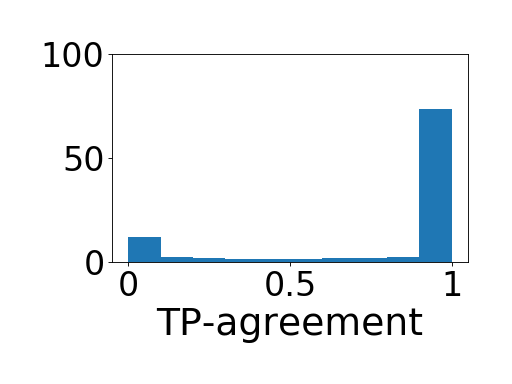}
\end{subfigure}

b)
\begin{subfigure}{.135\textwidth}
  \centering
  \includegraphics[width=1\linewidth]{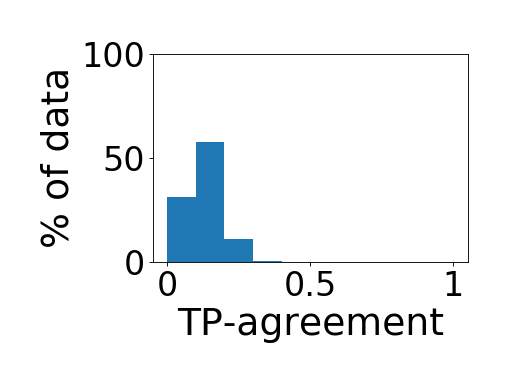}
\end{subfigure}
\begin{subfigure}{.135\textwidth}
  \centering
  \includegraphics[width=1\linewidth]{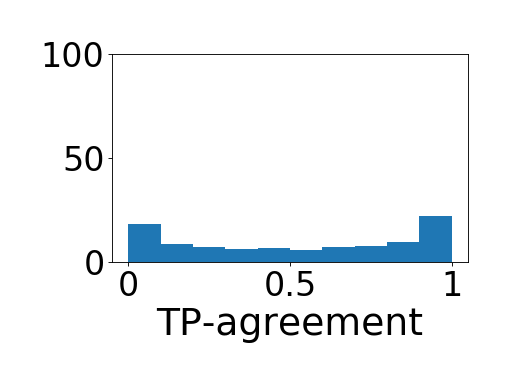}
\end{subfigure}
\begin{subfigure}{.135\textwidth}
  \centering
  \includegraphics[width=1\linewidth]{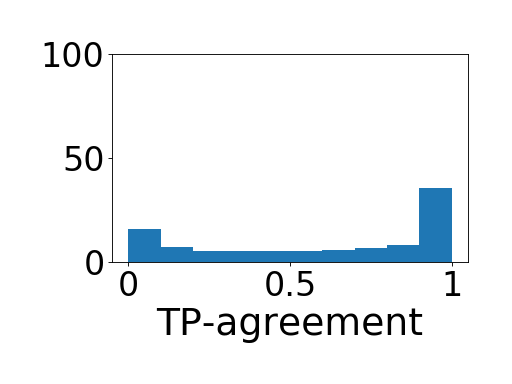}
\end{subfigure}
\begin{subfigure}{.135\textwidth}
  \centering
  \includegraphics[width=1\linewidth]{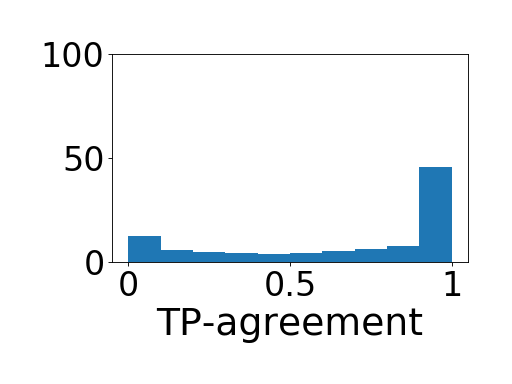}
\end{subfigure}
\begin{subfigure}{.135\textwidth}
  \centering
  \includegraphics[width=1\linewidth]{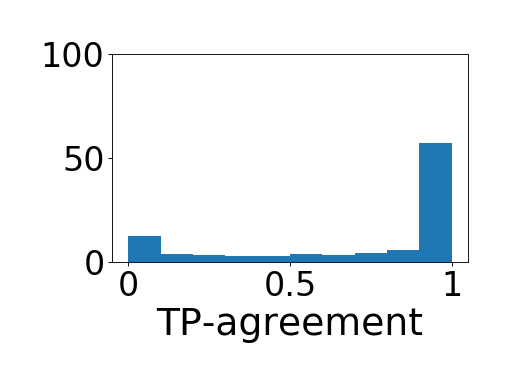}
\end{subfigure}
\begin{subfigure}{.135\textwidth}
  \centering
  \includegraphics[width=1\linewidth]{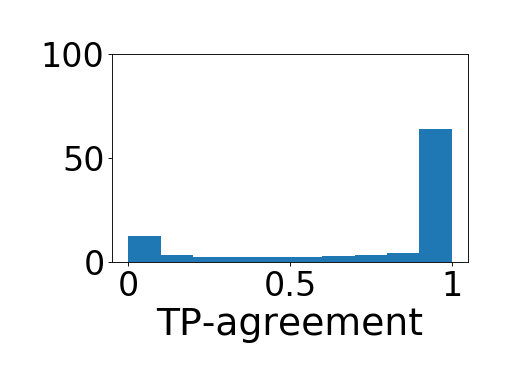}
\end{subfigure}
\begin{subfigure}{.135\textwidth}
  \centering
  \includegraphics[width=1\linewidth]{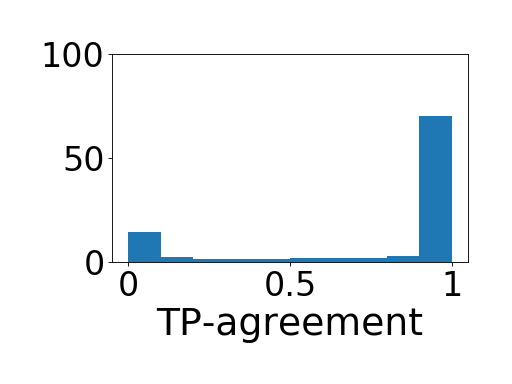}
\end{subfigure}

c)
\begin{subfigure}{.135\textwidth}
  \centering
  \includegraphics[width=1\linewidth]{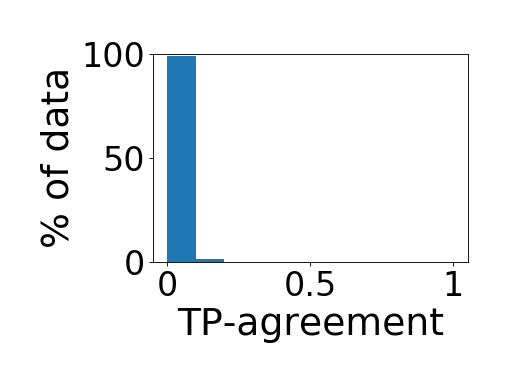}
\end{subfigure}
\begin{subfigure}{.135\textwidth}
  \centering
  \includegraphics[width=1\linewidth]{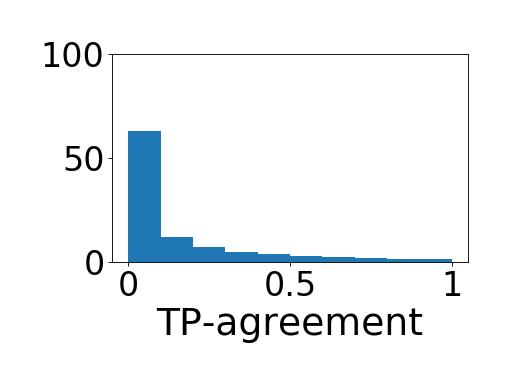}
\end{subfigure}
\begin{subfigure}{.135\textwidth}
  \centering
  \includegraphics[width=1\linewidth]{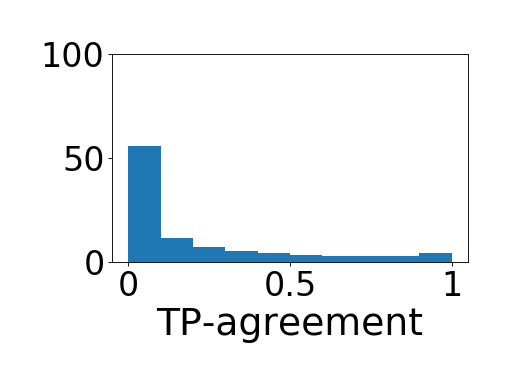}
\end{subfigure}
\begin{subfigure}{.135\textwidth}
  \centering
  \includegraphics[width=1\linewidth]{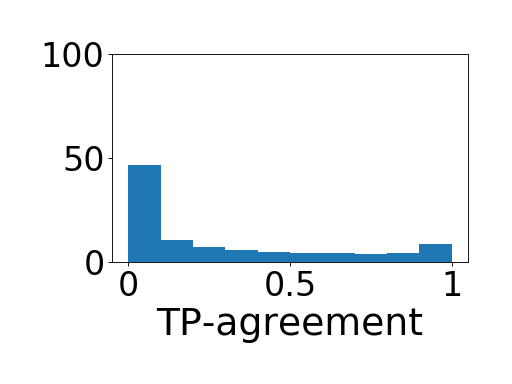}
\end{subfigure}
\begin{subfigure}{.135\textwidth}
  \centering
  \includegraphics[width=1\linewidth]{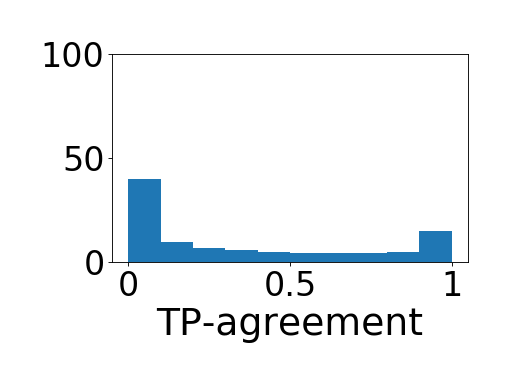}
\end{subfigure}
\begin{subfigure}{.135\textwidth}
  \centering
  \includegraphics[width=1\linewidth]{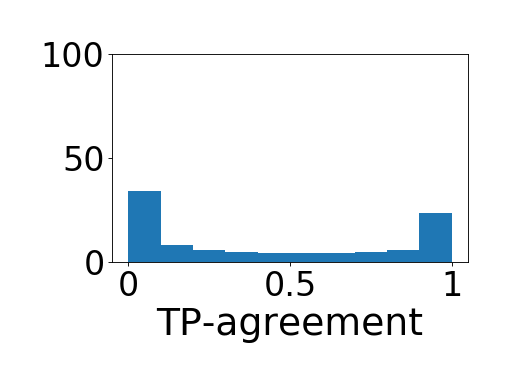}
\end{subfigure}
\begin{subfigure}{.135\textwidth}
  \centering
  \includegraphics[width=1\linewidth]{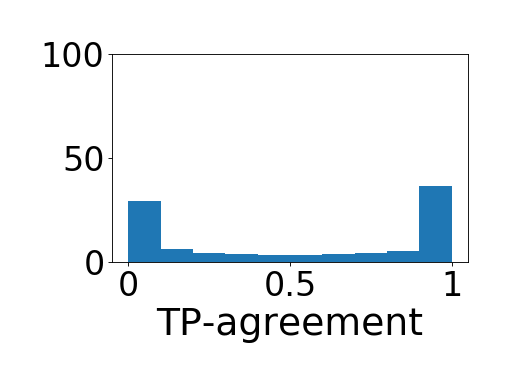}
\end{subfigure}

d)
\begin{subfigure}{.135\textwidth}
  \centering
  \includegraphics[width=1\linewidth]{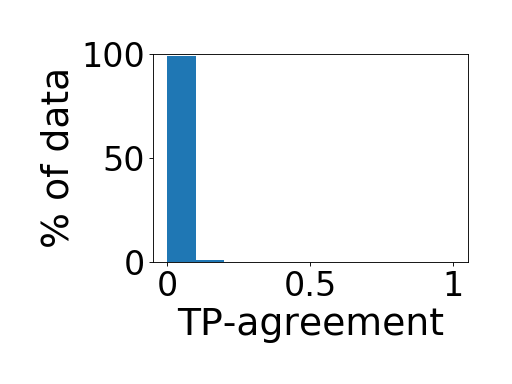}
\end{subfigure}
\begin{subfigure}{.135\textwidth}
  \centering
  \includegraphics[width=1\linewidth]{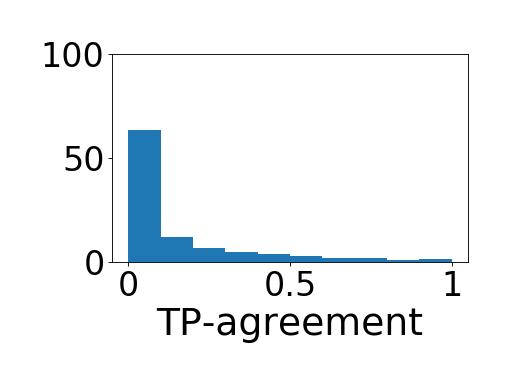}
\end{subfigure}
\begin{subfigure}{.135\textwidth}
  \centering
  \includegraphics[width=1\linewidth]{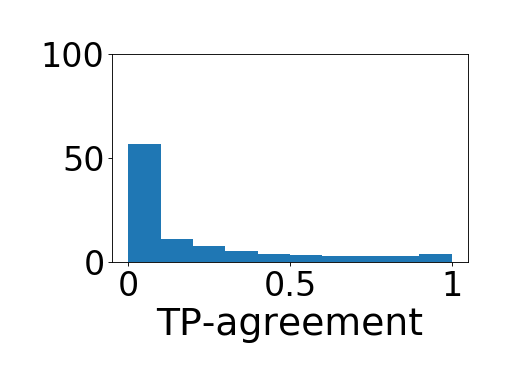}
\end{subfigure}
\begin{subfigure}{.135\textwidth}
  \centering
  \includegraphics[width=1\linewidth]{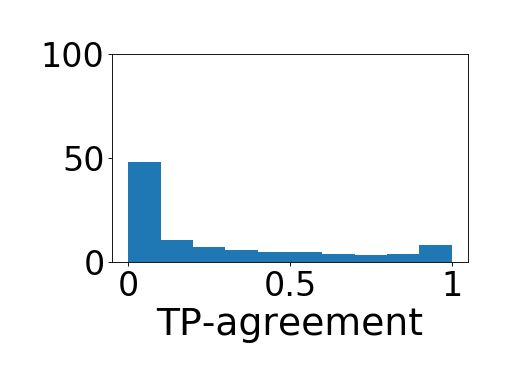}
\end{subfigure}
\begin{subfigure}{.135\textwidth}
  \centering
  \includegraphics[width=1\linewidth]{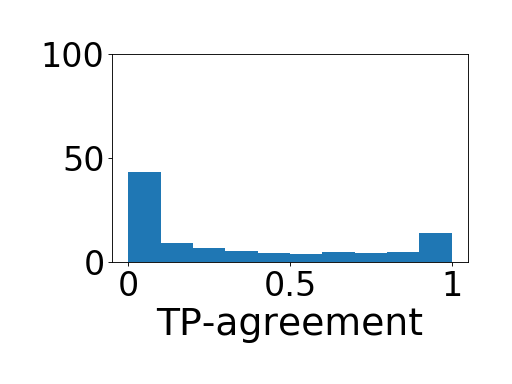}
\end{subfigure}
\begin{subfigure}{.135\textwidth}
  \centering
  \includegraphics[width=1\linewidth]{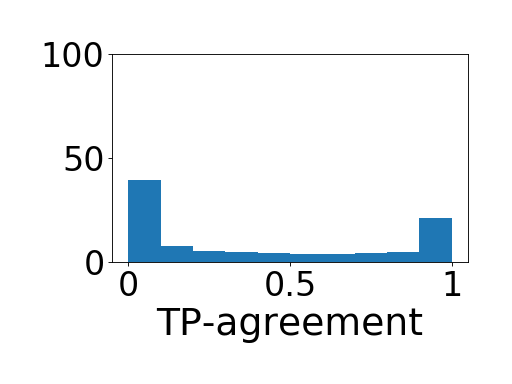}
\end{subfigure}
\begin{subfigure}{.135\textwidth}
  \centering
  \includegraphics[width=1\linewidth]{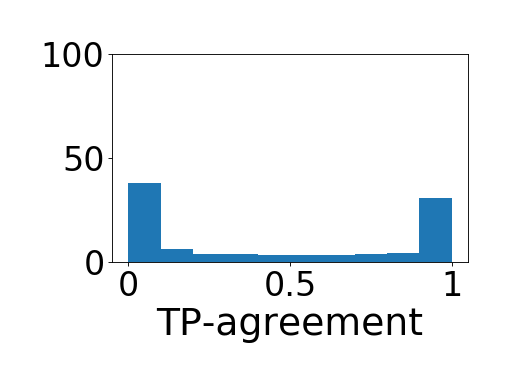}
\end{subfigure}
\caption{The distribution of TP-agreement scores during the learning process of $100$ instances of st-VGG (see \app~\ref{app:architectures}) trained on CIFAR-10 and CIFAR-100. Epochs shown: $0, 1, 2, 5, 10, 20, 40$. a-b) Train and test sets of CIFAR-10 respectively; c-d) train and test sets of CIFAR-100 respectively.}
  \label{appendix:full_results_st-vgg-cifar10}
  \label{appendix:full_results_st-vgg-cifar100}
\end{figure*}

\begin{figure*}[hbt]
a)
\begin{subfigure}{.135\textwidth}
  \centering
  \includegraphics[width=1\linewidth]{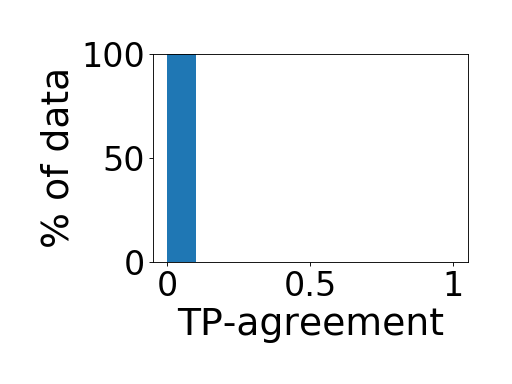}
\end{subfigure}
\begin{subfigure}{.135\textwidth}
  \centering
  \includegraphics[width=1\linewidth]{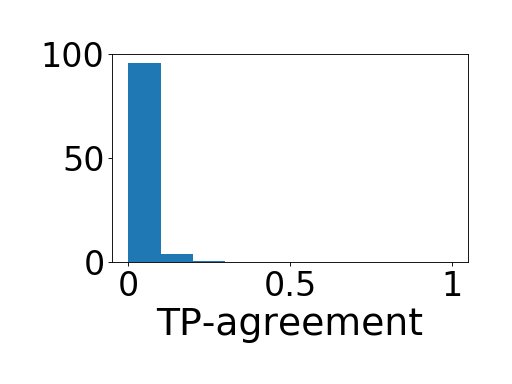}
\end{subfigure}
\begin{subfigure}{.135\textwidth}
  \centering
  \includegraphics[width=1\linewidth]{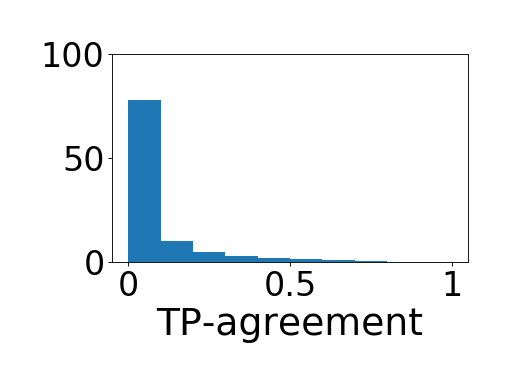}
\end{subfigure}
\begin{subfigure}{.135\textwidth}
  \centering
  \includegraphics[width=1\linewidth]{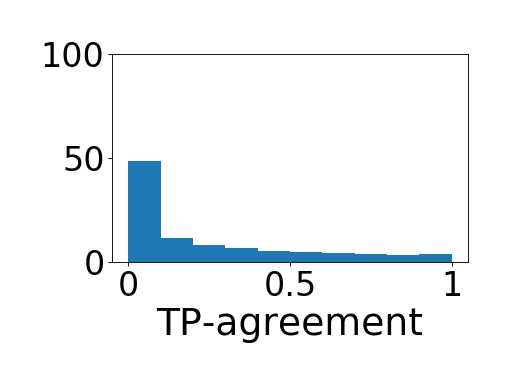}
\end{subfigure}
\begin{subfigure}{.135\textwidth}
  \centering
  \includegraphics[width=1\linewidth]{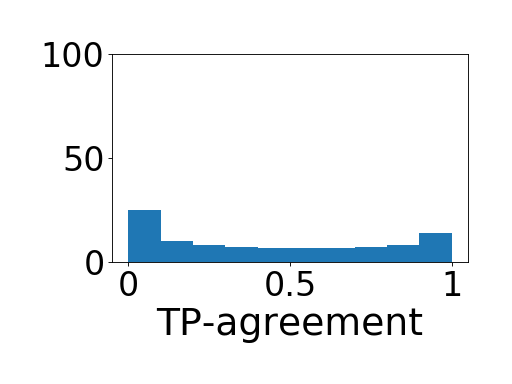}
\end{subfigure}
\begin{subfigure}{.135\textwidth}
  \centering
  \includegraphics[width=1\linewidth]{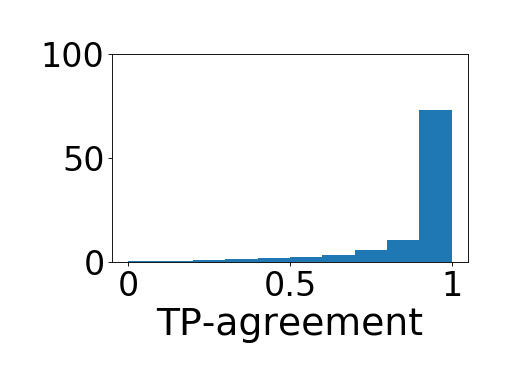}
\end{subfigure}
\begin{subfigure}{.135\textwidth}
  \centering
  \includegraphics[width=1\linewidth]{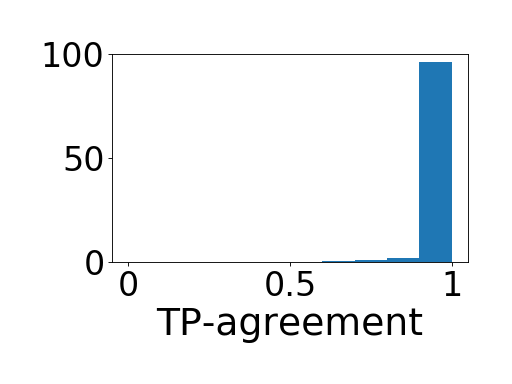}
\end{subfigure}

b)
\begin{subfigure}{.135\textwidth}
  \centering
  \includegraphics[width=1\linewidth]{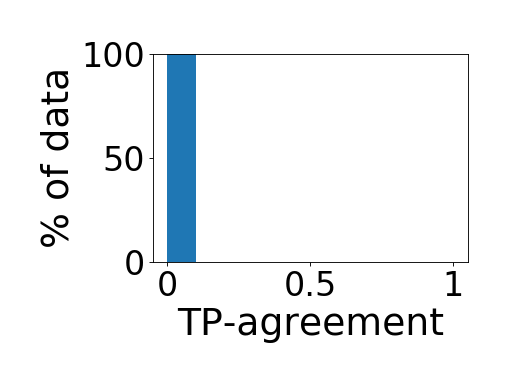}
\end{subfigure}
\begin{subfigure}{.135\textwidth}
  \centering
  \includegraphics[width=1\linewidth]{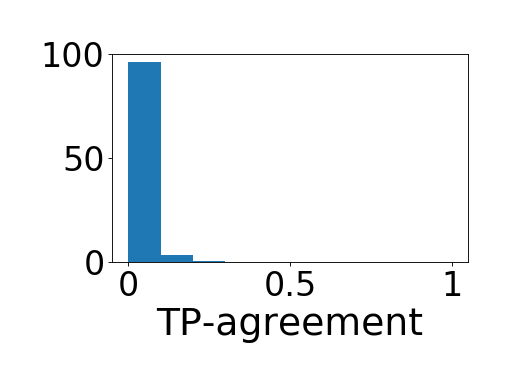}
\end{subfigure}
\begin{subfigure}{.135\textwidth}
  \centering
  \includegraphics[width=1\linewidth]{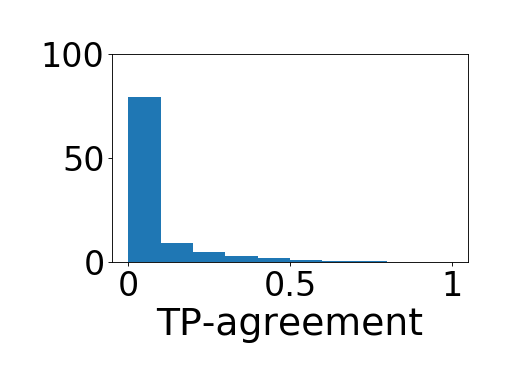}
\end{subfigure}
\begin{subfigure}{.135\textwidth}
  \centering
  \includegraphics[width=1\linewidth]{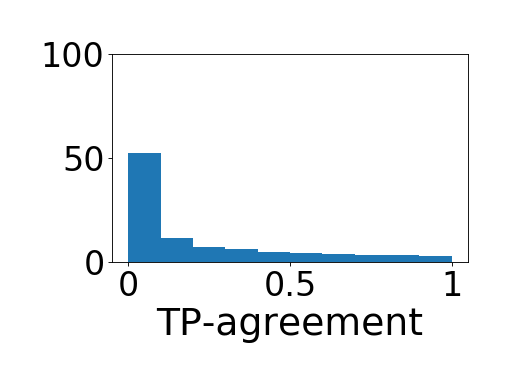}
\end{subfigure}
\begin{subfigure}{.135\textwidth}
  \centering
  \includegraphics[width=1\linewidth]{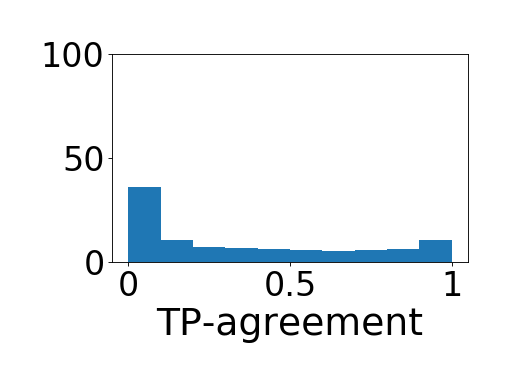}
\end{subfigure}
\begin{subfigure}{.135\textwidth}
  \centering
  \includegraphics[width=1\linewidth]{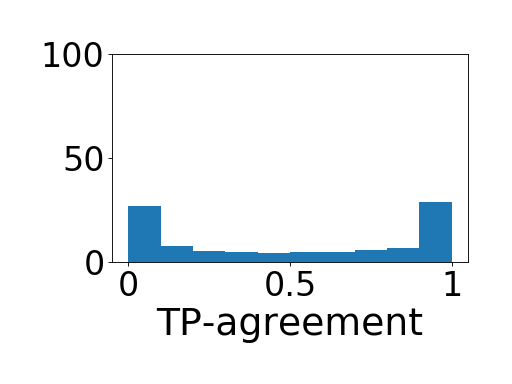}
\end{subfigure}
\begin{subfigure}{.135\textwidth}
  \centering
  \includegraphics[width=1\linewidth]{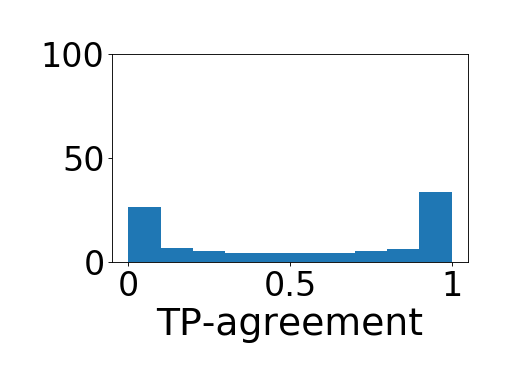}
\end{subfigure}
\caption{The distribution of TP-agreement scores during the learning process of $100$ instances of st-VGG (see \app~\ref{app:architectures}) trained on Tiny ImageNet. Epochs shown: $0, 1, 5, 10, 20, 50, 70$. a) Train set; b) Test set.}
  \label{appendix:full_results_tiny_imagenet}
\end{figure*}

\begin{figure*}[hbt]
a)
\begin{subfigure}{.135\textwidth}
  \centering
  \includegraphics[width=1\linewidth]{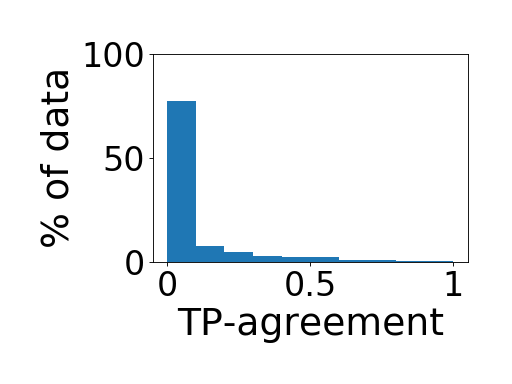}
\end{subfigure}
\begin{subfigure}{.135\textwidth}
  \centering
  \includegraphics[width=1\linewidth]{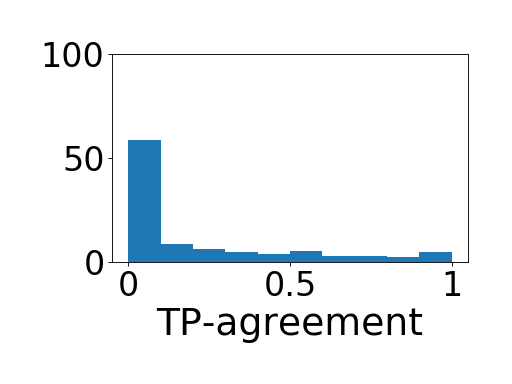}
\end{subfigure}
\begin{subfigure}{.135\textwidth}
  \centering
  \includegraphics[width=1\linewidth]{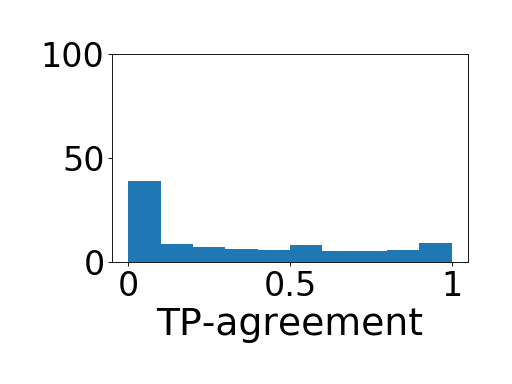}
\end{subfigure}
\begin{subfigure}{.135\textwidth}
  \centering
  \includegraphics[width=1\linewidth]{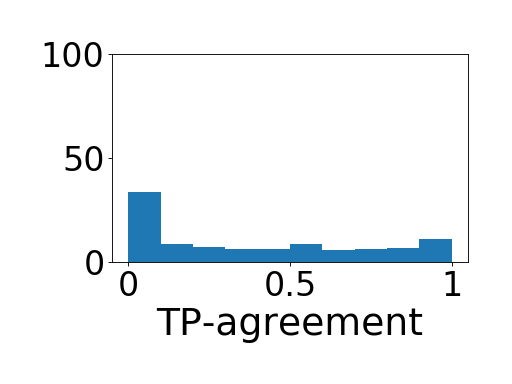}
\end{subfigure}
\begin{subfigure}{.135\textwidth}
  \centering
  \includegraphics[width=1\linewidth]{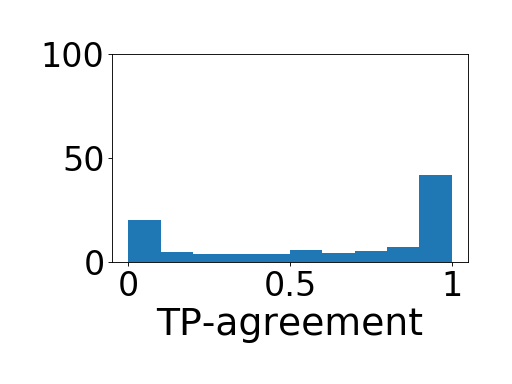}
\end{subfigure}
\begin{subfigure}{.135\textwidth}
  \centering
  \includegraphics[width=1\linewidth]{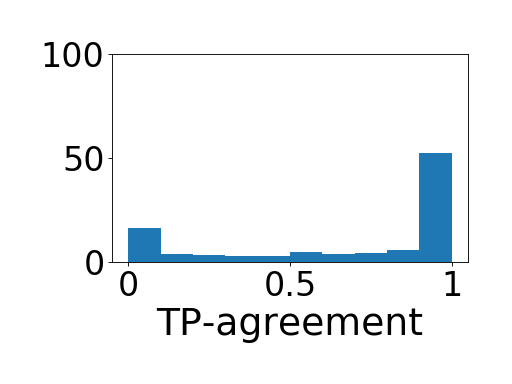}
\end{subfigure}
\begin{subfigure}{.135\textwidth}
  \centering
  \includegraphics[width=1\linewidth]{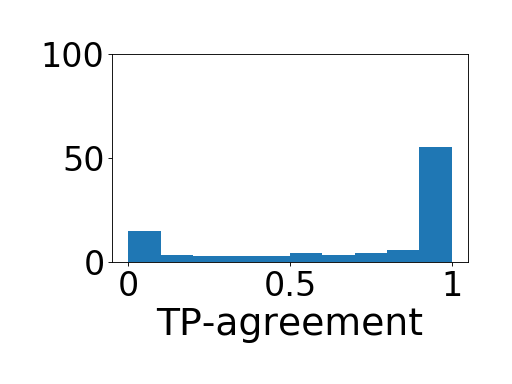}
\end{subfigure}

b)
\begin{subfigure}{.135\textwidth}
  \centering
  \includegraphics[width=1\linewidth]{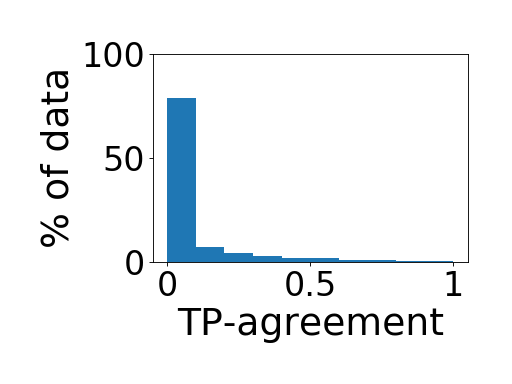}
\end{subfigure}
\begin{subfigure}{.135\textwidth}
  \centering
  \includegraphics[width=1\linewidth]{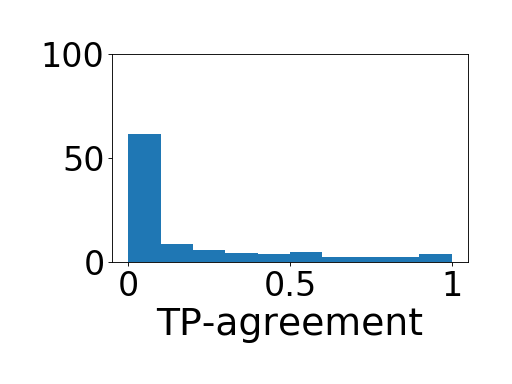}
\end{subfigure}
\begin{subfigure}{.135\textwidth}
  \centering
  \includegraphics[width=1\linewidth]{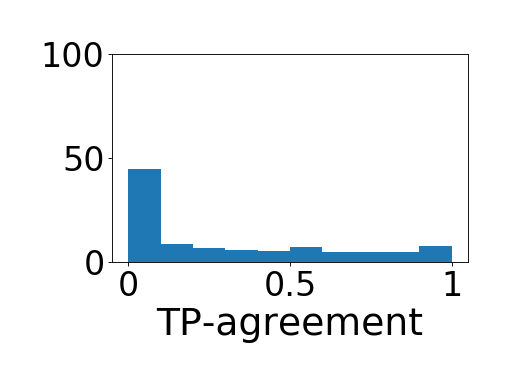}
\end{subfigure}
\begin{subfigure}{.135\textwidth}
  \centering
  \includegraphics[width=1\linewidth]{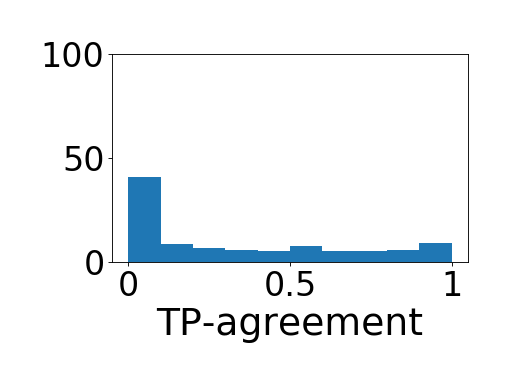}
\end{subfigure}
\begin{subfigure}{.135\textwidth}
  \centering
  \includegraphics[width=1\linewidth]{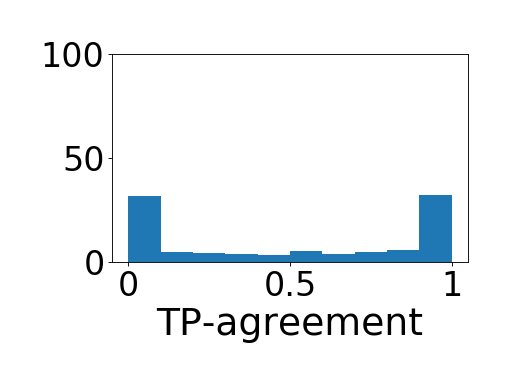}
\end{subfigure}
\begin{subfigure}{.135\textwidth}
  \centering
  \includegraphics[width=1\linewidth]{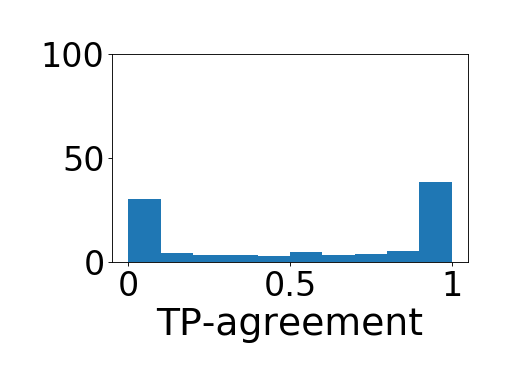}
\end{subfigure}
\begin{subfigure}{.135\textwidth}
  \centering
  \includegraphics[width=1\linewidth]{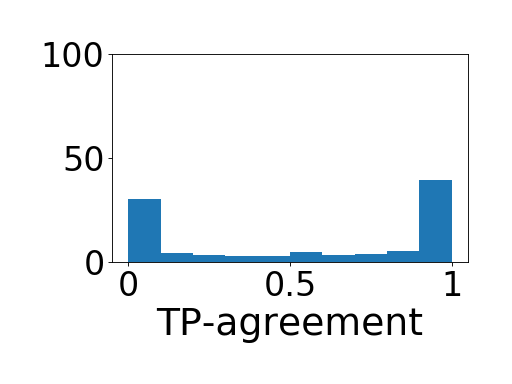}
\end{subfigure}

c)
\begin{subfigure}{.135\textwidth}
  \centering
  \includegraphics[width=1\linewidth]{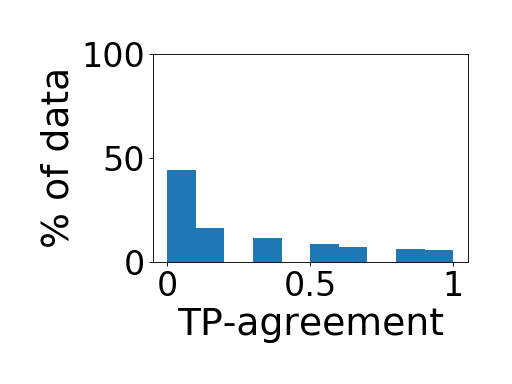}
\end{subfigure}
\begin{subfigure}{.135\textwidth}
  \centering
  \includegraphics[width=1\linewidth]{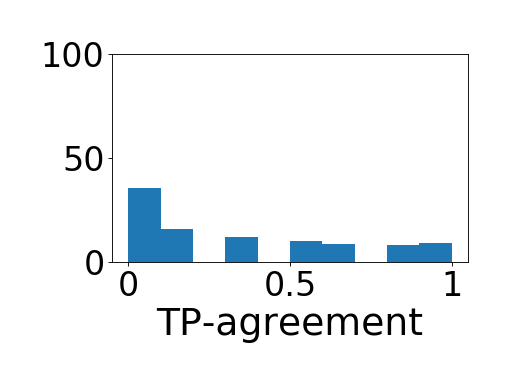}
\end{subfigure}
\begin{subfigure}{.135\textwidth}
  \centering
  \includegraphics[width=1\linewidth]{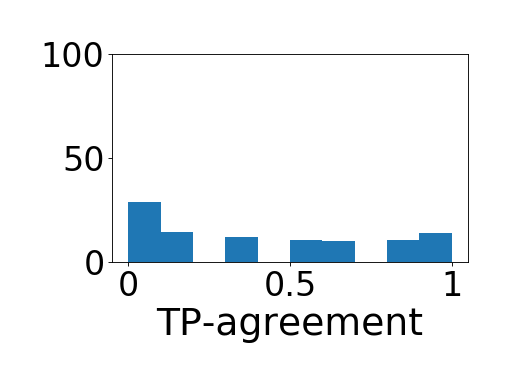}
\end{subfigure}
\begin{subfigure}{.135\textwidth}
  \centering
  \includegraphics[width=1\linewidth]{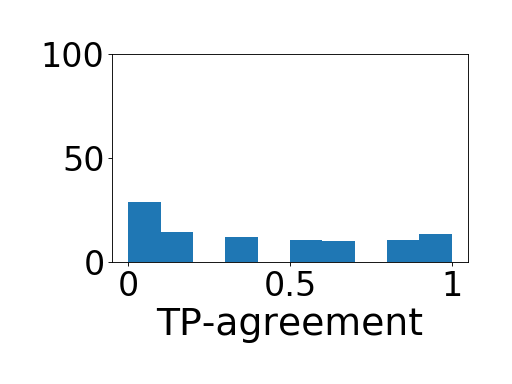}
\end{subfigure}
\begin{subfigure}{.135\textwidth}
  \centering
  \includegraphics[width=1\linewidth]{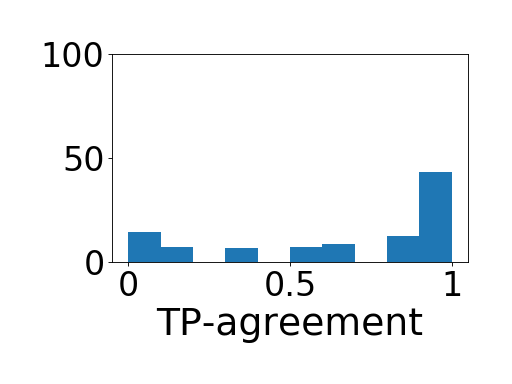}
\end{subfigure}
\begin{subfigure}{.135\textwidth}
  \centering
  \includegraphics[width=1\linewidth]{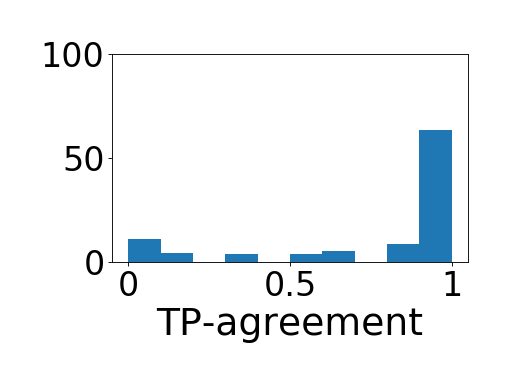}
\end{subfigure}
\begin{subfigure}{.135\textwidth}
  \centering
  \includegraphics[width=1\linewidth]{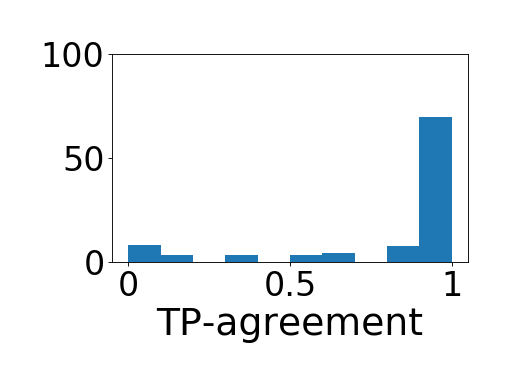}
\end{subfigure}

d)
\begin{subfigure}{.135\textwidth}
  \centering
  \includegraphics[width=1\linewidth]{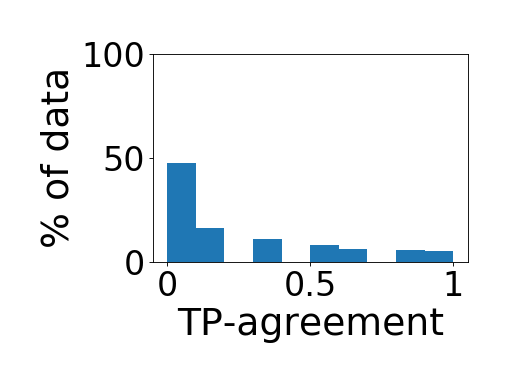}
\end{subfigure}
\begin{subfigure}{.135\textwidth}
  \centering
  \includegraphics[width=1\linewidth]{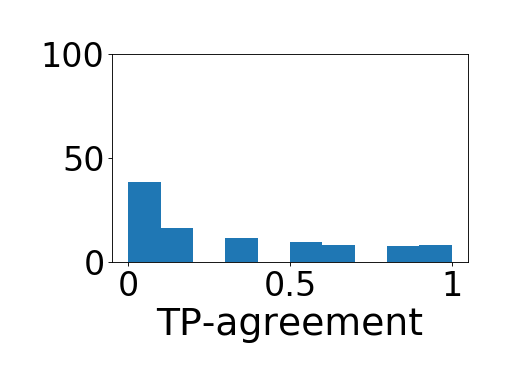}
\end{subfigure}
\begin{subfigure}{.135\textwidth}
  \centering
  \includegraphics[width=1\linewidth]{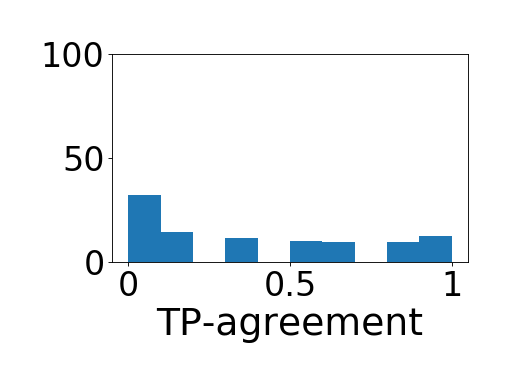}
\end{subfigure}
\begin{subfigure}{.135\textwidth}
  \centering
  \includegraphics[width=1\linewidth]{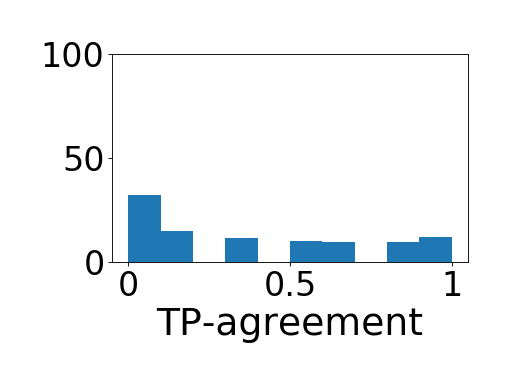}
\end{subfigure}
\begin{subfigure}{.135\textwidth}
  \centering
  \includegraphics[width=1\linewidth]{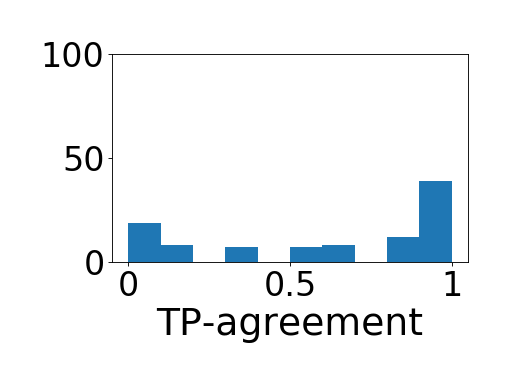}
\end{subfigure}
\begin{subfigure}{.135\textwidth}
  \centering
  \includegraphics[width=1\linewidth]{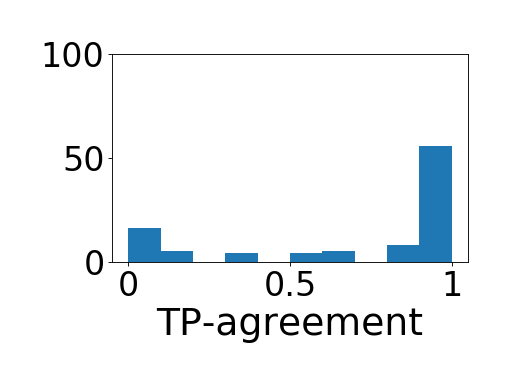}
\end{subfigure}
\begin{subfigure}{.135\textwidth}
  \centering
  \includegraphics[width=1\linewidth]{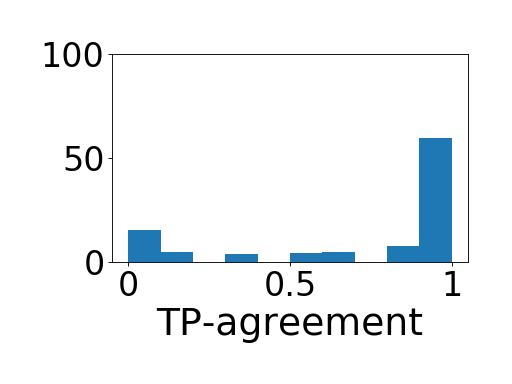}
\end{subfigure}

\caption{The distribution of TP-agreement scores of different architectures over ImageNet. Epochs shown: $1, 2, 10, 30, 50, 70, 100$. a-b) 22 instances of AlexNet over the train and test sets respectively. c-d) 6 instances of DenseNet over the train and test sets respectively. These results follow the same protocol described in Fig.~\ref{fig:consistency_dynamics}.}
  \label{appendix:full_results_imagenet_consistency_architectures}
\end{figure*}

As mentioned in \S\ref{section:robusteness}, results were obtained on the following combinations of datasets and architectures: ImageNet, trained on ResNet-50, DenseNet and AlexNet (Figs.~\ref{fig:consistency_dynamics},\ref{appendix:full_results_imagenet_consistency_architectures}). CIFAR-100 and CIFAR-10, both trained on VGG19 and st-vgg (Figs.~\ref{fig:full_results_vggcifar100},\ref{appendix:full_results_vggcifar10},\ref{appendix:full_results_st-vgg-cifar10}). MNIST trained on various small architectures (see \app~\ref{app:architectures} and Fig.~\ref{appendix:full_results_mnist}). Fashion-MNIST, trained on the same architectures as MNIST (Fig.~\ref{appendix:full_results_fashion_mnist}), Tiny ImageNet, trained on VGG19 (Fig.~\ref{appendix:full_results_tiny_imagenet}). VggFaces2, trained on st-VGG (Fig.~\ref{appendix:full_results_vggfaces2}). Small Mammals, Insect and Fish datasets, all trained on st-VGG (Fig.~\ref{appendix:full_results_subset1}), Cats and Dogs dataset, trained on st-VGG (Fig.~\ref{appendix:full_results_cats_dogs}). For all datasets and architectures, we used the hyper-parameters, augmentation and initialization as suggested in their original papers if available. In cases were such parameters were not available, we did some minimal parameter fine-tuning in order to achieve competitive top-1 accuracy. It is important to note, that the qualitative results remained the same for all hyper-parameters values we tested, both before and after fine-tuning.

We also reproduced our results while changing hyper-parameters either directly and indirectly. The indirect hyper-parameter change occurred as we tested different architectures, which has different hyper-parameters by definition. Hyper-parameters which were modified this way: i) Learning rate -- a wide range of values across different architectures, both with and without decay over epochs (ranges between $10^{-4}$ to $1$). ii) Optimizers -- SGD, Adam, AdaDelta all with various ranges of decay and momentum. iii) Batch size: wide range between $16$ and $256$. iv) Dropout: between $0$ and $0.5$. v) L2-regularization: between $0$ and $5^{-10}$. vi) Width/length/depth/size of kernels: we checked both networks with few layers and deep networks such as ResNet50 and VGG19, which also included various types of layers. vii) Initialization: Xavier. viii) Activations: Relu, Elu, Linear.
In addition, we also added experiments which change directly only specific hyper-parameters, on a toy problem of training the small mammals dataset on st-VGG. All hyper-parameter changes resulted in similar qualitative results, and included: i) Activations: Relu, Elu, Linear, tanh. ii) Initializations: Xavier, He normal, LeCun normal, truncated normal. iii) Batch size: between $1$ and $2500$. iv) Learning rate: between $10^{-3}$ to $0.1$. v) Dropout: $0$ to $0.5$.

\begin{figure*}[hbt]
a)
\begin{subfigure}{.135\textwidth}
  \centering
  \includegraphics[width=1\linewidth]{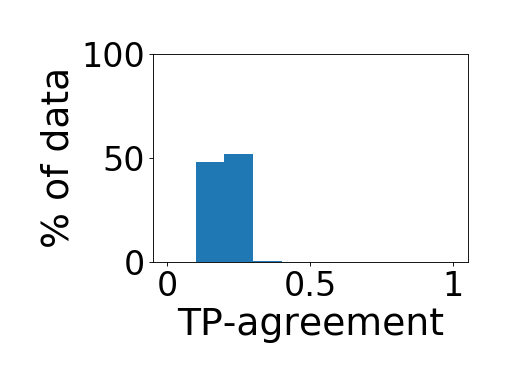}
\end{subfigure}
\begin{subfigure}{.135\textwidth}
  \centering
  \includegraphics[width=1\linewidth]{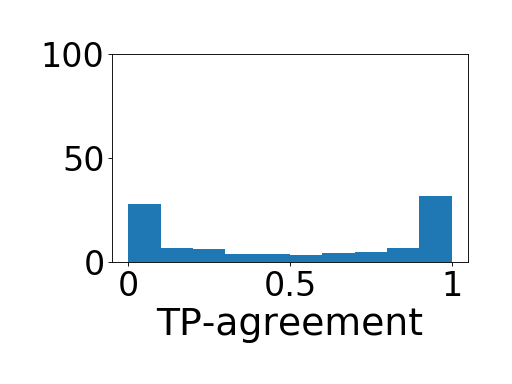}
\end{subfigure}
\begin{subfigure}{.135\textwidth}
  \centering
  \includegraphics[width=1\linewidth]{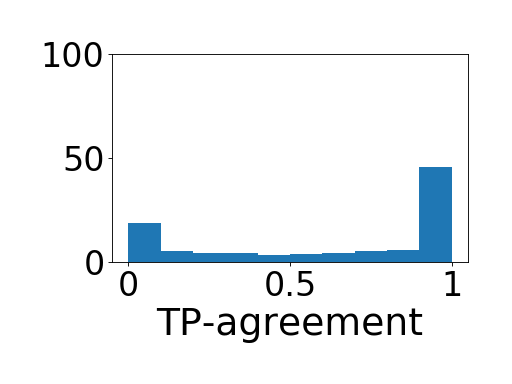}
\end{subfigure}
\begin{subfigure}{.135\textwidth}
  \centering
  \includegraphics[width=1\linewidth]{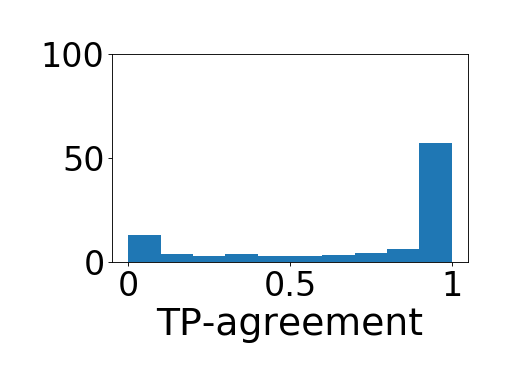}
\end{subfigure}
\begin{subfigure}{.135\textwidth}
  \centering
  \includegraphics[width=1\linewidth]{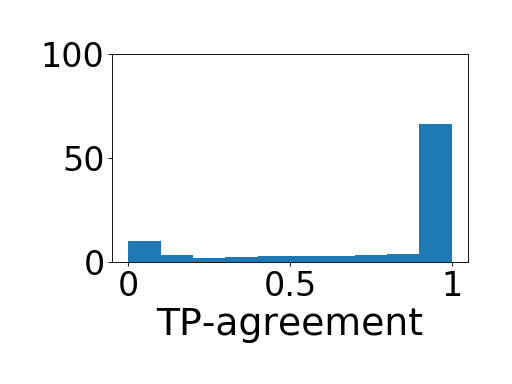}
\end{subfigure}
\begin{subfigure}{.135\textwidth}
  \centering
  \includegraphics[width=1\linewidth]{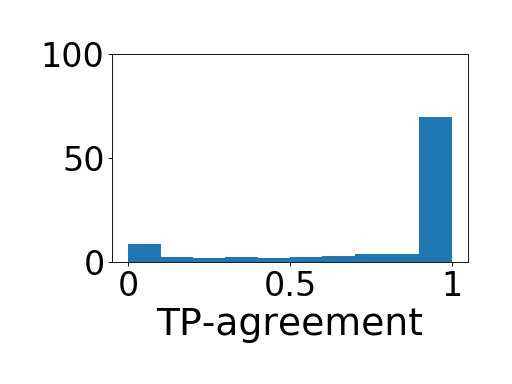}
\end{subfigure}
\begin{subfigure}{.135\textwidth}
  \centering
  \includegraphics[width=1\linewidth]{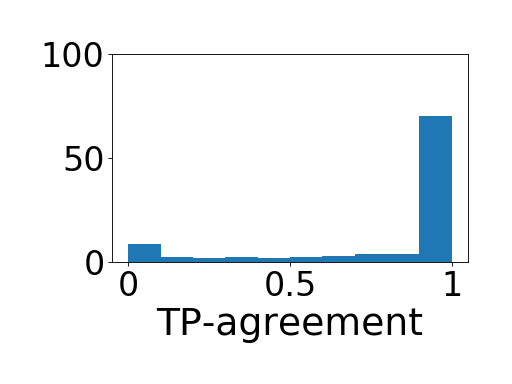}
\end{subfigure}

b)
\begin{subfigure}{.135\textwidth}
  \centering
  \includegraphics[width=1\linewidth]{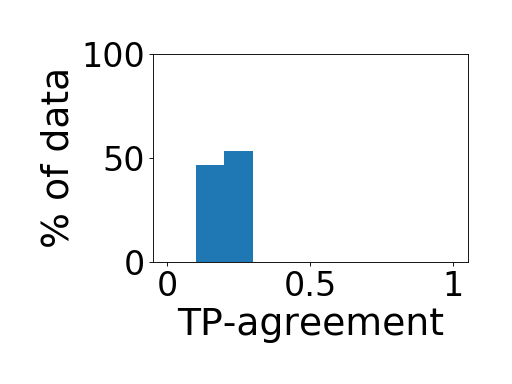}
\end{subfigure}
\begin{subfigure}{.135\textwidth}
  \centering
  \includegraphics[width=1\linewidth]{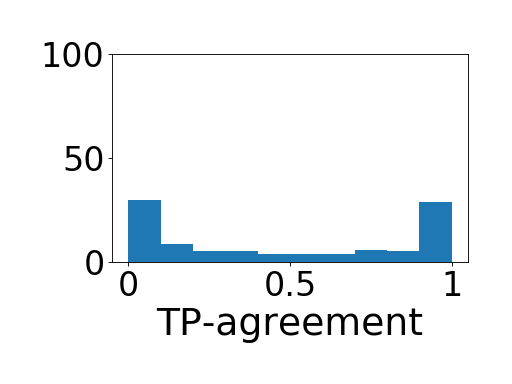}
\end{subfigure}
\begin{subfigure}{.135\textwidth}
  \centering
  \includegraphics[width=1\linewidth]{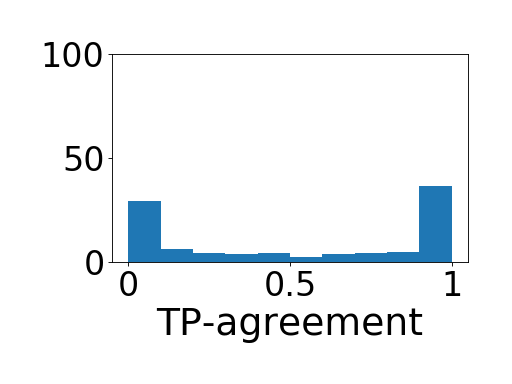}
\end{subfigure}
\begin{subfigure}{.135\textwidth}
  \centering
  \includegraphics[width=1\linewidth]{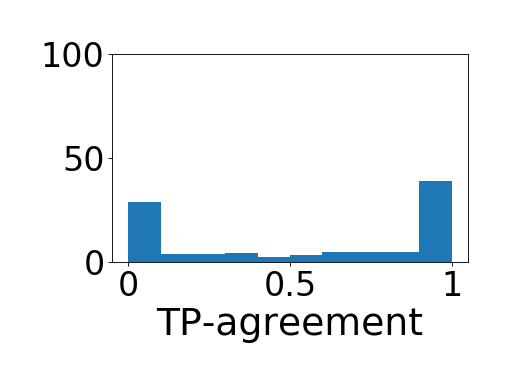}
\end{subfigure}
\begin{subfigure}{.135\textwidth}
  \centering
  \includegraphics[width=1\linewidth]{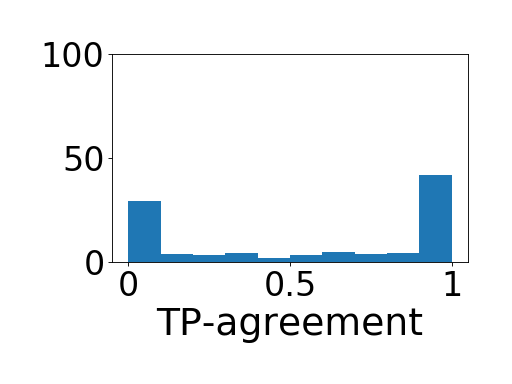}
\end{subfigure}
\begin{subfigure}{.135\textwidth}
  \centering
  \includegraphics[width=1\linewidth]{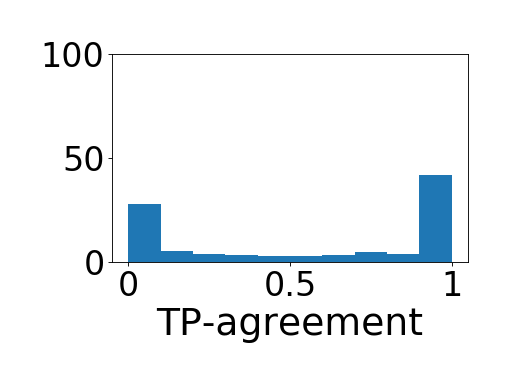}
\end{subfigure}
\begin{subfigure}{.135\textwidth}
  \centering
  \includegraphics[width=1\linewidth]{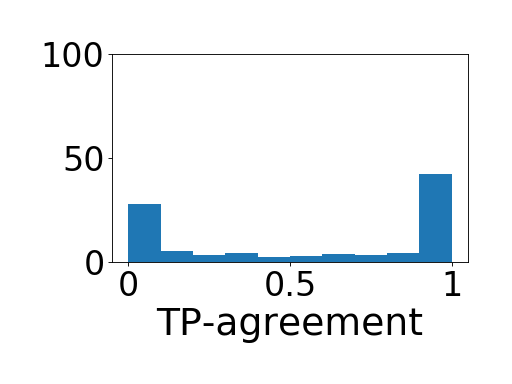}
\end{subfigure}
\caption{The distribution of TP-agreement scores during the learning process of $100$ instances of small st-VGG (see \app~\ref{app:architectures}) trained on the small-mammals dataset (see \app~\ref{app:datasets}). Epochs shown: $0, 10, 30, 60, 90, 120, 140$. a) Train set; b) Test set.}
  \label{appendix:full_results_mediumnet_subset16}
\end{figure*}

\begin{figure*}[hbt]
a)
\begin{subfigure}{.135\textwidth}
  \centering
  \includegraphics[width=1\linewidth]{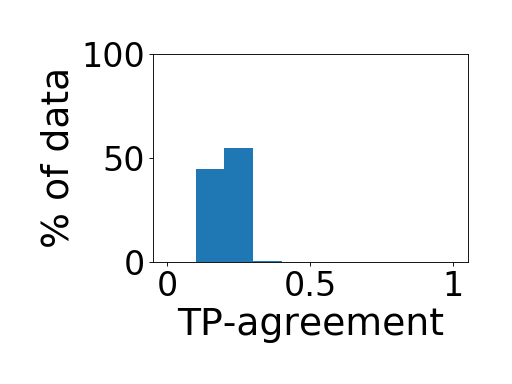}
\end{subfigure}
\begin{subfigure}{.135\textwidth}
  \centering
  \includegraphics[width=1\linewidth]{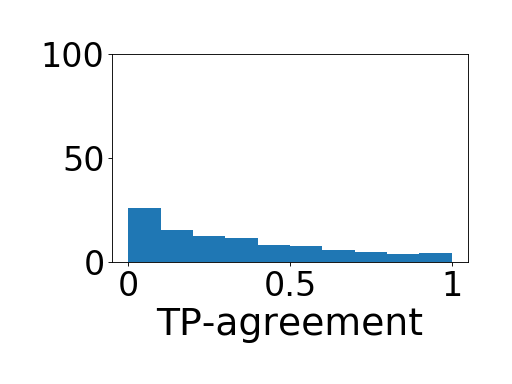}
\end{subfigure}
\begin{subfigure}{.135\textwidth}
  \centering
  \includegraphics[width=1\linewidth]{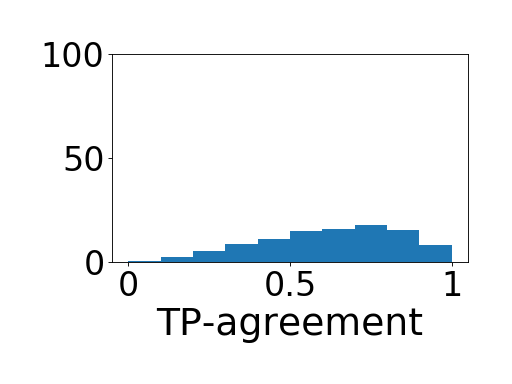}
\end{subfigure}
\begin{subfigure}{.135\textwidth}
  \centering
  \includegraphics[width=1\linewidth]{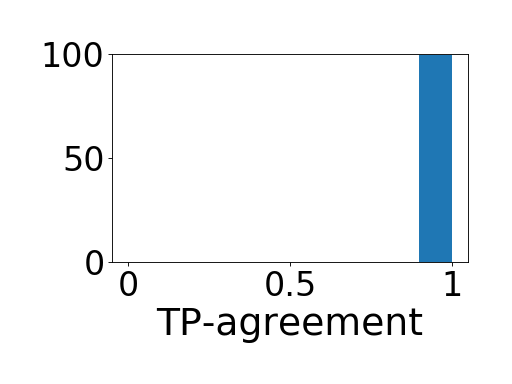}
\end{subfigure}
\begin{subfigure}{.135\textwidth}
  \centering
  \includegraphics[width=1\linewidth]{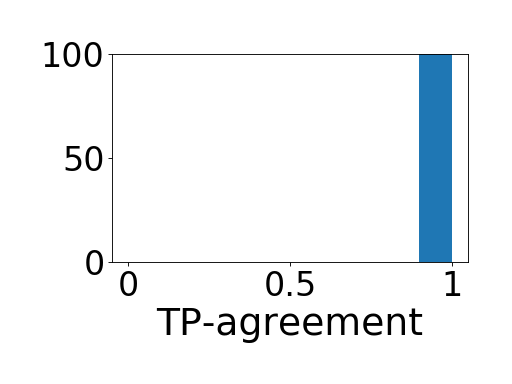}
\end{subfigure}
\begin{subfigure}{.135\textwidth}
  \centering
  \includegraphics[width=1\linewidth]{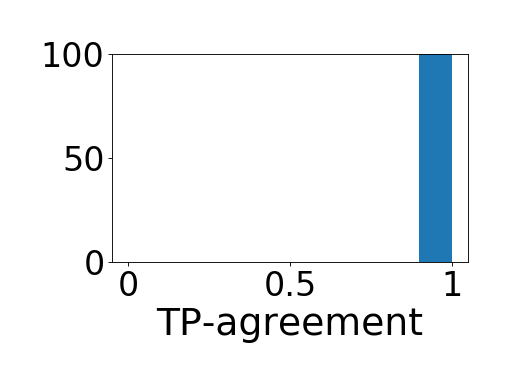}
\end{subfigure}
\begin{subfigure}{.135\textwidth}
  \centering
  \includegraphics[width=1\linewidth]{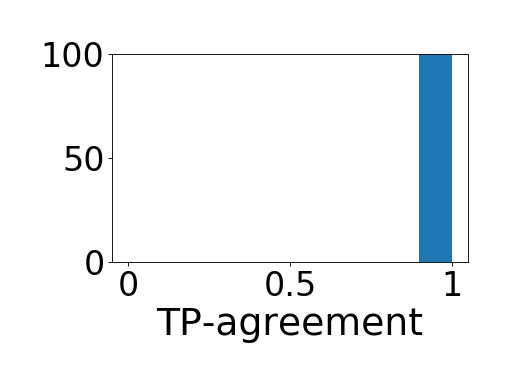}
\end{subfigure}

b)
\begin{subfigure}{.135\textwidth}
  \centering
  \includegraphics[width=1\linewidth]{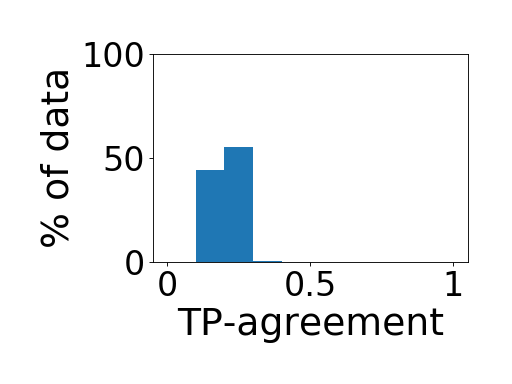}
\end{subfigure}
\begin{subfigure}{.135\textwidth}
  \centering
  \includegraphics[width=1\linewidth]{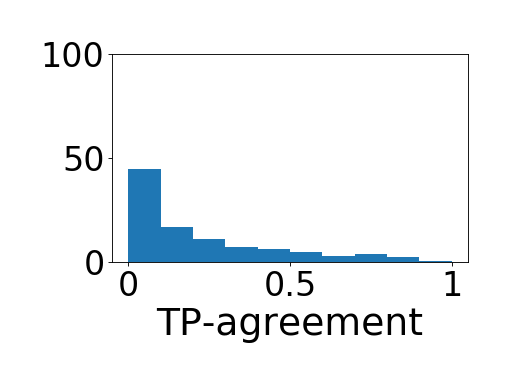}
\end{subfigure}
\begin{subfigure}{.135\textwidth}
  \centering
  \includegraphics[width=1\linewidth]{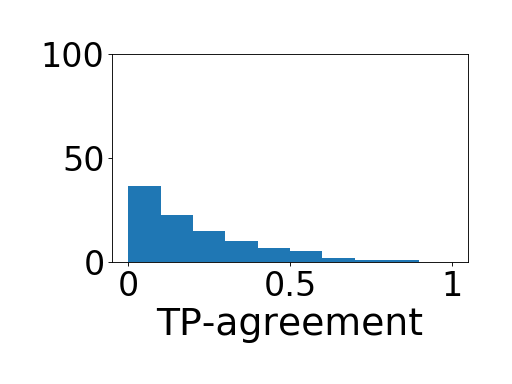}
\end{subfigure}
\begin{subfigure}{.135\textwidth}
  \centering
  \includegraphics[width=1\linewidth]{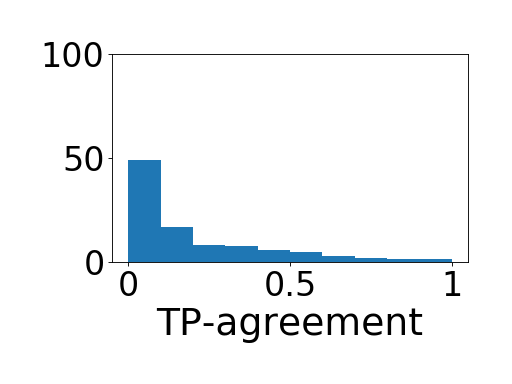}
\end{subfigure}
\begin{subfigure}{.135\textwidth}
  \centering
  \includegraphics[width=1\linewidth]{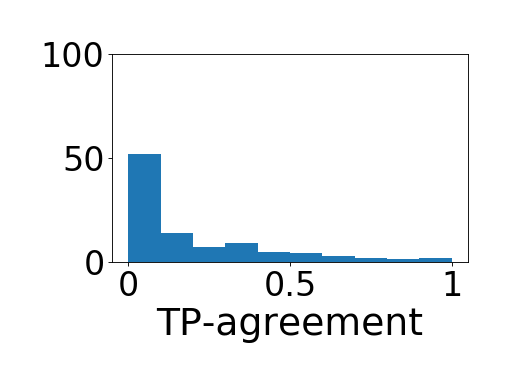}
\end{subfigure}
\begin{subfigure}{.135\textwidth}
  \centering
  \includegraphics[width=1\linewidth]{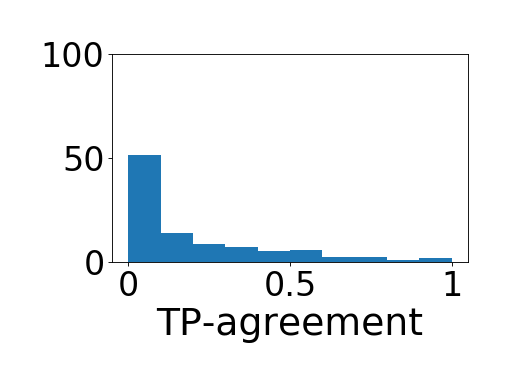}
\end{subfigure}
\begin{subfigure}{.135\textwidth}
  \centering
  \includegraphics[width=1\linewidth]{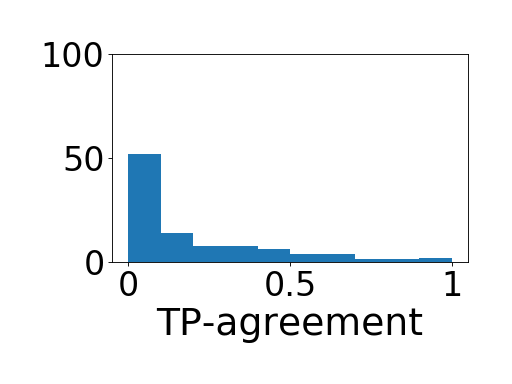}
\end{subfigure}
\caption{The distribution of TP-agreement scores during the learning process of $100$ instances of st-VGG (see \app~\ref{app:architectures}) trained on the small-mammals dataset (see \app~\ref{app:datasets}), where labels were assigned randomly to images, as done in \citet{zhang2016understanding}. Epochs shown: $0, 10, 30, 60, 90, 120, 140$. a) Train set; b) Test set.}
  \label{appendix:full_results_random_labels_subset16}
\end{figure*}

\begin{figure*}[hbt]
a)
\begin{subfigure}{.135\textwidth}
  \centering
  \includegraphics[width=1\linewidth]{linearnet-subset16/histogram_consistency_train_epoch0.png}
\end{subfigure}
\begin{subfigure}{.135\textwidth}
  \centering
  \includegraphics[width=1\linewidth]{linearnet-subset16/histogram_consistency_train_epoch10.png}
\end{subfigure}
\begin{subfigure}{.135\textwidth}
  \centering
  \includegraphics[width=1\linewidth]{linearnet-subset16/histogram_consistency_train_epoch30.png}
\end{subfigure}
\begin{subfigure}{.135\textwidth}
  \centering
  \includegraphics[width=1\linewidth]{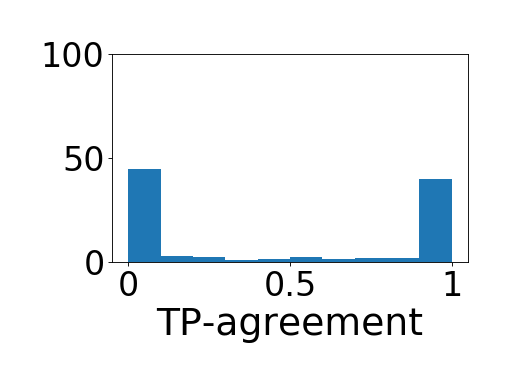}
\end{subfigure}
\begin{subfigure}{.135\textwidth}
  \centering
  \includegraphics[width=1\linewidth]{linearnet-subset16/histogram_consistency_train_epoch90.png}
\end{subfigure}
\begin{subfigure}{.135\textwidth}
  \centering
  \includegraphics[width=1\linewidth]{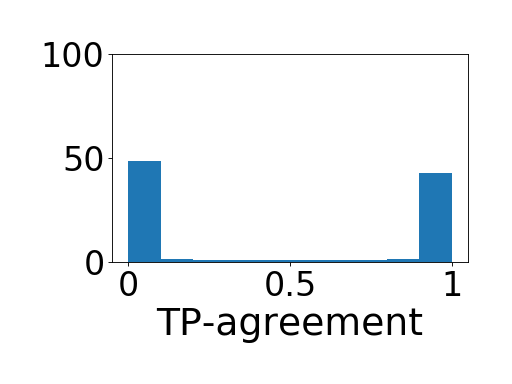}
\end{subfigure}
\begin{subfigure}{.135\textwidth}
  \centering
  \includegraphics[width=1\linewidth]{linearnet-subset16/histogram_consistency_train_epoch140.png}
\end{subfigure}

b)
\begin{subfigure}{.135\textwidth}
  \centering
  \includegraphics[width=1\linewidth]{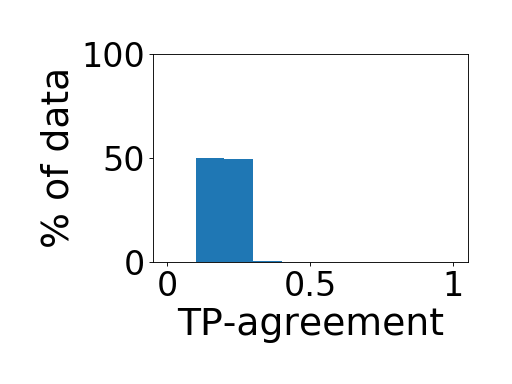}
\end{subfigure}
\begin{subfigure}{.135\textwidth}
  \centering
  \includegraphics[width=1\linewidth]{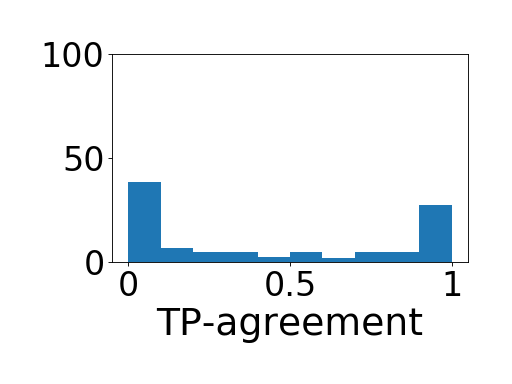}
\end{subfigure}
\begin{subfigure}{.135\textwidth}
  \centering
  \includegraphics[width=1\linewidth]{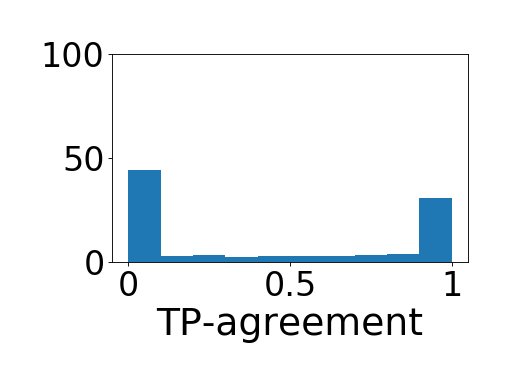}
\end{subfigure}
\begin{subfigure}{.135\textwidth}
  \centering
  \includegraphics[width=1\linewidth]{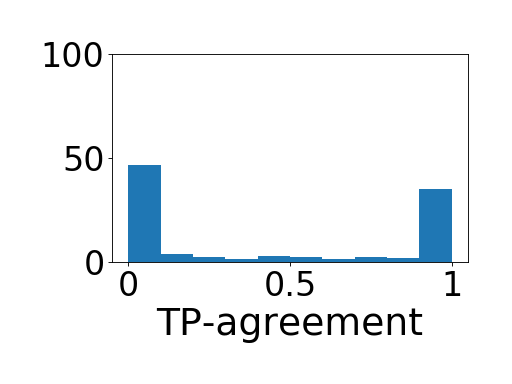}
\end{subfigure}
\begin{subfigure}{.135\textwidth}
  \centering
  \includegraphics[width=1\linewidth]{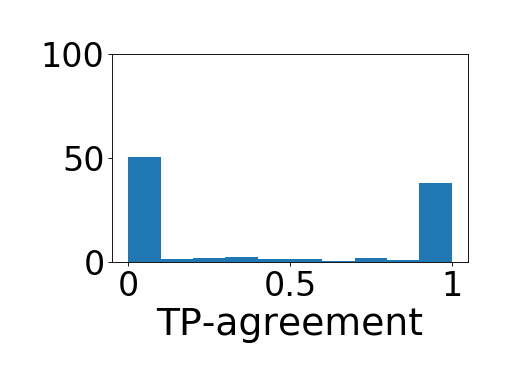}
\end{subfigure}
\begin{subfigure}{.135\textwidth}
  \centering
  \includegraphics[width=1\linewidth]{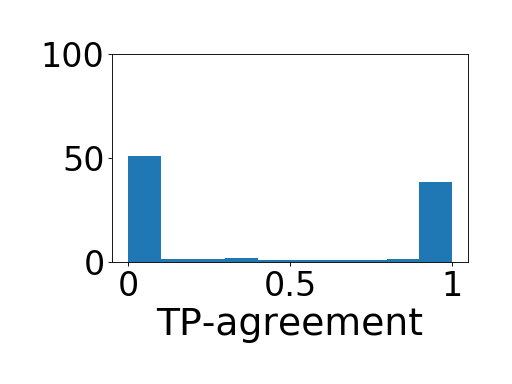}
\end{subfigure}
\begin{subfigure}{.135\textwidth}
  \centering
  \includegraphics[width=1\linewidth]{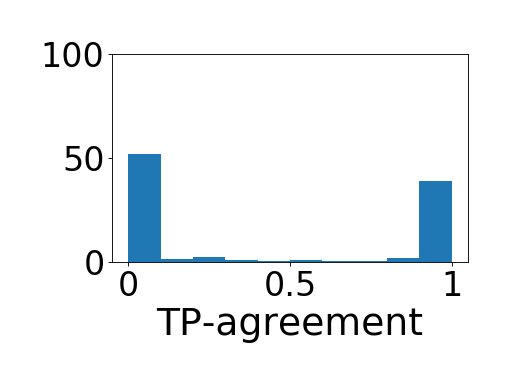}
\end{subfigure}
\caption{The distribution of TP-agreement scores during the learning process of $100$ instances of linear st-VGG (see \app~\ref{app:architectures}) trained on the small-mammals dataset (see \app~\ref{app:datasets}). Epochs shown: $0, 10, 30, 60, 90, 120, 140$. a) Train set; b) Test set.}
  \label{appendix:full_results_linearnet-subset16}
\end{figure*}

\begin{figure*}[hbt]
a)
\begin{subfigure}{.135\textwidth}
  \centering
  \includegraphics[width=1\linewidth]{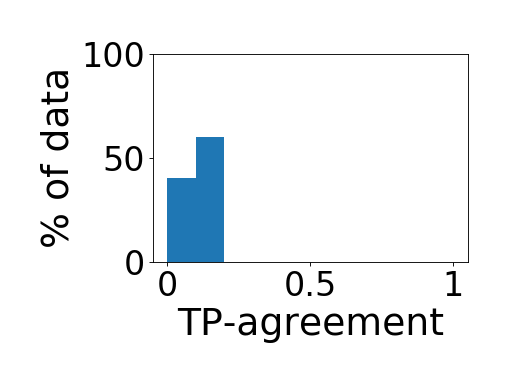}
\end{subfigure}
\begin{subfigure}{.135\textwidth}
  \centering
  \includegraphics[width=1\linewidth]{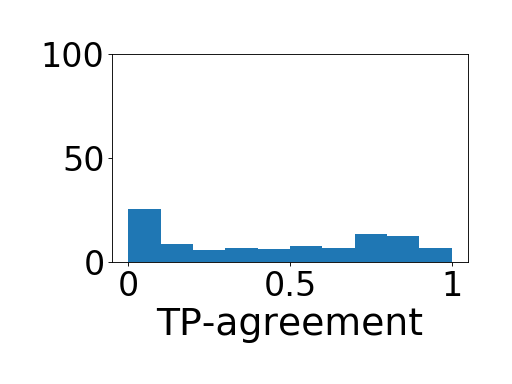}
\end{subfigure}
\begin{subfigure}{.135\textwidth}
  \centering
  \includegraphics[width=1\linewidth]{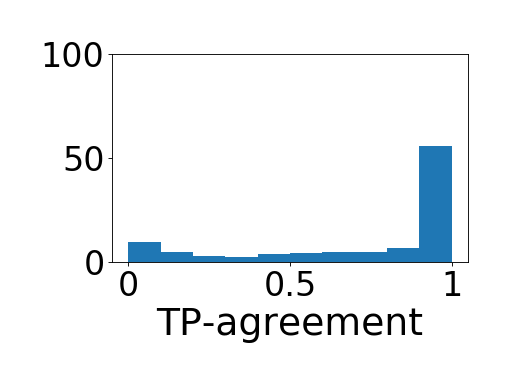}
\end{subfigure}
\begin{subfigure}{.135\textwidth}
  \centering
  \includegraphics[width=1\linewidth]{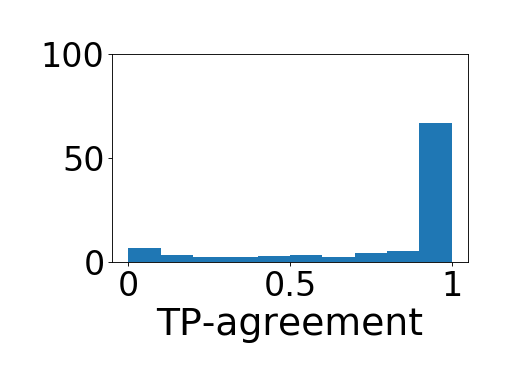}
\end{subfigure}
\begin{subfigure}{.135\textwidth}
  \centering
  \includegraphics[width=1\linewidth]{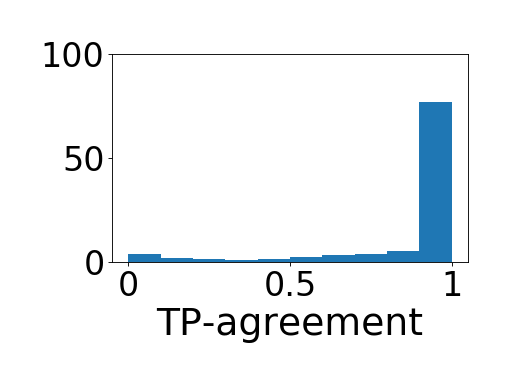}
\end{subfigure}
\begin{subfigure}{.135\textwidth}
  \centering
  \includegraphics[width=1\linewidth]{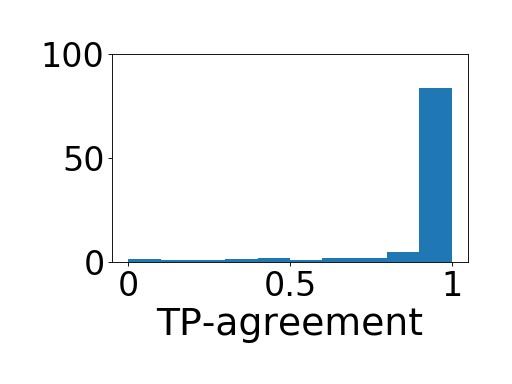}
\end{subfigure}
\begin{subfigure}{.135\textwidth}
  \centering
  \includegraphics[width=1\linewidth]{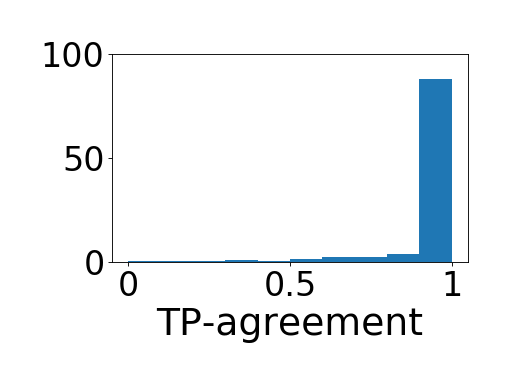}
\end{subfigure}

b)
\begin{subfigure}{.135\textwidth}
  \centering
  \includegraphics[width=1\linewidth]{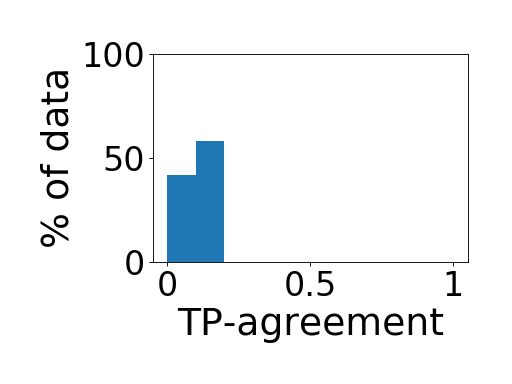}
\end{subfigure}
\begin{subfigure}{.135\textwidth}
  \centering
  \includegraphics[width=1\linewidth]{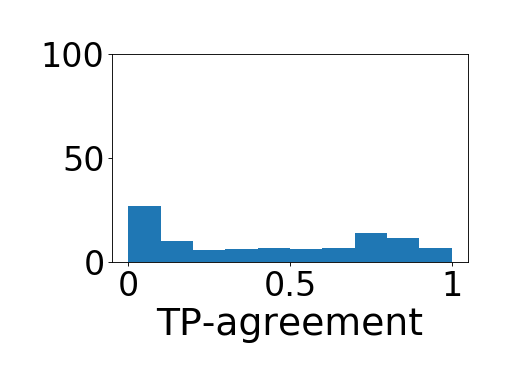}
\end{subfigure}
\begin{subfigure}{.135\textwidth}
  \centering
  \includegraphics[width=1\linewidth]{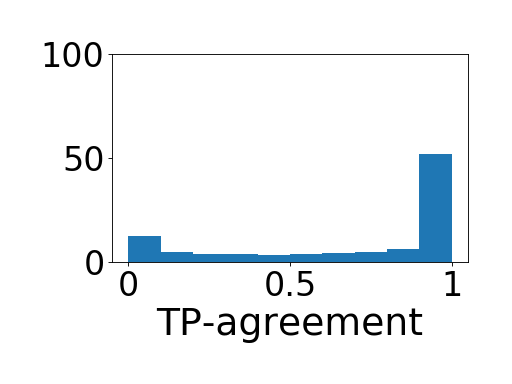}
\end{subfigure}
\begin{subfigure}{.135\textwidth}
  \centering
  \includegraphics[width=1\linewidth]{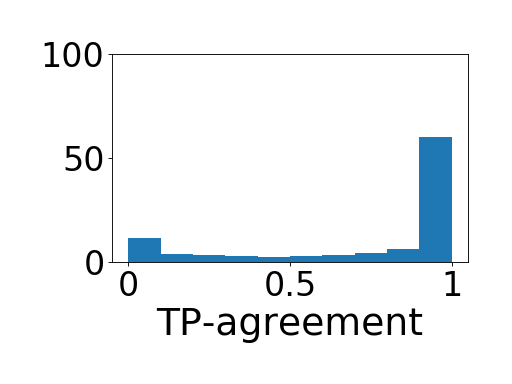}
\end{subfigure}
\begin{subfigure}{.135\textwidth}
  \centering
  \includegraphics[width=1\linewidth]{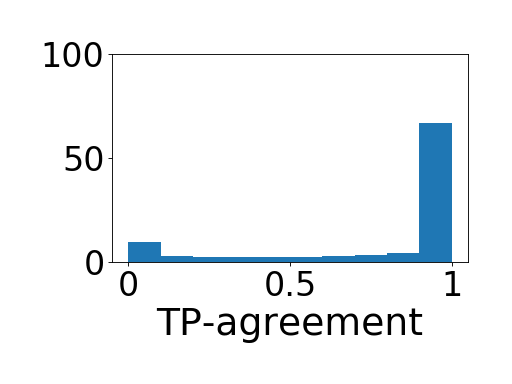}
\end{subfigure}
\begin{subfigure}{.135\textwidth}
  \centering
  \includegraphics[width=1\linewidth]{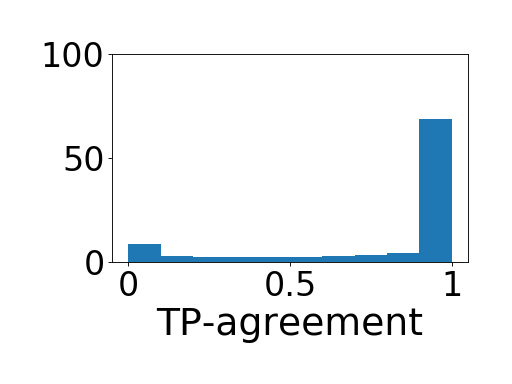}
\end{subfigure}
\begin{subfigure}{.135\textwidth}
  \centering
  \includegraphics[width=1\linewidth]{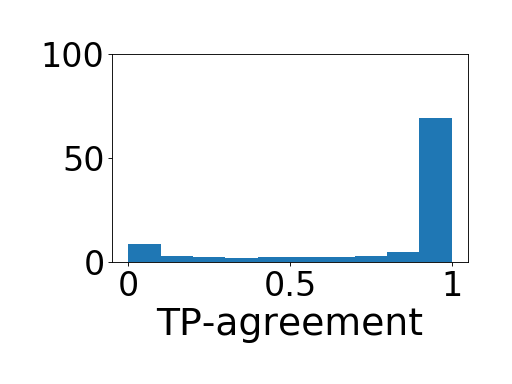}
\end{subfigure}

c)
\begin{subfigure}{.135\textwidth}
  \centering
  \includegraphics[width=1\linewidth]{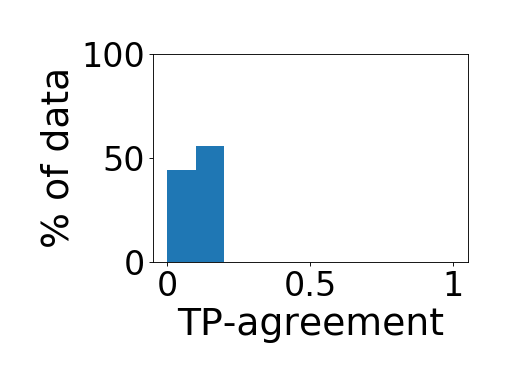}
\end{subfigure}
\begin{subfigure}{.135\textwidth}
  \centering
  \includegraphics[width=1\linewidth]{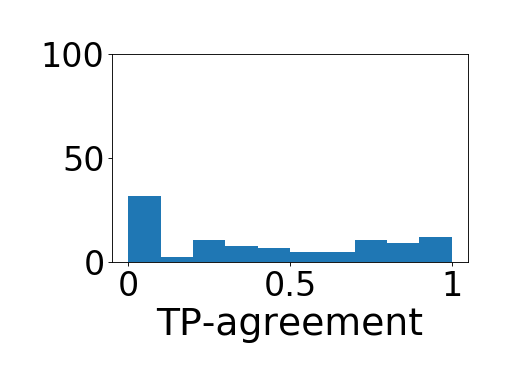}
\end{subfigure}
\begin{subfigure}{.135\textwidth}
  \centering
  \includegraphics[width=1\linewidth]{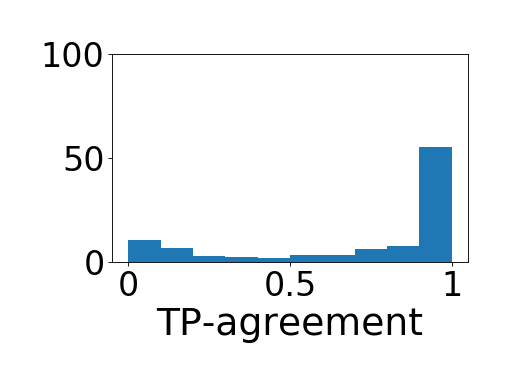}
\end{subfigure}
\begin{subfigure}{.135\textwidth}
  \centering
  \includegraphics[width=1\linewidth]{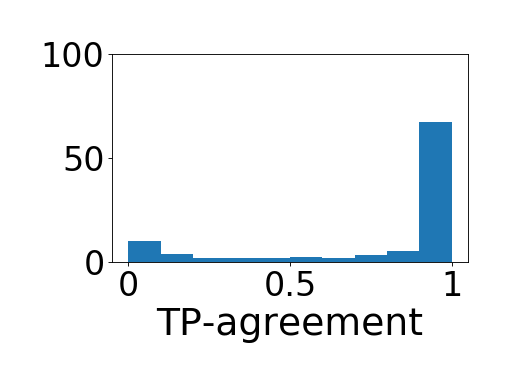}
\end{subfigure}
\begin{subfigure}{.135\textwidth}
  \centering
  \includegraphics[width=1\linewidth]{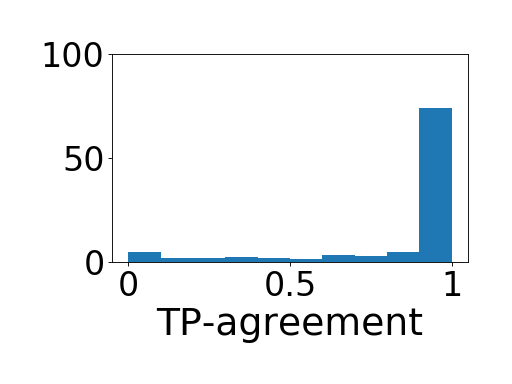}
\end{subfigure}
\begin{subfigure}{.135\textwidth}
  \centering
  \includegraphics[width=1\linewidth]{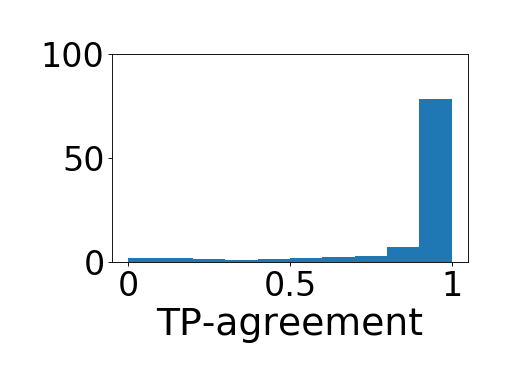}
\end{subfigure}
\begin{subfigure}{.135\textwidth}
  \centering
  \includegraphics[width=1\linewidth]{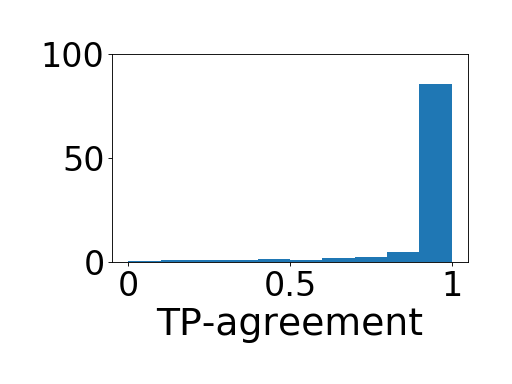}
\end{subfigure}

d)
\begin{subfigure}{.135\textwidth}
  \centering
  \includegraphics[width=1\linewidth]{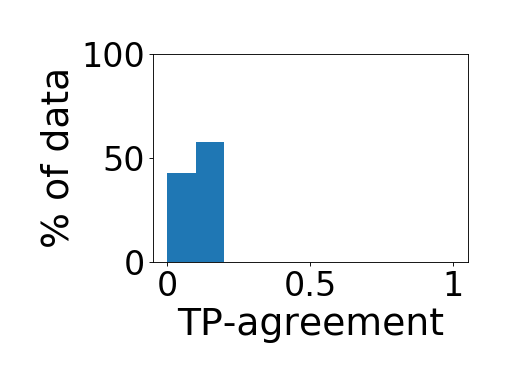}
\end{subfigure}
\begin{subfigure}{.135\textwidth}
  \centering
  \includegraphics[width=1\linewidth]{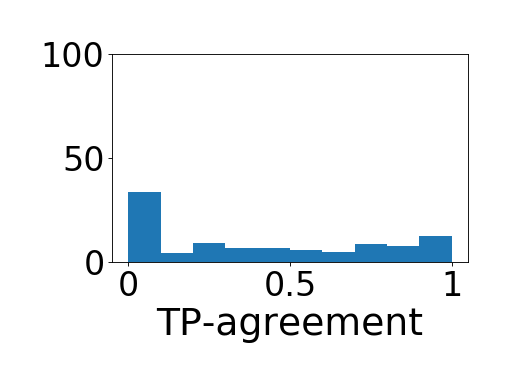}
\end{subfigure}
\begin{subfigure}{.135\textwidth}
  \centering
  \includegraphics[width=1\linewidth]{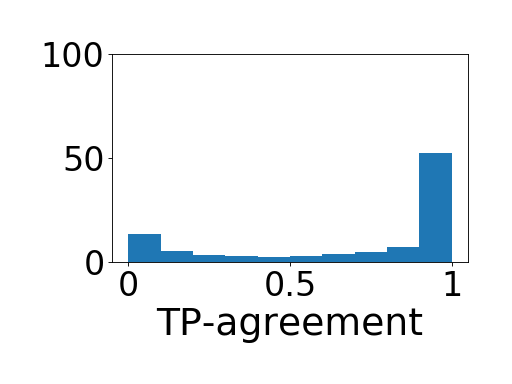}
\end{subfigure}
\begin{subfigure}{.135\textwidth}
  \centering
  \includegraphics[width=1\linewidth]{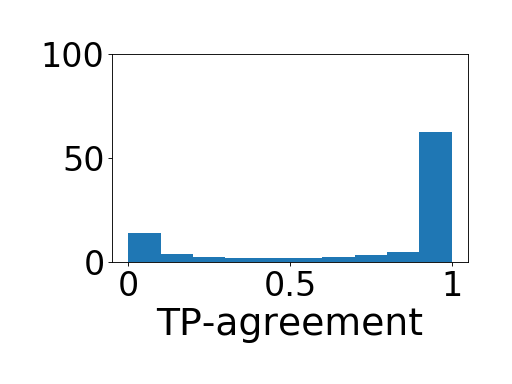}
\end{subfigure}
\begin{subfigure}{.135\textwidth}
  \centering
  \includegraphics[width=1\linewidth]{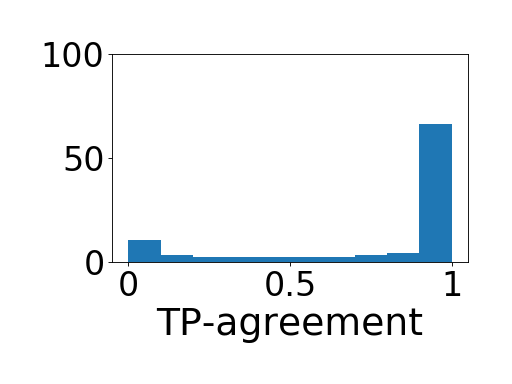}
\end{subfigure}
\begin{subfigure}{.135\textwidth}
  \centering
  \includegraphics[width=1\linewidth]{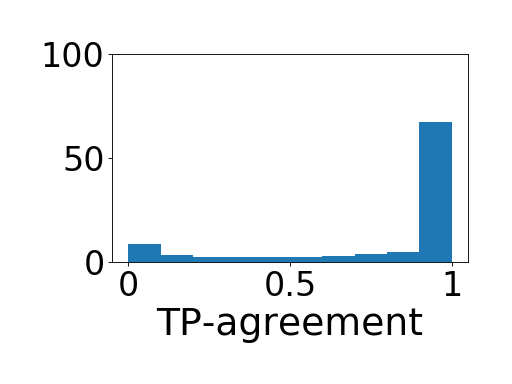}
\end{subfigure}
\begin{subfigure}{.135\textwidth}
  \centering
  \includegraphics[width=1\linewidth]{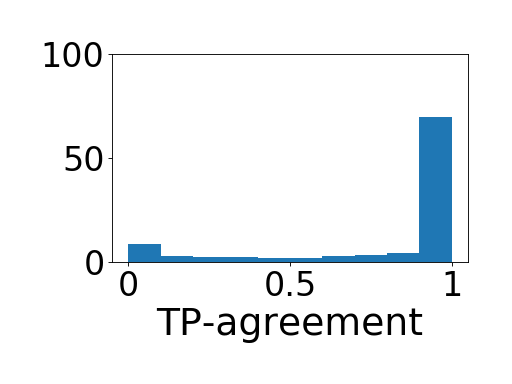}
\end{subfigure}

e)
\begin{subfigure}{.135\textwidth}
  \centering
  \includegraphics[width=1\linewidth]{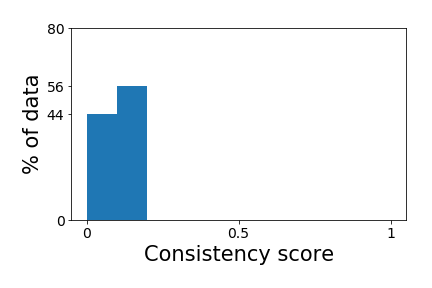}
\end{subfigure}
\begin{subfigure}{.135\textwidth}
  \centering
  \includegraphics[width=1\linewidth]{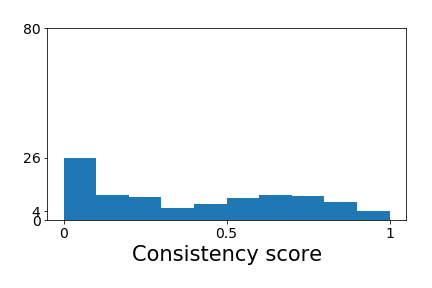}
\end{subfigure}
\begin{subfigure}{.135\textwidth}
  \centering
  \includegraphics[width=1\linewidth]{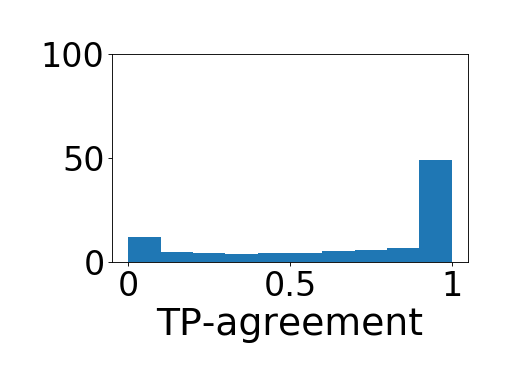}
\end{subfigure}
\begin{subfigure}{.135\textwidth}
  \centering
  \includegraphics[width=1\linewidth]{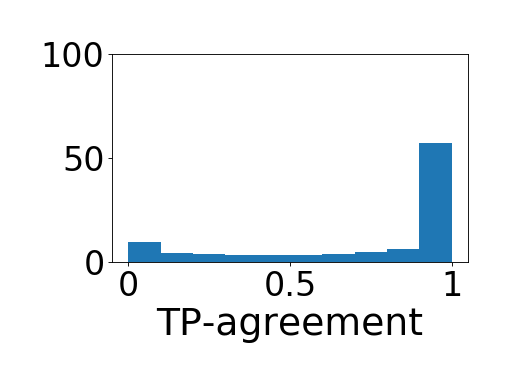}
\end{subfigure}
\begin{subfigure}{.135\textwidth}
  \centering
  \includegraphics[width=1\linewidth]{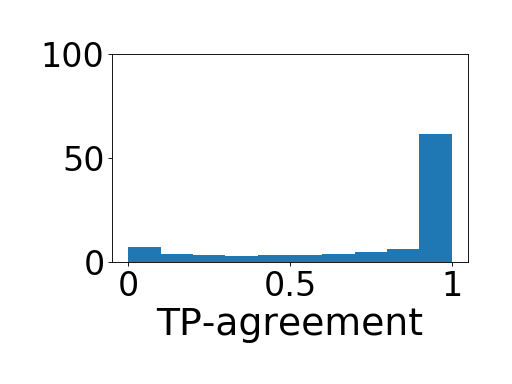}
\end{subfigure}
\begin{subfigure}{.135\textwidth}
  \centering
  \includegraphics[width=1\linewidth]{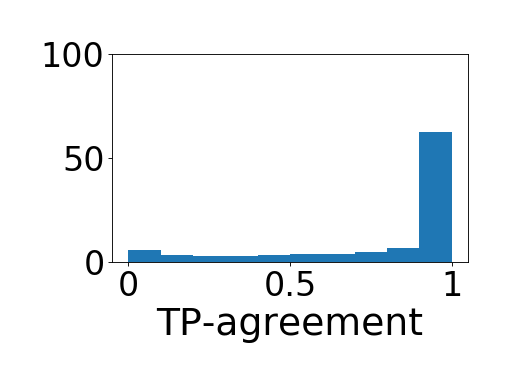}
\end{subfigure}
\begin{subfigure}{.135\textwidth}
  \centering
  \includegraphics[width=1\linewidth]{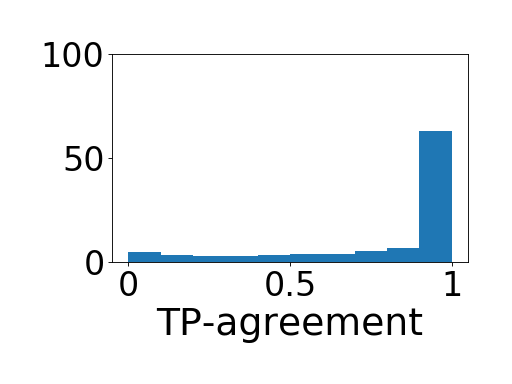}
\end{subfigure}
\caption{The distribution of TP-agreement scores during the learning process of $100$ instances of st-VGG (see \app~\ref{app:architectures}) trained on the parts of Fashion-Mnist. We divided the train set of Fashion-Mnist into 60 parts of $1000$ images. Epochs shown: $0, 1, 5, 10, 20, 30, 40$. a-b) Train and test sets of $100$ instances trained on a random part of Fashion-Mnist; c-d) train and test sets of $100$ instances trained on another random part of Fashion-Mnist. e) The average distribution of all 60 collections on the Fashion-Mnist test set. The bi-modality presented here indicates that although learned on different training sets, all collections have similar learning order.}
  \label{appendix:full_results_split_fashion_mnist}
\end{figure*}

\begin{figure*}[hbt]
a)
\begin{subfigure}{.24\textwidth}
  \centering
  \includegraphics[width=1\linewidth]{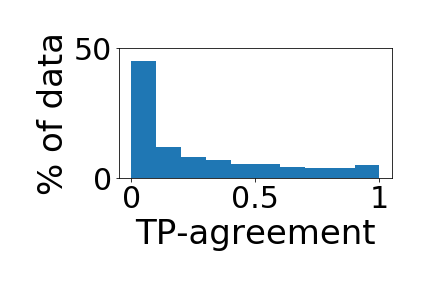}
\end{subfigure}
\begin{subfigure}{.24\textwidth}
  \centering
  \includegraphics[width=1\linewidth]{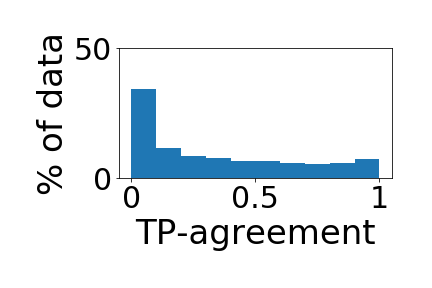}
\end{subfigure}
\begin{subfigure}{.24\textwidth}
  \centering
  \includegraphics[width=1\linewidth]{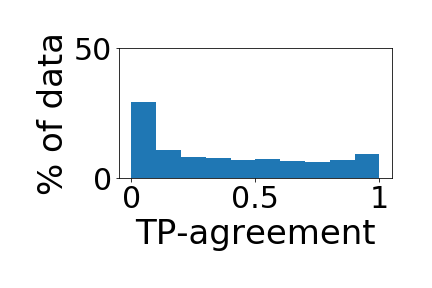}
\end{subfigure}
\begin{subfigure}{.24\textwidth}
  \centering
  \includegraphics[width=1\linewidth]{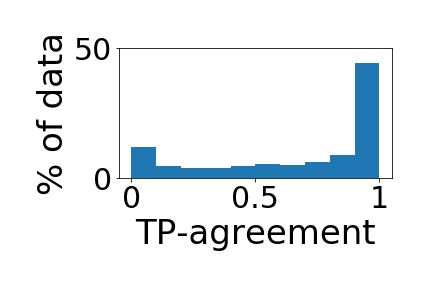}
\end{subfigure}

b)
\begin{subfigure}{.24\textwidth}
  \centering
  \includegraphics[width=1\linewidth]{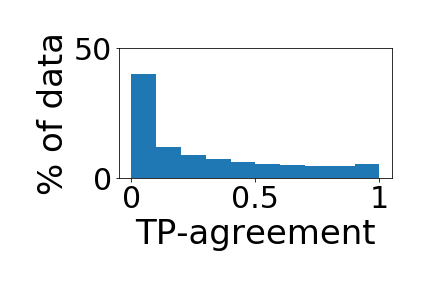}
\end{subfigure}
\begin{subfigure}{.24\textwidth}
  \centering
  \includegraphics[width=1\linewidth]{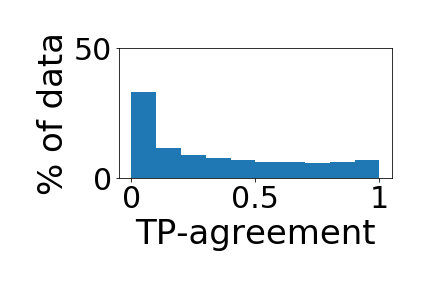}
\end{subfigure}
\begin{subfigure}{.24\textwidth}
  \centering
  \includegraphics[width=1\linewidth]{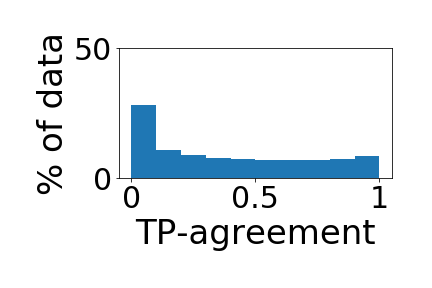}
\end{subfigure}
\begin{subfigure}{.24\textwidth}
  \centering
  \includegraphics[width=1\linewidth]{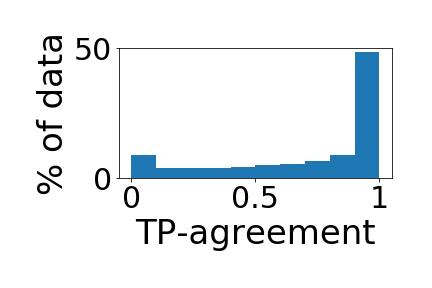}
\end{subfigure}

c)
\begin{subfigure}{.24\textwidth}
  \centering
  \includegraphics[width=1\linewidth]{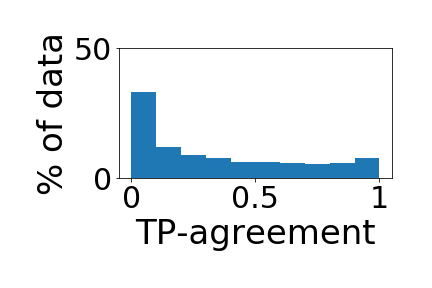}
\end{subfigure}
\begin{subfigure}{.24\textwidth}
  \centering
  \includegraphics[width=1\linewidth]{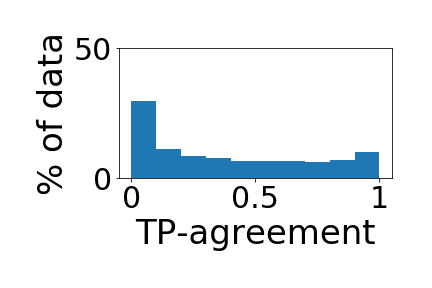}
\end{subfigure}
\begin{subfigure}{.24\textwidth}
  \centering
  \includegraphics[width=1\linewidth]{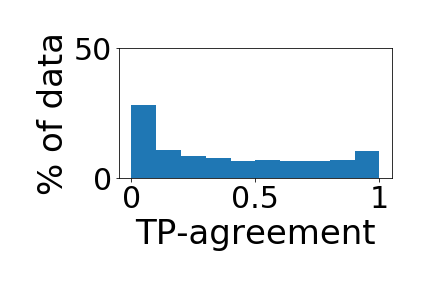}
\end{subfigure}
\begin{subfigure}{.24\textwidth}
  \centering
  \includegraphics[width=1\linewidth]{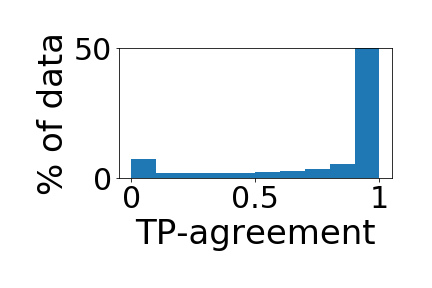}
\end{subfigure}
\caption{Averaged TP-agreement dynamics of pairs of collections of ImageNet architectures. a)  6 DenseNet instances with 22 AlexNet instances, plotted on accuracies of $26\%, 33\%, 38\%, 67\%$. b) 27 ResNet-50 instances with 22 AlexNet instances, plotted on accuracies of $28\%, 33\%, 38\%, 70\%$. c) 27 ResNet-50 instances with 6 DenseNet, plotted on accuracies of $35\%, 39\%, 40\%, 83\%$. These results follow the same protocol described in Fig.~\ref{fig:alexnet_vs_resnet}.}
  \label{appendix:full_results_imagenet_dynamics}
\end{figure*}

\begin{figure*}[hbt]
\begin{subfigure}{.49\textwidth}
  \centering
  \includegraphics[width=1\linewidth]{imagenet_pair_correlations/resnet_vs_densenet_r97_p0.png}
\end{subfigure}
\begin{subfigure}{.49\textwidth}
  \centering
  \includegraphics[width=1\linewidth]{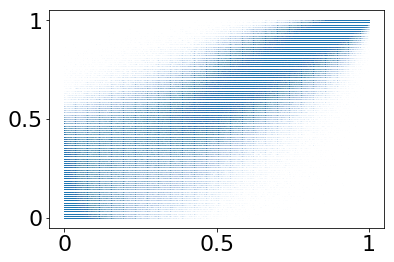}
\end{subfigure}
\caption{The \emph{accessibility score} in 2 pairs of collections of different architectures trained on ImageNet. Left: $27$ instances of ResNet-50 and $6$ instances of DenseNet, $r=0.97, p<10^{-50}$. Right: 22 instances of AlexNet and 6 instances of DenseNet $r=0.87, p<10^{-50}$. These results follow the same protocol described in Fig.~\ref{fig:alexnet_vs_resnet}.}
  \label{appendix:full_results_imagenet_correlations}
\end{figure*}

\begin{figure*}[hbt]
\begin{subfigure}{.33\textwidth}
  \centering
  \includegraphics[width=1\linewidth]{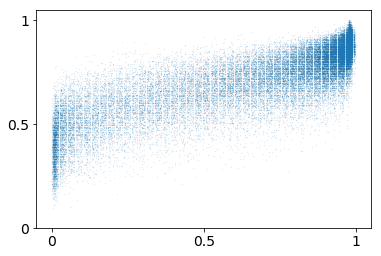}
  \caption{CIFAR-10}
\end{subfigure}
\begin{subfigure}{.33\textwidth}
  \centering
  \includegraphics[width=1\linewidth]{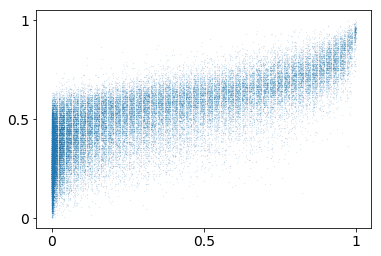}
  \caption{CIFAR-100}
\end{subfigure}
\begin{subfigure}{.33\textwidth}
  \centering
  \includegraphics[width=1\linewidth]{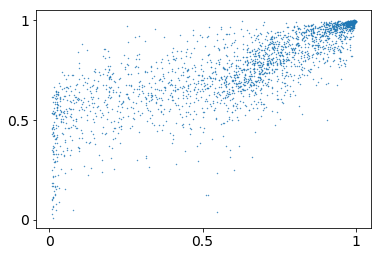}
  \caption{Small mammals}
\end{subfigure}
\caption{
The \emph{accessibility score} of different architectures train on the same dataset. a) $100$ instances of st-VGG and $19$ instances of VGG19 trained on CIFAR-10. $r=0.83, p<10^{-50}$. b) $100$ instances of st-VGG and $20$ instances of VGG19 trained on CIFAR-100. $r=0.78, p<10^{-50}$. c) $100$ instances of st-VGG and $100$ instances of small st-VGG on the samll-mammals dataset $r=0.82, p<10^{-50}$. These results follow the same protocol described in Fig.~\ref{fig:alexnet_vs_resnet}.}
  \label{appendix:full_results_correlations_comparison}
\end{figure*}

\section{Additional results}
\label{app:additional_results}
\begin{figure*}[htb]
\begin{subfigure}{.245\textwidth}
  \centering
  \includegraphics[width=1\linewidth]{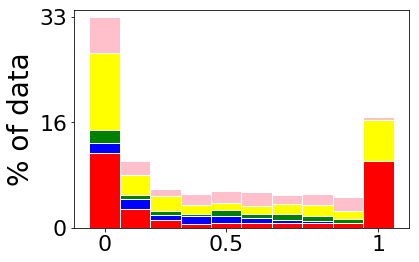}
  \caption{Epoch 1}
\end{subfigure}
\begin{subfigure}{.245\textwidth}
  \centering
  \includegraphics[width=1\linewidth]{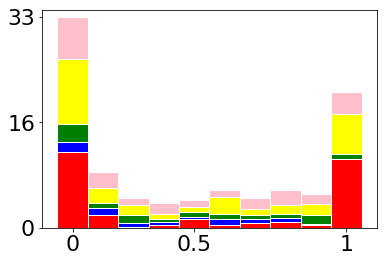}
  \caption{Epoch 2}
\end{subfigure}
\begin{subfigure}{.245\textwidth}
  \centering
  \includegraphics[width=1\linewidth]{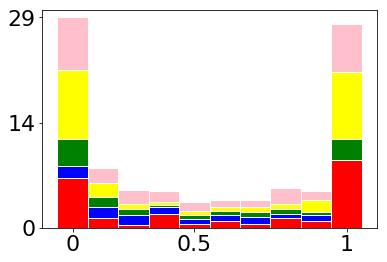}
  \caption{Epoch 30}
\end{subfigure}
\begin{subfigure}{.245\textwidth}
  \centering
  \includegraphics[width=1\linewidth]{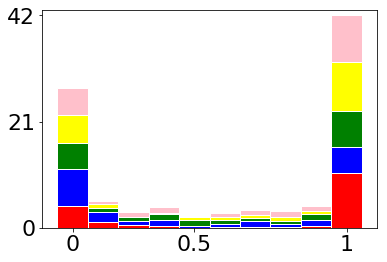}
  \caption{Epoch 140}
\end{subfigure}

\caption{}
\label{fig:class_hierarchy}
\end{figure*}

\paragraph{Induced class hierarchy.}
The ranking of training examples induced by the TP-agreement typically induces a hierarchical structure over the different classes as well. To see this, we train $100$ instances of st-VGG on the small-mammals dataset, and calculate for each image the most frequent class label assigned to it by the collection of networks. In Fig.~\ref{fig:class_hierarchy} we plot the histogram of the TP-agreement (as in Fig.~\ref{fig:consistency_dynamics}), but this time each image is assigned a color, which identifies its most frequent class label (1 of 5 colors). It can be readily seen that at the beginning of learning, only images from 2 classes reach a TP-agreement of 1. As learning proceeds, more class labels slowly emerge. This result suggests that classes are learned in a specific order, across all networks. Moreover, we can see a pattern in the erroneous label assignments, which suggests that the classifiers initially use fewer class labels, and only become more specific later on in the learning process.

\paragraph{Dynamics of individual image TP-agreement.}
\label{app:individual_learning_dynamics}
We now focus on TP-agreement scores of individual images, as they evolve throughout the entire learning process. For the majority of examples, the score may climb up from random (0.2) to 1 in 1 epoch, it may dip down to 0 and then go up to 1 after a few epochs, or it may go rapidly down to 0. Either way, the score remains 1 or 0 during most of the learning procedure. This type of dynamic also suggests that once an example is learned by most networks, it is rarely forgotten -- this observation allow us to analyze the order of learning of specific architectures.

\begin{figure}[htb]
\begin{center}
  \centering
  \includegraphics[width=1\linewidth]{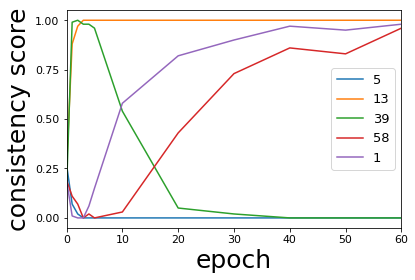}
\end{center}

\caption{TP-agreement as a function of the epoch, of 5 example images. Results achieved when training st-VGG on the small-mammals dataset.}
\label{fig:single_image_dynamics}
\end{figure}

These patterns are shown in Fig.~\ref{fig:single_image_dynamics}, and support the bi-modality results we report above. The duration in which a certain example maintains a TP-agreement 0 correlates with the order of learning: the longer it has 0 TP-agreement, the more difficult it is. A minority of the training examples exhibit different patterns of learning. For example, a few images (the green curve in Fig.~\ref{fig:single_image_dynamics}) begin with a high TP-agreement (near 1), but after a few epochs their score drops to 0 and remains there. The amount of examples with this type of dynamics is negligible.

\paragraph{Diversity in single architecture.}
The bi-modal phase we report in \S\ref{section:self-consistency}, was seen in all unmodified datasets we tested, across all architectures. Specifically, we've tested ImageNet on AlexNet ($N=22$), ResNet-50 ($N=27$), DenseNet ($N=7$). Mnist on the Mnist architecture (see \app~\ref{app:hand_crafted_architecture}) with $N=100$, CIFAR-10 and CIFAR-100 with VGG-16 ($N=20$) and st-VGG ($N=100$), tiny ImageNet with st-VGG ($N=100$), small-mammals dataset with st-VGG ($N=100$) and small st-VGG ($N=100$), and finally randomly picked super-classes of CIFAR-100, specifically "aquatic-mammals", "insects" and "household furniture" with st-VGG ($N=100$). The number of instances $N$ is chosen according to our computational capabilities. However, in all cases, picking much smaller $N$ suffice to yield the same qualitative results. 

In addition to hyper-parameters which may differ between various architectures, we also experimented with changing the hyper-parameters of st-VGG trained on the small-mammals dataset, always observing the same qualitative result. All experiments used $N=100$ instances. Specifically, we tried a large range of learning rates, learning rate decay, SGD and Adam optimizers, large range of batch sizes, dropout and L2-regularization.

\paragraph{Cross architectures diversity.}
In addition to the results in \S\ref{section:cross-consistency}, the same qualitative results were obtained for all 2 architectures we trained on the same unmodified dataset. We conducted the following experiments:  ImageNet dataset: ResNet-50 vs DenseNet, AlexNet vs DenseNet. Aquatic-mammals and small-mammals super-classes of CIFAR-100: st-VGG vs small st-VGG, Tiny ImageNet: st-VGG vs small st-VGG, CIFAR-10 and CIFAR-100: VGG19 vs st-VGG. All of which yielding similar results to the ones analyzed in \S\ref{section:cross-consistency}.

\begin{figure}[htb]
\begin{subfigure}{.23\textwidth}
  \centering
  \includegraphics[width=1\linewidth]{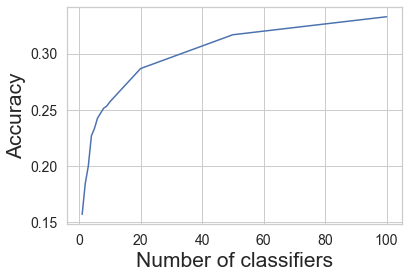}
  \caption{Grey levels}
\end{subfigure}
\begin{subfigure}{.23\textwidth}
  \centering
  \includegraphics[width=1\linewidth]{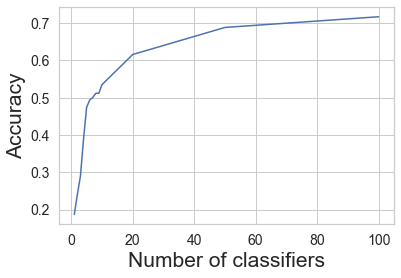}
  \caption{Transfer learning}
\end{subfigure}
\caption{Accuracy of AdaBoost trained on CIFAR-10 as a function of number of weak classifiers. a) AdaBoost trained on grey-levels. b) AdaBoost trained on the Inception-V3 trained on ImageNet penultimate representation of the CIFAR-10}
\label{fig:appendix_adaboost}
\end{figure}

\section{Other learning paradigms}
\label{app:other_learning_paradigms}
\paragraph{Boosting linear classifiers.}
We used AdaBoost \cite{hastie2009multi} with $k$ up to $100$ weak linear classifiers. We trained the AdaBoost over CIFAR-10, where each channel in each image was normalized to $0$ mean and $1$ standard deviation. Then, each image tensor was flattened into a vector. The accuracy of AdaBoost increased as we added more linear classifiers, as can be seen in Fig.~\ref{fig:appendix_adaboost}. 
In addition to the results presented in \S\ref{section:cross-consistency-other}, we also repeated the experiments for the small-mammals, fish, insect, cats and dogs, and the ImageNet cats dataset, all depicting similar results. Datasets with more classes (such as CIFAR-100 or the entire ImageNet) were ruled out, as the basic AdaBoost do not reach suffiecnt accuracy on them in order to compare them to neural networks.

\end{document}